\documentclass[sigconf,nonacm]{acmart}

\AtBeginDocument{}
\setcopyright{none}
\renewcommand\footnotetextcopyrightpermission[1]{}

\usepackage{amsmath}
\usepackage{bm}
\usepackage{booktabs}
\usepackage{multirow}
\usepackage{graphicx}
\usepackage[table]{xcolor}
\usepackage{subcaption}

\newcommand{\ours}{NSL-SLAM}

\definecolor{rankone}{RGB}{217,237,214}
\definecolor{ranktwo}{RGB}{255,242,204}
\definecolor{rankthree}{RGB}{252,228,214}
\newcommand{\best}[1]{\cellcolor{rankone}\textbf{#1}}
\newcommand{\second}[1]{\cellcolor{ranktwo}#1}
\newcommand{\third}[1]{\cellcolor{rankthree}#1}

\graphicspath{{figures/}}

\begin{document}

\title{NSL-SLAM: High-Fidelity Neural Structured-Light Depth\texorpdfstring{\newline}{ }for Practical SLAM and Reconstruction}

\author{Jiaheng Li}
\email{jiahengli25@stu.pku.edu.cn}
\affiliation{
  \institution{Wangxuan Institute of Computer Technology, Peking University}}
\affiliation{
  \institution{Beijing Academy of Artificial Intelligence, Beijing, China}}

\author{Binsheng Zhang}
\email{23371228@buaa.edu.cn}
\affiliation{
  \institution{School of Computer Science and Technology, Beihang University}}

\author{Xinhai Chang}
\email{changxinhai@stu.pku.edu.cn}
\affiliation{
  \institution{Yuanpei College, Peking University}}

\author{Wenzheng Chen}
\authornote{Corresponding author.}
\email{wenzhengchen@pku.edu.cn}
\affiliation{
  \institution{Wangxuan Institute of Computer Technology, Peking University}
}
\affiliation{
  \institution{Beijing Academy of Artificial Intelligence, Beijing, China}}

\renewcommand{\shortauthors}{Li et al.}

\begin{abstract}

Structured-light (SL) cameras power depth sensing in millions of devices, and recent neural SL decoding methods have substantially improved their depth quality.
SLAM systems can benefit greatly from such strong depth sensing, where reliable geometry enables stable tracking and faithful reconstruction.

In this work, we present {\ours}, a practical SLAM system tailored for high-fidelity structured-light depth. We first strengthen SL depth sensing: inspired by the neural structured-light (NSL) method~\cite{NSL}, we further incorporate strong monocular depth priors into the SL stereo decoding,
reducing depth RMSE by \textbf{35\%} on Replica-SL compared to NSL.

We then build a depth-centric SLAM pipeline with this stronger depth: because structured-light geometry is dense and metrically accurate, we keep it as the primary tracking signal, and add only sparse visual correspondences for geometrically degenerate cases and lightweight bundle adjustment for long-range drift.

Our depth estimator and SLAM design reinforce each other: \\ stronger depth makes a simple SLAM pipeline effective, and the depth-centric pipeline ensures this advantage transfers to downstream reconstruction. Experimentally, on the synthetic Replica-SL benchmark, {\ours} achieves the best tracking accuracy and improves reconstruction F-score by \textbf{1.6} points over the SOTA baseline under a shared-depth protocol. On a real benchmark of 8 challenging scenes, it is the only method that avoids catastrophic failure on all sequences while achieving \textbf{43.3\%} lower trajectory deviation than selected baselines. The SLAM system runs online at 20.9 FPS, demonstrating that stronger structured-light depth and depth-centric system design together enable practical, robust SLAM.

\end{abstract}


\begin{teaserfigure}
  \includegraphics[width=\textwidth]{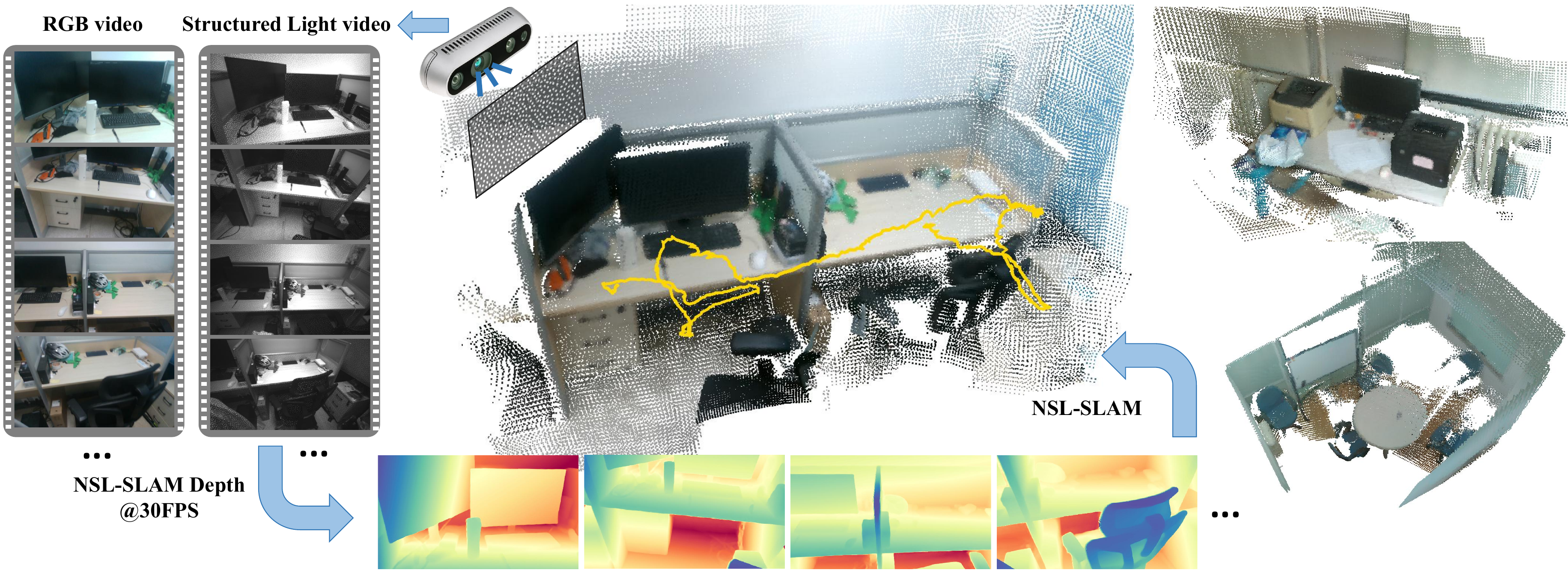}
  \caption{Our system, NSL-SLAM, captures synchronized RGB and single-shot structured-light video, predicts high-quality neural structured-light depth at \textasciitilde30 FPS, and drives a depth-centric SLAM pipeline with the improved depth. The system is stable and robust on real-world data, producing higher-quality scene reconstruction with finer structural details.}
  \Description{A teaser figure showing the input structured-light images, predicted depth maps, and the final 3D reconstruction results with detail comparisons.}
  \label{fig:teaser}
\end{teaserfigure}

\maketitle

\section{Introduction}
\label{sec:intro}

Simultaneous localization and mapping (SLAM) underpins many multimedia applications, including immersive content capture, robotics, digital twins, and more.
Among these systems, depth sensing plays a critical role, as better depth can directly benefit both camera tracking and scene reconstruction.
Structured light (SL), as one of the mainstream depth sensing approaches, is typically composed of a camera and a projector: the projector casts a known pattern onto the scene, and depth is recovered by matching and decoding the observed image against the pattern.
This design has been widely used in real products,
and recently neural structured light has emerged as a promising direction for recovering denser and more precise depth from this sensing modality~\cite{NSL}.

Structured-light depth and SLAM are naturally complementary: stronger dense depth supports more stable tracking and more faithful reconstruction, and a SLAM pipeline tailored for structured-light sensing can in turn ensure that the depth advantage is fully exploited.
In this work, we push the boundary of turning accurate neural structured-light depth into a practical SLAM system (see Fig.~\ref{fig:teaser} for an overview).

Specifically, we first enhance structured-light depth sensing.
Following neural structured light (NSL)~\cite{NSL}, we adopt a similar neural decoding scheme to acquire dense and precise SL depth.
However, unlike NSL, which applies a foundation model as post-hoc refinement, we feed foundation-model visual cues (frozen Depth Anything~\cite{DepthAnythingV2} features) into the iterative disparity updating process, leveraging the visual prior to guide every depth decoding update step and yielding better depth boundaries, finer structures, and stronger real-world generalization.

We then build a depth-centric SLAM pipeline around dense depth as the main geometric signal.
To keep the system stable in practice, we add auxiliary sparse visual correspondences to handle the cases where depth-only alignment becomes unreliable.
We further combine an online 3D Gaussian scene representation with lightweight bundle adjustment to maintain consistency over long sequences.

Based on the above design, we build \ours{}, an integrated neural structured-light SLAM system where the stronger depth and depth-centric design reinforce each other: high-fidelity geometry makes a simple pipeline effective, and the pipeline ensures that the depth advantage transfers to downstream reconstruction.
Established components are not individually claimed as novel; our contribution is their coordination around high-fidelity structured-light depth.

We evaluate \ours{} on both synthetic and real structured-light benchmarks.
Our depth estimator reduces RMSE by \textbf{35\%} on Replica-SL compared to NSL.
The full system achieves state-of-the-art tracking accuracy on Replica-SL scenes and improves reconstruction F-score by \textbf{1.6} points over the strongest baseline under a shared-depth protocol, where all methods receive the same structured-light depth input.
We further collect a real-world dataset composed of 8 challenging D435 sequences, where \ours{} is the only method that avoids catastrophic failure across all scenes while achieving \textbf{43.3\%} lower trajectory deviation than selected baselines.
The SLAM system runs online at 20.9 FPS on a single GPU, demonstrating that neural structured-light depth can serve as a practical foundation for robust SLAM and 3D reconstruction.

Our contributions are three-fold:
\begin{enumerate}
    \item We present a stronger neural structured-light depth estimator that integrates frozen monocular depth priors into the iterative neural structured-light stereo decoding process---rather than as post-hoc refinement---reducing depth RMSE by 35\% on Replica-SL over NSL.
    \item We build a depth-centric SLAM pipeline that keeps dense depth as the primary geometric signal, stabilizes degenerate cases with sparse visual correspondences, and maintains long-range consistency with an online Gaussian scene representation and lightweight bundle adjustment, achieving the best tracking on Replica-SL scenes and avoiding catastrophic failures across all 8 real sequences.
    \item We integrate the above into a complete neural structured-light SLAM system and evaluate it on three benchmarks. Ablations show that stronger depth alone reduces tracking error by an order of magnitude (1.55\,m $\to$ 0.15\,m ATE), and each subsequent component contributes measurable further improvement.
\end{enumerate}

\section{Related Work}
\label{sec:related}

\subsection{Single-Shot Structured-Light Depth Estimation}

Structured light (SL) has long been a mainstream solution for active depth sensing.
Standard SL systems recover depth by projecting a known pattern and decoding correspondences through triangulation.
Temporal coding schemes, such as Gray codes and phase shifting, can achieve high accuracy but require multiple exposures and are therefore less suitable for dynamic scenes~\cite{Salvi2004,Geng2011}.
Single-shot designs encode spatial uniqueness into one pattern, making them attractive for real-time use, but they also make decoding more sensitive to reflectance variation, occlusion, and local ambiguity.
Traditional decoding pipelines based on block matching or semi-global matching~\cite{SGM} remain efficient, yet they often struggle to preserve fine structures and boundaries in challenging scenes.

Learning-based methods have substantially improved this picture.
Early approaches such as HyperDepth~\cite{HyperDepth}, UltraStereo~\cite{UltraStereo}, and ActiveStereoNet~\cite{ActiveStereoNet} demonstrated that learned matching can outperform hand-crafted decoding in active stereo.
Later work pushed stronger decoding or supervision strategies for structured-light settings~\cite{MonoStereoFusion,PolkaLines}.
Most directly related to ours, NSL~\cite{NSL} reformulated single-shot structured-light decoding as feature-space stereo matching and showed that neural structured light can generalize surprisingly well from synthetic training data.
Our work continues this line by further strengthening the depth estimator and then asking the next system question: how should such depth be used in a practical SLAM pipeline?

Another relevant direction improves depth through completion or refinement rather than decoding alone.
Recent methods such as LingBot-Depth treat incomplete or imperfect depth as the input to a refinement model and show strong recovery performance~\cite{LingBotDepth}.
This direction is relevant because it highlights the value of strong depth priors, but it still depends on an upstream estimator for the initial metric depth and scale.
In contrast, our structured-light depth estimator improves the matching and decoding stage itself.
These directions are complementary rather than conflicting: refinement can repair a depth estimate after prediction, while our goal is to make the structured-light depth estimation itself stronger and more reliable before entering the SLAM system.

\begin{figure*}[t]
  \centering
  \includegraphics[width=\textwidth]{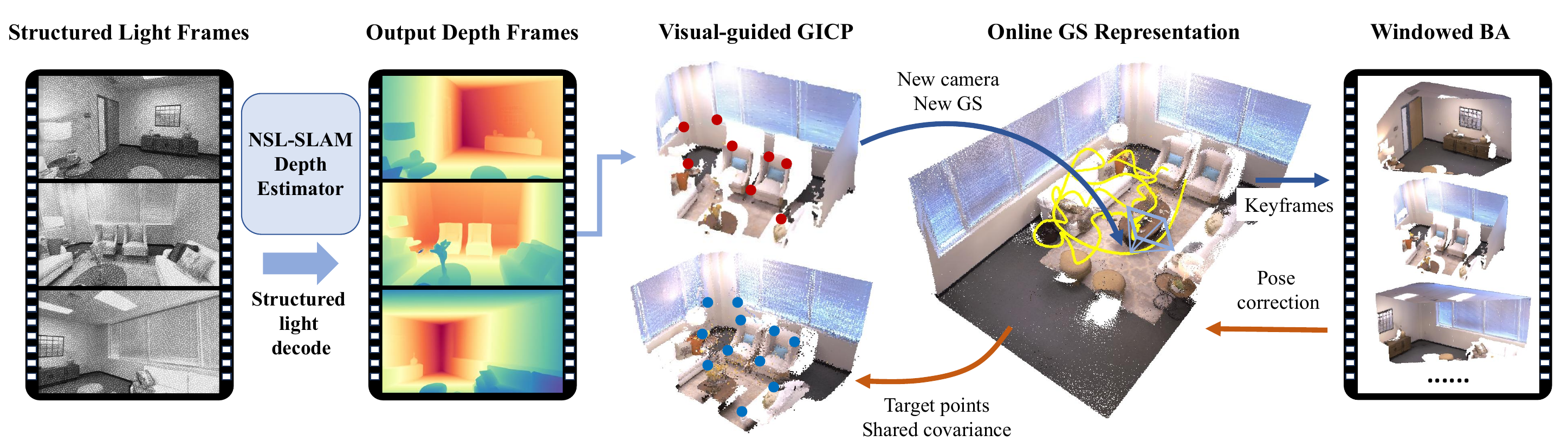}

  \caption{Pipeline of NSL-SLAM. Structured-light frames are decoded into high-quality depth maps, which drive a depth-centric SLAM pipeline. In tracking, visual-guided GICP keeps depth as the primary geometric signal and uses sparse visual correspondences as complementary cues, especially in challenging cases. The tracked frames are fused into an online Gaussian Splatting representation, and a windowed bundle adjustment refines keyframe poses for improved reconstruction quality. Backend pose corrections are fed back to the frontend through the Gaussian scene map, which absorbs global updates by updating keyframe poses and subsequently adjusting Gaussian positions during online optimization.}
  \label{fig:pipeline}
\end{figure*}

\subsection{RGB-D SLAM Under Imperfect Depth}

Classical RGB-D SLAM systems were built around the error characteristics of commodity depth sensors.
Fusion-based methods such as KinectFusion, ElasticFusion, BundleFusion, and ReFusion~\cite{KinectFusion,ElasticFusion,BundleFusion,ReFusion} improve geometric quality through repeated integration, robust alignment, and global consistency mechanisms.
Feature- and optimization-based systems such as ORB-SLAM3, BAD-SLAM, and DROID-SLAM~\cite{ORBSLAM3,BADSLAM,DROIDSLAM} combine visual correspondences, depth, and bundle adjustment to maintain stable camera motion estimates over longer sequences.
MegaSaM~\cite{MegaSaM} and VIPE~\cite{VIPE} further integrate visual foundation models to improve global consistency and system accuracy under dynamic scenes on the basis of DROID-SLAM.
These methods do not ignore imperfect depth; rather, they are designed to absorb the typical noise, holes, and drift patterns of commodity RGB-D sensing.

Neural structured-light depth changes this operating point.
Compared with conventional consumer depth, it can provide denser and sharper geometry, yet the remaining failure cases are different in character: they tend to appear in a smaller number of critical regions, but they can still destabilize tracking when dense registration becomes locally ambiguous.
This motivates a system that remains depth-centric, because dense geometry is the main strength of the modality, while still introducing targeted support for the local failure cases that can destabilize tracking.

\subsection{Neural and Gaussian RGB-D SLAM}

Neural scene representations have opened a new route for dense SLAM and reconstruction.
Implicit systems such as iMAP, NICE-SLAM, Vox-Fusion, Co-SLAM, ESLAM, and Point-SLAM~\cite{iMAP,NICESLAM,VoxFusion,CoSLAM,ESLAM,PointSLAM} use learned scene representations to couple tracking and mapping in a unified optimization loop.
More recently, explicit 3D Gaussian representations have made this direction significantly more efficient for online reconstruction, leading to systems such as SplaTAM, MonoGS, Photo-SLAM, RTG-SLAM, and CaRtGS~\cite{SplaTAM,MonoGS,PhotoSLAM,RTGSLAM,CaRtGS}.
Other learned SLAM variants further improve global consistency or representation quality through stronger optimization and learned priors~\cite{GOSLAM,SplatSLAM,LoopSplat,RemixFusion,MegaSaM,VIPE}.

Among these methods, GS-ICP SLAM~\cite{GSICPSLAM} is the closest system-level reference for our work because it provides an efficient dense-registration-plus-Gaussian framework in which the Gaussian map supports online registration.
However, it is built around a different assumption: depth is treated as sufficiently reliable for dense alignment and for directly driving the online map.
Our work departs from this assumption in two ways.
First, we explicitly design for the residual failure cases that remain even with strong structured-light depth, using auxiliary sparse visual correspondences rather than making image matching the primary tracking signal.
Second, we treat the Gaussian map as an online scene representation rather than the final geometric output, and use lightweight backend correction to coordinate global consistency with online stability.

\subsection{Positioning of This Work}

We position this work as a bridge from stronger neural structured-light depth to practical SLAM. Rather than redesigning general RGB-D SLAM with heavier optimization, we show that reliable structured-light geometry supports a simple depth-centric pipeline, where dense depth drives tracking and reconstruction, while sparse visual cues provide complementary support.

\section{Method}\label{sec:method}

\subsection{Overview and Sensing Setup}\label{sec:problem}

We consider a calibrated structured-light camera that provides, at each time step, a left infrared (IR) image with the projected pattern and a synchronized RGB image, together with either a right IR observation from the stereo baseline or the known projector pattern by treating the projector as a virtual camera. In a single-shot structured-light system, the projected pattern injects spatially varying texture in one exposure, and metric depth is recovered by decoding correspondences across the calibrated pair.
Given the recovered disparity $d$, depth follows triangulation:
\begin{equation}\label{eq:triangulation}
  z = \frac{fB}{d},
\end{equation}
where $f$ is the focal length and $B$ is the stereo baseline.
This relation makes disparity decoding the first determinant of the geometry available to SLAM.

\ours{} aims to build a better structured-light SLAM system by pairing stronger structured-light depth with a simple depth-centric design. Better per-frame depth improves local geometry and supports dense alignment, while complementary visual cues provide robustness in challenging cases.

As shown in Fig.~\ref{fig:pipeline}, the system has four tightly coupled parts: a stronger neural structured-light depth estimator, depth-centric tracking augmented by sparse visual correspondences, an online 3D Gaussian scene representation, and a lightweight streaming backend for long-range consistency.

\subsection{Stronger Neural Structured-Light Depth}\label{sec:depth}

\begin{figure}[t]
  \centering
  \includegraphics[width=\linewidth]{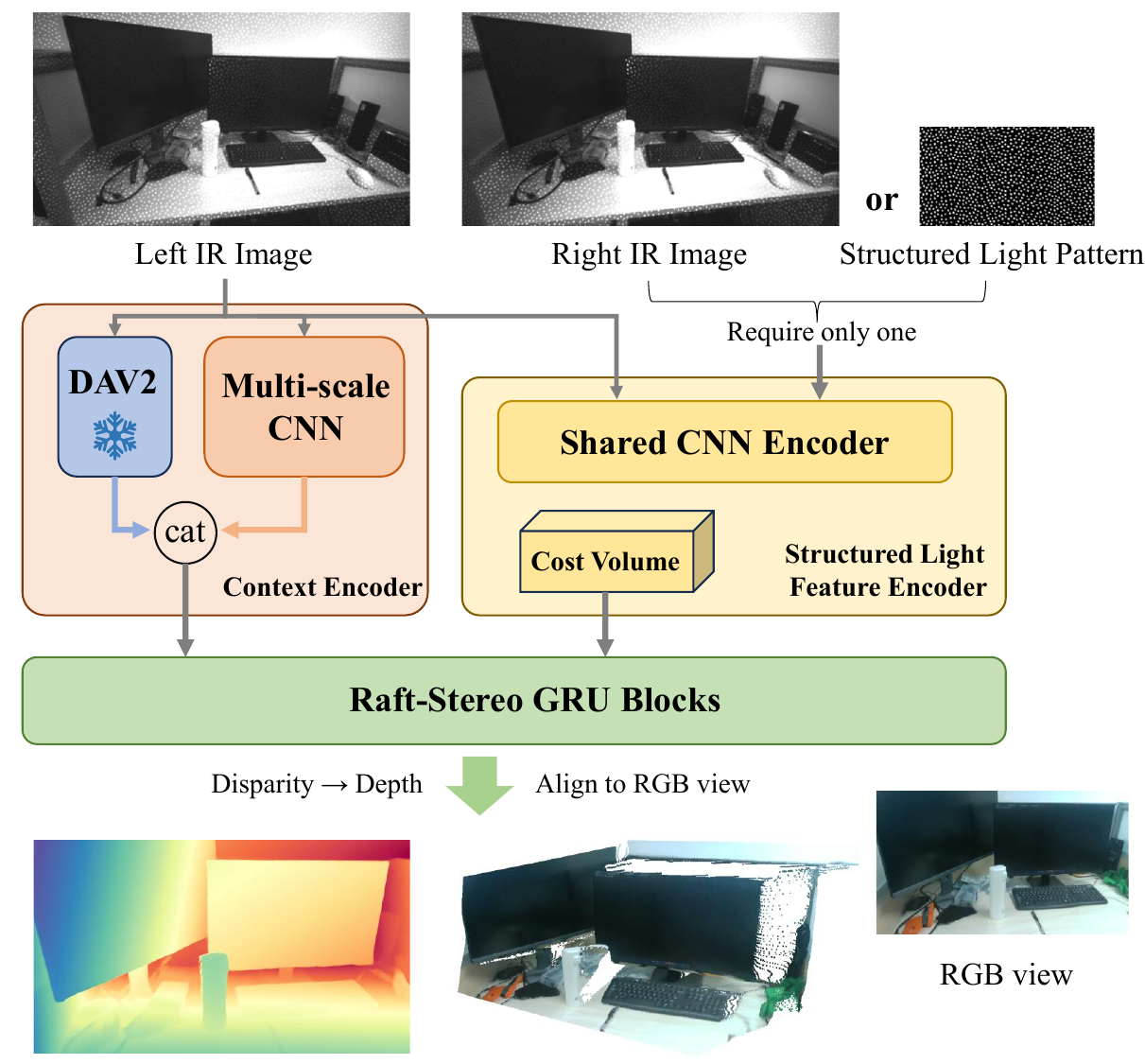}
  \caption{Depth estimation in \ours{}. Stereo matching remains geometry-driven on the structured-light pair, while frozen DAv2 features enrich the context pathway used by iterative disparity refinement.}
  \label{fig:depth_frontend}
\end{figure}

Our design is motivated by the need for stronger global information in single-shot structured-light decoding. While depth is still recovered through calibrated matching on the structured-light pair, local pattern evidence alone is often insufficient to support robust decoding across an entire image. Foundation models are well suited to provide this missing global context.

We therefore incorporate monocular foundation-model features directly into the structured-light decoder, rather than using them only after decoding as a post-hoc refinement step as in NSL~\cite{NSL}. Specifically, we build on RAFT-Stereo~\cite{RAFTStereo} and follow the feature-injection design of FoundationStereo~\cite{FoundationStereo}, feeding frozen Depth Anything v2 (DAv2)~\cite{DepthAnythingV2} features into the context pathway of iterative disparity refinement. This allows global visual features to guide structured-light matching throughout decoding, while keeping the estimator grounded in calibrated stereo geometry.
FoundationStereo targets passive RGB stereo, whereas our method decodes active IR structured light; the two modalities are not directly numerically comparable.

We choose DAv2 because, among the monocular models we tested, it was the most stable on structured-light inputs. Although newer models exist, DAv2 consistently produced more reliable features on projected-pattern and IR imagery in our setting, making it the best frozen prior for our decoder.

By letting global information participate throughout the matching process, rather than only after decoding, our model—trained on the same synthetic data protocol as NSL—produces more robust, higher-quality depth maps with richer geometric details on real captures, especially for low-cost, lower-quality structured-light sensors such as the RealSense D435.

\subsection{Depth-Centric Tracking with Sparse Visual Correspondences}\label{sec:tracking}

With stronger per-frame depth, the simple depth-centric frontend like geometric registration becomes practical.
We use Generalized ICP (GICP)~\cite{GICP} as the main tracking mechanism.
Unlike point-to-point ICP, GICP models local surface uncertainty through per-point covariance matrices, which makes it better suited to anisotropic scene geometry.
Following the GS-ICP-SLAM design~\cite{GSICPSLAM}, our online Gaussian map provides not only the registration target points but also the target-side anisotropic covariances used by GICP.

Let $\mathbf{T}=(\mathbf{R},\mathbf{t}) \in SE(3)$ be the pose of the current frame.
The dense geometric objective is
\begin{equation}\label{eq:gicp}
  E_{\text{geo}}(\mathbf{T}) =
  \sum_{i \in \mathcal{D}}
  \left(\mathbf{R}\mathbf{p}_i^{s} + \mathbf{t} - \mathbf{p}_i^{t}\right)^{\!\top}
  \left(\mathbf{C}_i^{t} + \mathbf{R}\mathbf{C}_i^{s}\mathbf{R}^{\top}\right)^{-1}
  \left(\mathbf{R}\mathbf{p}_i^{s} + \mathbf{t} - \mathbf{p}_i^{t}\right),
\end{equation}
where $\mathbf{p}_i^{s}$ and $\mathbf{p}_i^{t}$ are source and target points, and $\mathbf{C}_i^{s}$ and $\mathbf{C}_i^{t}$ are the corresponding local covariance matrices.

Our frontend is built around dense registration on structured-light depth, since the modality provides dense geometric observations at every frame. In addition to this dense geometric term, we incorporate sparse visual correspondences as complementary constraints for views where the geometry alone provides weaker localization cues, such as large planar regions or locally degraded depth.

To this end, matched image keypoints with valid depth are back-projected to 3D and used as sparse 3D--3D visual anchors. Their residual is
\begin{equation}\label{eq:anchor}
  E_{\text{anc}}(\mathbf{T}) =
  \sum_{j \in \mathcal{A}} c_j
  \left\|\mathbf{R}\mathbf{a}_j + \mathbf{t} - \mathbf{b}_j\right\|_2^2,
\end{equation}
where $\mathbf{a}_j$ and $\mathbf{b}_j$ are anchor correspondences and $c_j$ is the matcher confidence.
The full frontend objective becomes
\begin{align}
  \mathbf{T}^{*} &= \arg\min_{\mathbf{T}}
  E_{\text{geo}}(\mathbf{T}) + \alpha E_{\text{anc}}(\mathbf{T}), \label{eq:combined}\\
  \alpha &= \mathrm{clip}\!\left(
  \frac{\mathrm{tr}(\mathbf{H}_{\text{geo}})}
       {\mathrm{tr}(\mathbf{H}_{\text{anc}})},
  \alpha_{\min}, \alpha_{\max}\right), \label{eq:balance}
\end{align}
where $\mathbf{H}_{\text{geo}}$ and $\mathbf{H}_{\text{anc}}$ are the local Hessians of the geometric and anchor terms.
This adaptive scaling sets the contribution of the sparse anchor term according to the relative strength of the dense and sparse objectives. The clipping in Eq.~\eqref{eq:balance} bounds the influence of the anchor term under extreme Hessian ratios, yielding a stable combination in which dense depth remains the primary signal while sparse correspondences provide additional support when the geometric term becomes weak.
We set $\alpha_{\min}=10^{-3}$ and $\alpha_{\max}=10^{3}$. The supplementary material reports a one-factor sensitivity analysis.

\subsection{Online Gaussian Mapping and Lightweight Global Correction}\label{sec:gs}

The frontend operates against a persistent scene representation for repeated frame-to-model registration. Following GS-ICP~\cite{GSICPSLAM}, we maintain an online 3D Gaussian scene representation as the central scene memory of the system. This Gaussian map plays a dual role. On the one hand, it serves as the global map that accumulates scene structure over time. On the other hand, it acts as the registration target for the frontend by supplying local target geometry together with anisotropic covariances, in the same spirit as GICP-style registration. As the Gaussian scene is continuously optimized, it also regularizes noisy frame-wise depth observations and reduces the influence of depth outliers before they propagate into pose estimation.

Long trajectories still accumulate drift even with stronger depth and anchor-assisted tracking.
We address this with a lightweight streaming backend operating on keyframes.
For each new backend keyframe, we establish feature correspondences to sequential neighbors and loop candidates, initialize landmarks from the matched observations and their associated depth, and optimize a factor graph in GTSAM~\cite{GTSAM}.
The objective can be summarized as
\begin{equation}\label{eq:ba}
\begin{aligned}
  \mathbf{X}^{*} = \arg\min_{\mathbf{X}}
  &\sum_{(k,l)\in\mathcal{O}}
  \left\|
  (\mathbf{T}_{k}^{-1}\mathbf{T}_{l}) \ominus \hat{\mathbf{T}}_{kl}^{\text{odom}}
  \right\|_{\mathbf{\Sigma}_{o}}^{2} \\
  &+
  \sum_{(i,j)\in\mathcal{R}}
  \rho\!\left(
  \left\|\pi(\mathbf{T}_{i}, \mathbf{L}_{j}) - \mathbf{u}_{ij}\right\|_{\mathbf{\Sigma}_{r}}^{2}
  \right) \\
  &+
  \sum_{(i,j)\in\mathcal{Q}}
  \left\|
  \|\mathbf{T}_{i}^{-1}\mathbf{L}_{j}\|_{2} - \hat{r}_{ij}
  \right\|_{\mathbf{\Sigma}_{d}}^{2},
\end{aligned}
\end{equation}
where $\mathbf{T}_{i}$ are keyframe poses, $\hat{\mathbf{T}}_{kl}^{\text{odom}}$ are frontend odometry priors, $\mathbf{L}_{j}$ are landmarks, $\mathbf{u}_{ij}$ are 2D feature observations, and $\hat{r}_{ij}$ are depth-derived metric range observations.
The odometry factors preserve local motion consistency, the reprojection factors enforce multi-view feature agreement, and the range factors keep the optimization tied to metric depth.

\paragraph{Map-mediated backend feedback.}
Backend corrections are fed back through the Gaussian map rather than applied as hard updates to the frontend state. After backend optimization, the corrected keyframe poses are written back into the Gaussian scene state, and we temporarily increase the learning rate of the Gaussian center coordinates. This lets the affected Gaussians move gradually toward positions consistent with the corrected trajectory during subsequent online optimization. By absorbing global corrections through continued Gaussian optimization, the map remains a stable registration target for the frontend while still converging toward the backend-corrected geometry.

For final output, we reconstruct the scene by reprojecting the per-frame dense depth maps along the backend-optimized trajectory. With the improved depth quality in our system, trajectory-based reprojection yields a denser final reconstruction.

\section{Experiments}
\label{sec:experiments}

We evaluate \ours{} from three complementary angles.
First, we test whether the proposed decoder improves structured-light depth itself.
Second, we test whether that stronger depth leads to better tracking and reconstruction inside a complete system.
Third, we examine whether the resulting system remains stable and avoids catastrophic failure on real structured-light captures.

\subsection{Experimental Setup}
\label{sec:setup}

\paragraph{Datasets.}
We evaluate on one controlled structured-light benchmark, one video benchmark for full-system evaluation, and one real benchmark.
\textbf{NSL-Test} is the synthetic benchmark used in prior neural structured-light work; each sample provides a pair of IR observations and ground-truth depth.
\textbf{Replica-SL}

renders structured-light observations on eight Replica~\cite{Replica} scenes and provides RGB images, IR images with the projected pattern, ground-truth depth, and ground-truth camera trajectories.
Supplementary material provides the Replica-SL synthesis details.
\textbf{D435 real} contains eight indoor sequences captured with an Intel RealSense D435 and a T265 tracker.
We collect this dataset because existing real RGB-D benchmarks typically use other sensing modalities or do not provide the IR observations required for structured-light decoding.
The D435 provides left/right IR images, RGB images, and raw sensor depth, while the T265 provides only a \emph{reference} trajectory rather than ground truth,
i.e., agreement with it does not prove correctness, but large divergence is a reliable indicator of catastrophic failure.

\paragraph{Metrics.}
For \textit{depth}, we report RMSE, MAE, AbsRel, and $\delta_{1.05}$.
For \textit{Replica-SL full-system evaluation}, we report first-frame-aligned ATE and reconstruction precision / recall / F-score at 5\,cm.
For \textit{real-sequence tracking}, we report the translation gap to the T265 reference and mark sequences with average gap above 0.3\,m as failed.

\paragraph{Baselines.}
For \textit{depth}, we compare against LibSGM, NSL~\cite{NSL}, and LingBot-Depth~\cite{LingBotDepth}.
LibSGM represents the traditional semi-global-matching family used in commodity structured-light pipelines.
NSL is the most relevant prior neural structured-light decoder, and LingBot-Depth is a strong refinement-based depth method.
For \textit{SLAM}, we compare against GS-ICP SLAM~\cite{GSICPSLAM}, RTG-SLAM~\cite{RTGSLAM}, DROID-SLAM~\cite{DROIDSLAM}, ViPE~\cite{VIPE}, and RemixFusion~\cite{RemixFusion}.
On Replica-SL, all SLAM baselines receive the same RGB-D input, namely RGB images paired with our structured-light depth, so the comparison isolates downstream system design rather than upstream depth quality.

\paragraph{Implementation details.}
Detailed training settings, frontend thresholds, and backend solver parameters are given in the supplementary material.
The supplementary material also reports hyperparameter sensitivity and the runtime breakdown.
In the main paper, we evaluate the default system configuration described in Sec.~\ref{sec:method}.

\subsection{Stronger Structured-Light Depth}
\label{sec:depth_eval}

Table~\ref{tab:depth_main} first evaluates the depth estimator in isolation.
On NSL-Test, which is a highly controlled in-domain synthetic benchmark, \ours{} is competitive with NSL but not uniformly better on every metric.
This is consistent with the role of that benchmark: it mainly measures decoding quality under limited distribution shift.
The more important result for this paper appears on Replica-SL, which is the benchmark that also drives the downstream SLAM system.
There, \ours{} reduces RMSE from 0.069\,m to 0.045\,m and MAE from 0.020\,m to 0.012\,m, clearly outperforming both NSL and LingBot-Depth.
The resulting picture is therefore not one of uniform in-domain superiority, but of a stronger practical depth regime for the downstream setting studied in the rest of the paper.
In other words, the synthetic in-domain benchmark is not the main operating point this paper is optimizing for; Replica-SL and the real captures are more faithful to the regime in which the depth model is ultimately used.

\begin{table}[t]
\caption{Depth estimation on the main structured-light benchmarks used in this paper. Lower is better except $\delta_{1.05}$. LingBot-Depth is unavailable on NSL-Test because the benchmark has no RGB input.}
\label{tab:depth_main}
\centering
\small
\begin{tabular}{l|cccc}
\toprule
\multicolumn{5}{c}{NSL-Test} \\
\midrule
Method & RMSE$\downarrow$ & MAE$\downarrow$ & AbsRel$\downarrow$ & $\delta_{1.05}\uparrow$ \\
\midrule
\textbf{\ours{}} & \second{0.079} & \best{0.027} & \second{0.028} & \best{0.945} \\
NSL & \best{0.075} & \second{0.030} & \best{0.025} & \second{0.926} \\
LingBot-Depth & -- & -- & -- & -- \\
LibSGM & 1.901 & 0.357 & 0.299 & 0.753 \\
\midrule
\multicolumn{5}{c}{Replica-SL} \\
\midrule
Method & RMSE$\downarrow$ & MAE$\downarrow$ & AbsRel$\downarrow$ & $\delta_{1.05}\uparrow$ \\
\midrule
\textbf{}{\ours{}} & \best{0.045} & \best{0.012} & \best{0.005} & \best{0.989} \\
NSL & \second{0.069} & \second{0.020} & \second{0.007} & \second{0.983} \\
LingBot-Depth & 0.055 & 0.022 & 0.009 & 0.982 \\
LibSGM & 2.840 & 0.317 & 0.217 & 0.963 \\
\bottomrule
\end{tabular}
\end{table}

The contrast between NSL-Test and Replica-SL is informative.
Replica-SL introduces different scene geometry, trajectories, and rendering conditions from the training distribution, making it a better proxy for the generalization problem faced by SLAM.
The stronger Replica-SL result therefore supports the intended effect of our design: the frozen visual prior improves the decoding process itself, which matters most when the scene distribution shifts and local ambiguity increases.

\begin{figure*}[t]
\centering
\setlength{\tabcolsep}{1pt}
\renewcommand{\arraystretch}{1.0}
\begin{tabular}{c c c c c c c}
RGB & D435 & LibSGM & LingBot-Depth & NSL & Ours & \\
\includegraphics[width=0.16\textwidth]{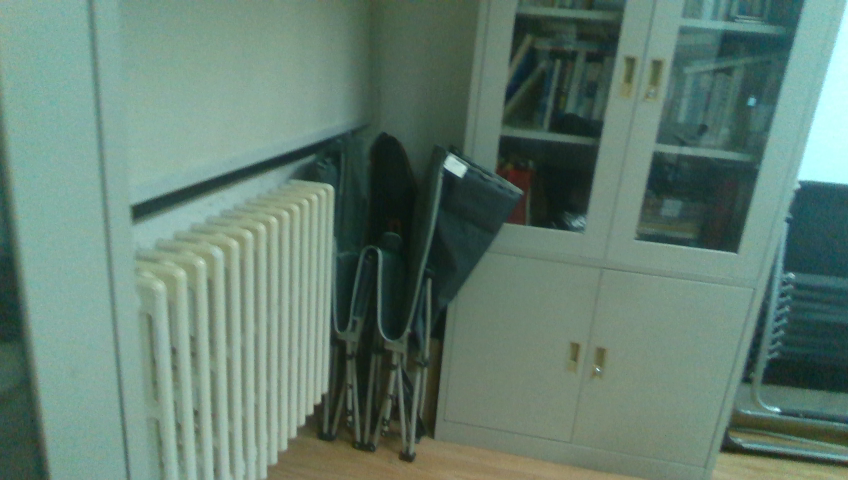} &
\includegraphics[width=0.16\textwidth]{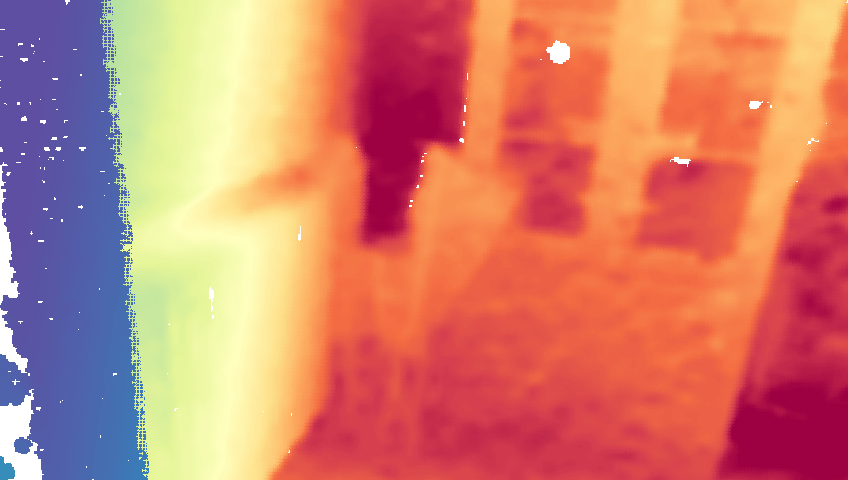} &
\includegraphics[width=0.16\textwidth]{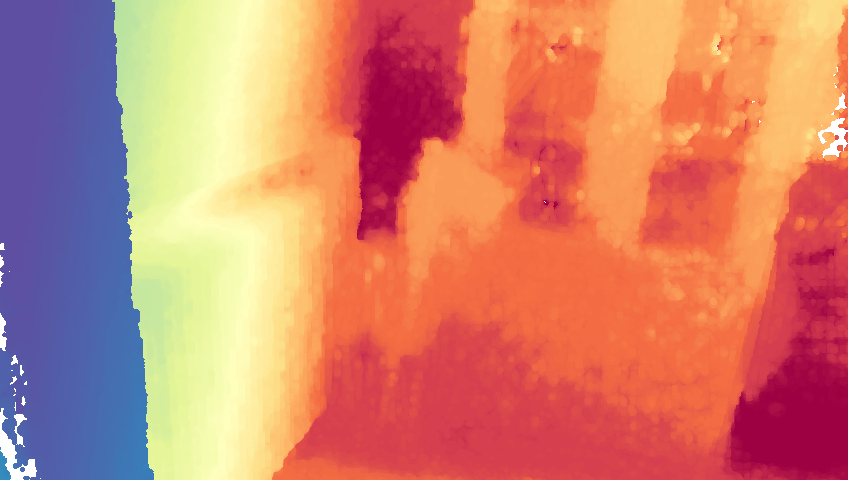} &
\includegraphics[width=0.16\textwidth]{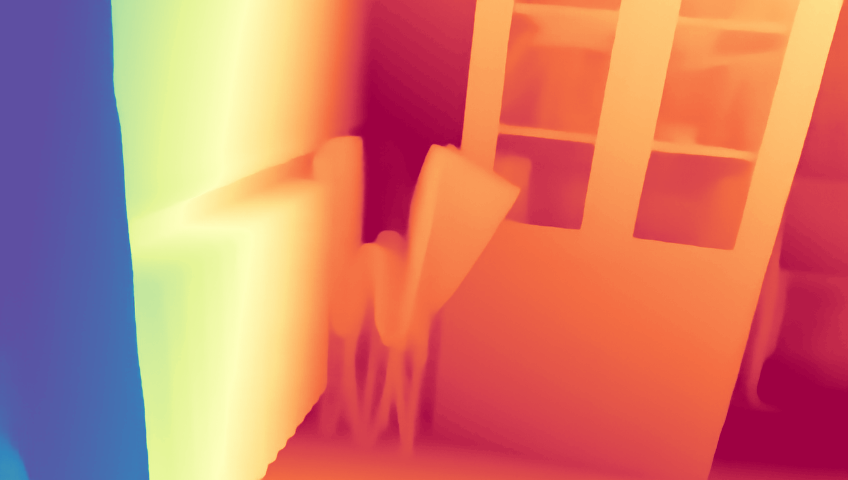} &
\includegraphics[width=0.16\textwidth]{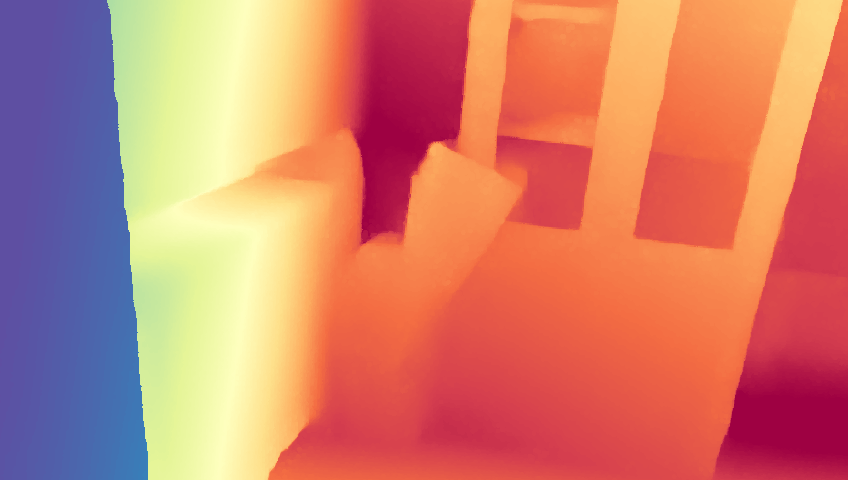} &
\includegraphics[width=0.16\textwidth]{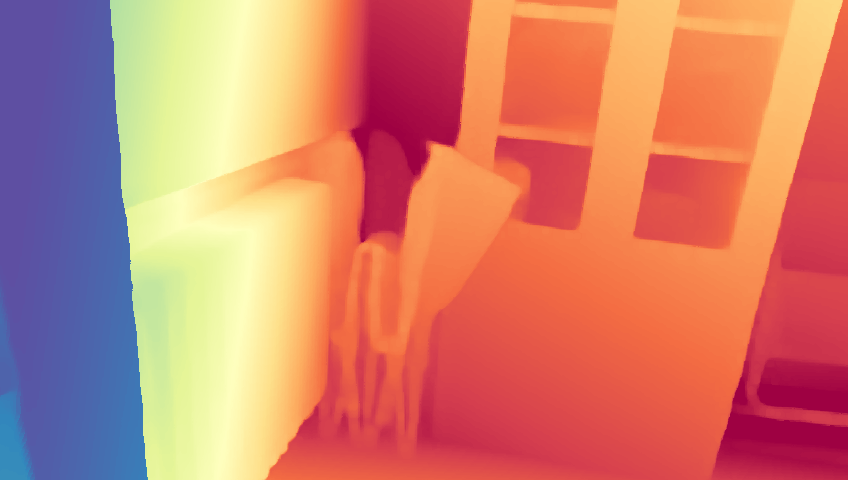} &
\raisebox{\height}{\begin{minipage}{0.02\textwidth}\centering\tiny 2.13\\\includegraphics[width=0.5\linewidth]{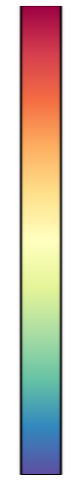}\\\tiny 0.34\end{minipage}} \\
\includegraphics[width=0.16\textwidth]{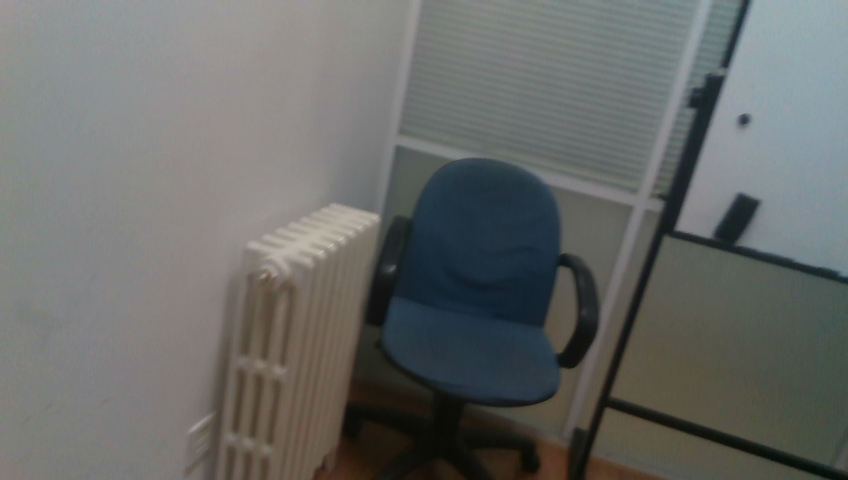} &
\includegraphics[width=0.16\textwidth]{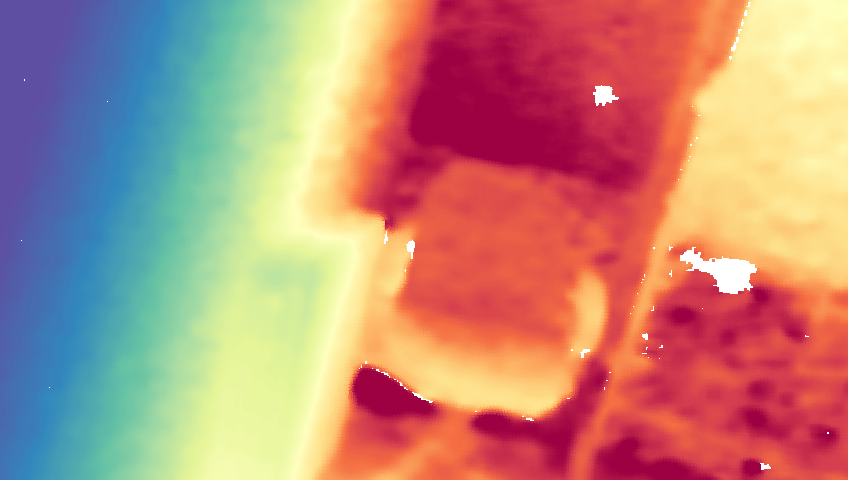} &
\includegraphics[width=0.16\textwidth]{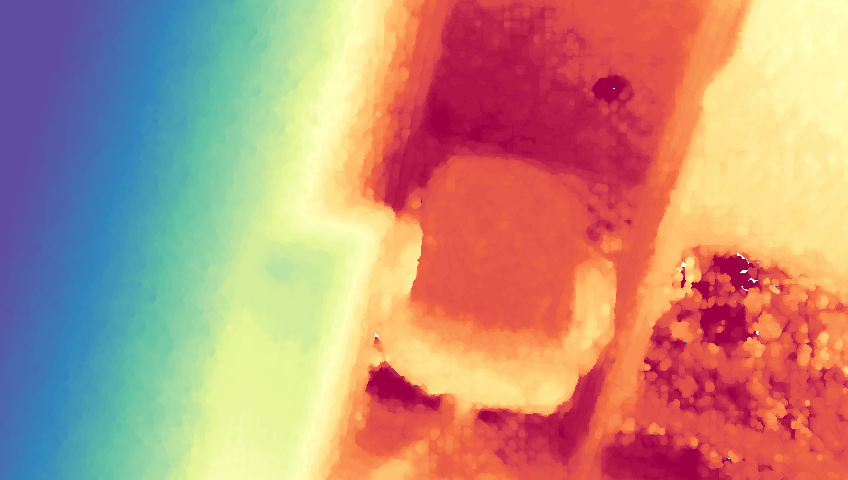} &
\includegraphics[width=0.16\textwidth]{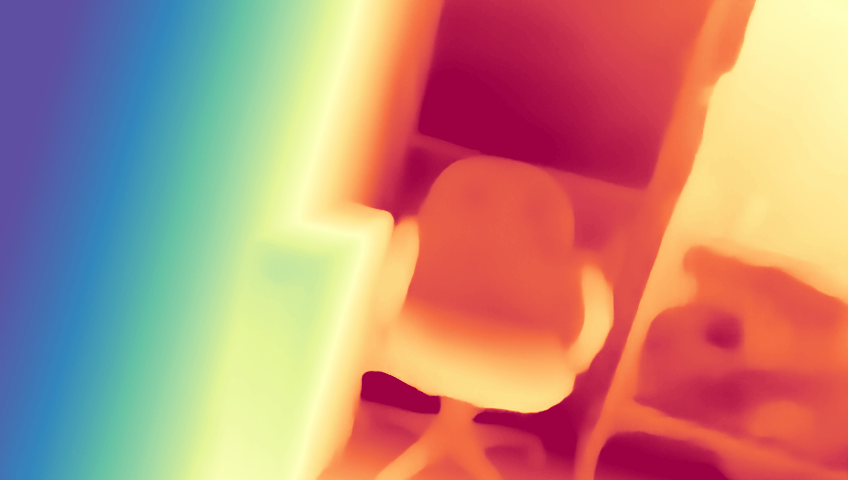} &
\includegraphics[width=0.16\textwidth]{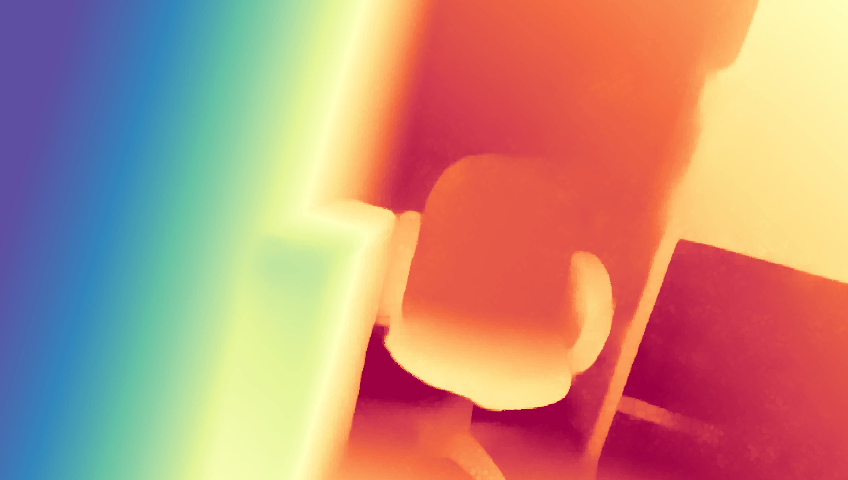} &
\includegraphics[width=0.16\textwidth]{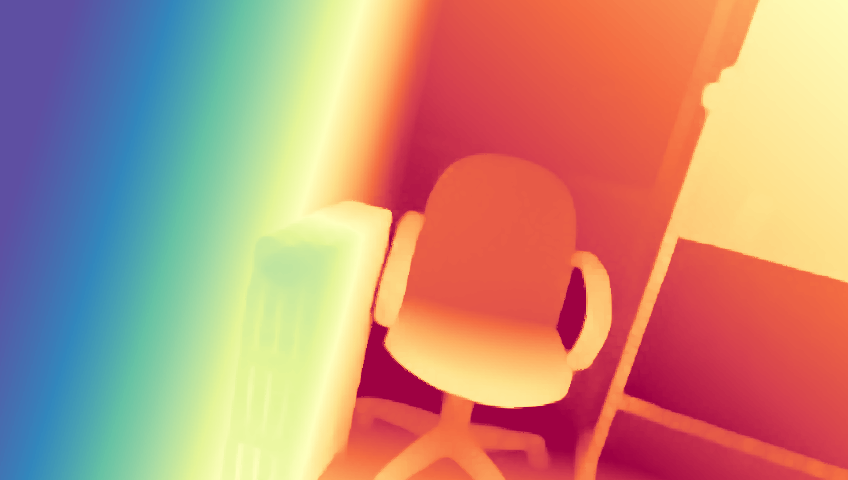} &
\raisebox{\height}{\begin{minipage}{0.02\textwidth}\centering\tiny 1.98\\\includegraphics[width=0.5\linewidth]{figures/colorbar.png}\\\tiny 0.62\end{minipage}} \\
\includegraphics[width=0.16\textwidth]{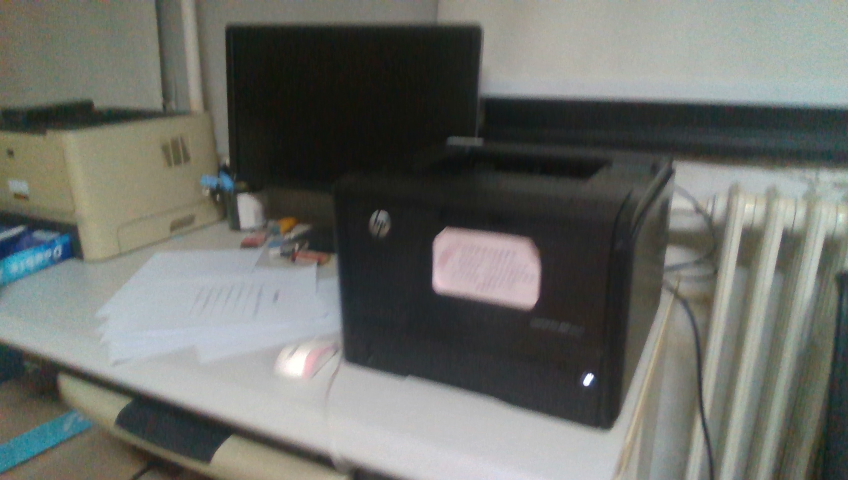} &
\includegraphics[width=0.16\textwidth]{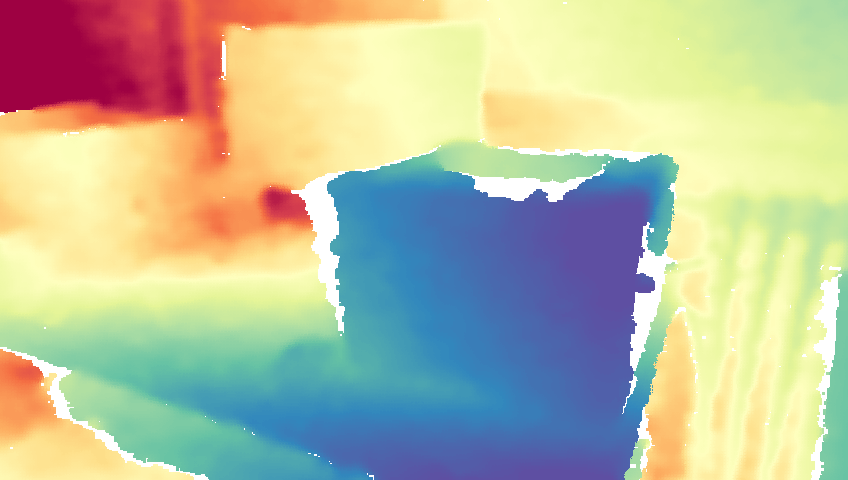} &
\includegraphics[width=0.16\textwidth]{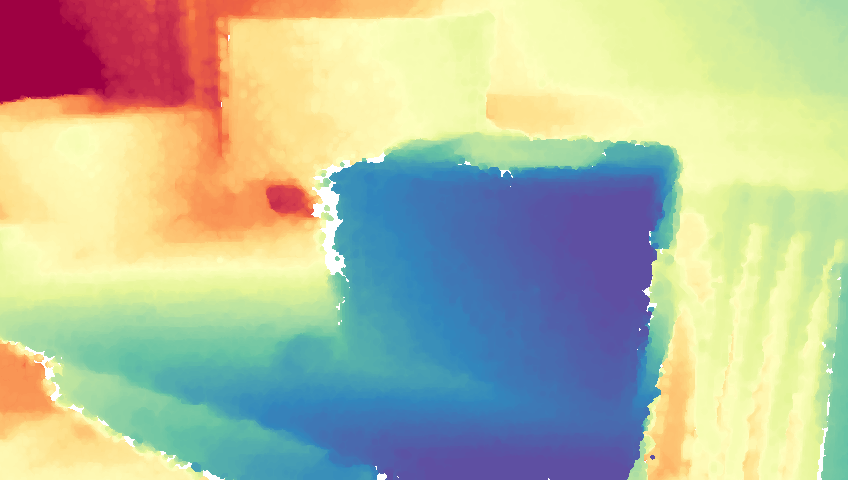} &
\includegraphics[width=0.16\textwidth]{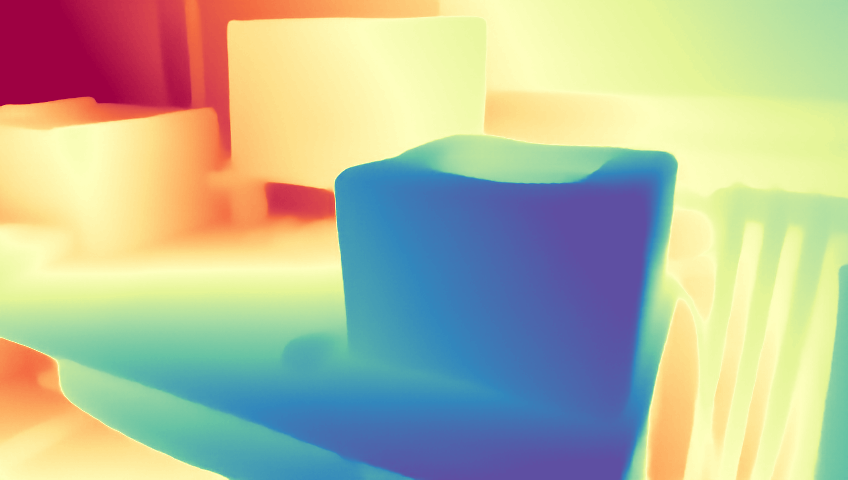} &
\includegraphics[width=0.16\textwidth]{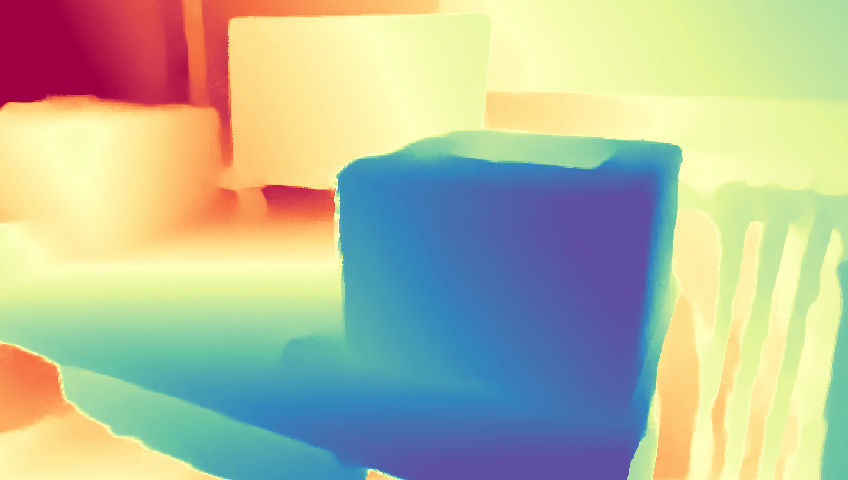} &
\includegraphics[width=0.16\textwidth]{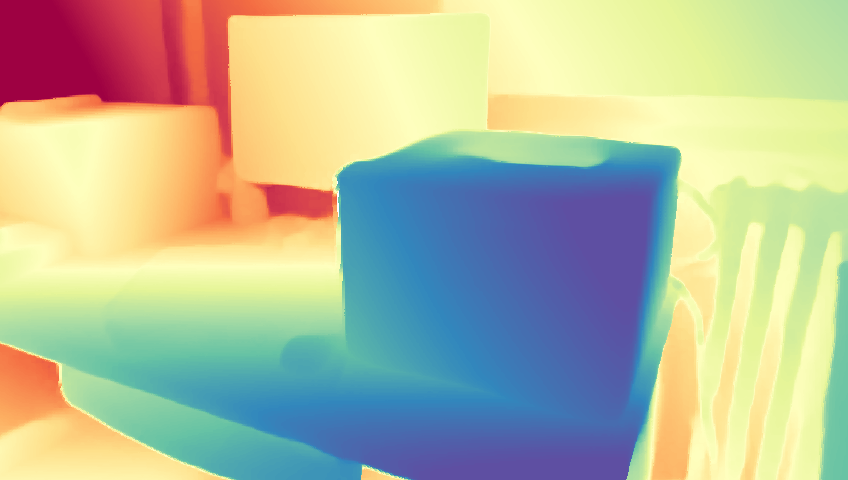} &
\raisebox{\height}{\begin{minipage}{0.02\textwidth}\centering\tiny 1.59\\\includegraphics[width=0.5\linewidth]{figures/colorbar.png}\\\tiny 0.60\end{minipage}} \\
\includegraphics[width=0.16\textwidth]{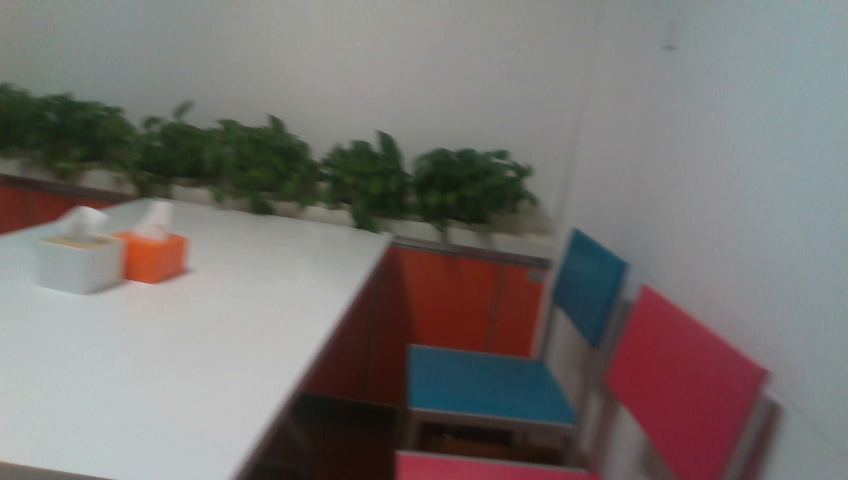} &
\includegraphics[width=0.16\textwidth]{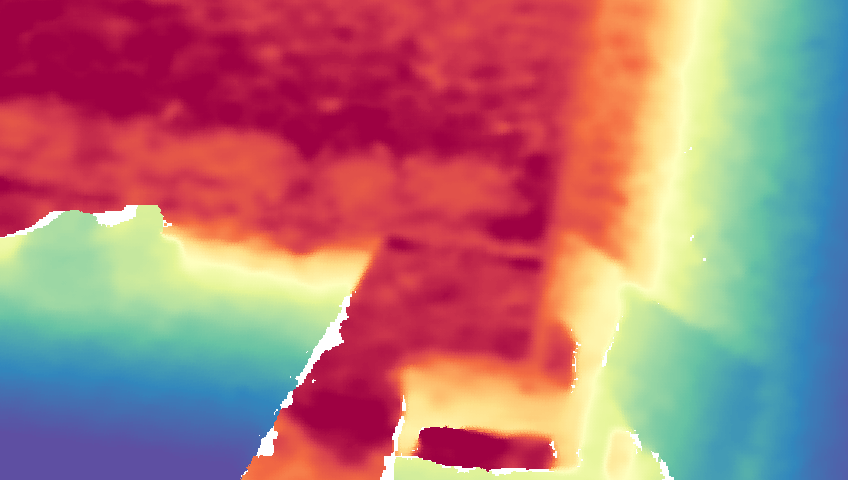} &
\includegraphics[width=0.16\textwidth]{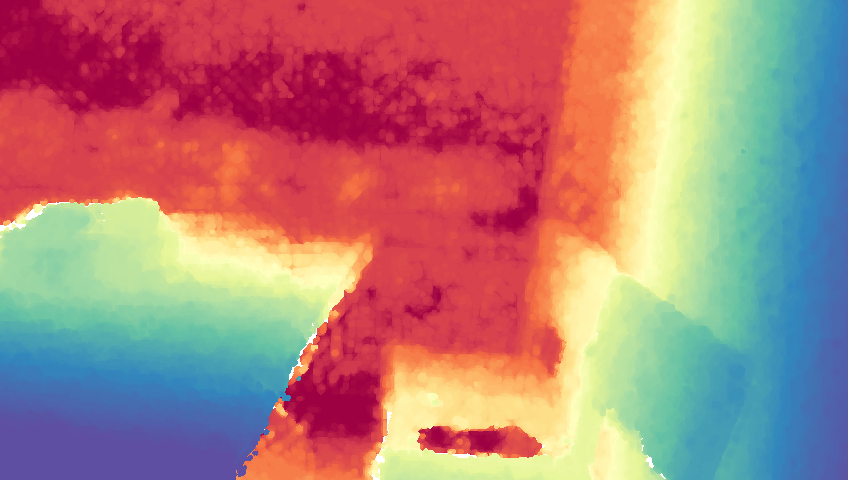} &
\includegraphics[width=0.16\textwidth]{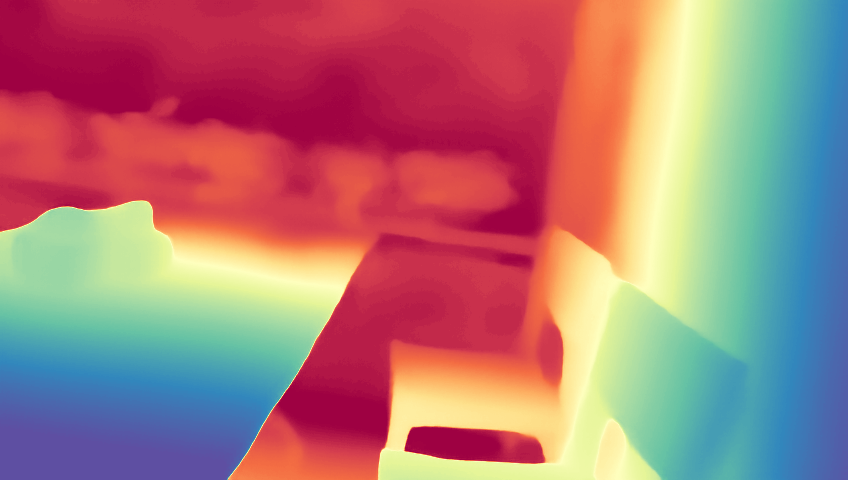} &
\includegraphics[width=0.16\textwidth]{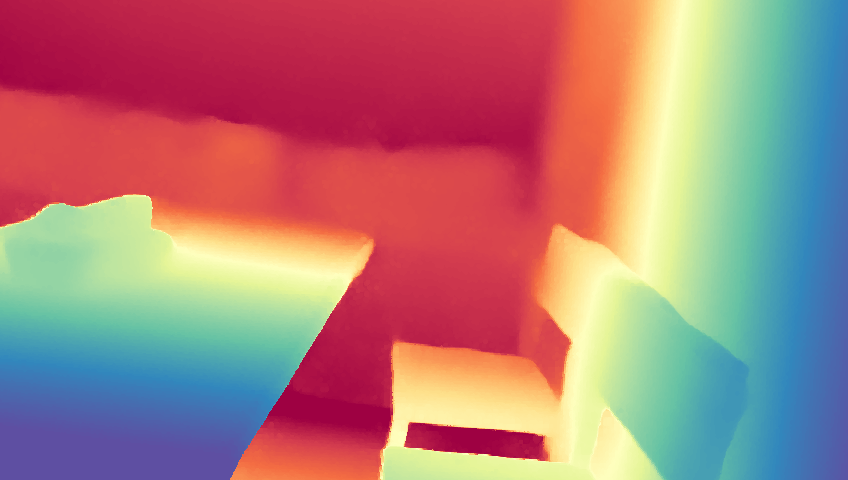} &
\includegraphics[width=0.16\textwidth]{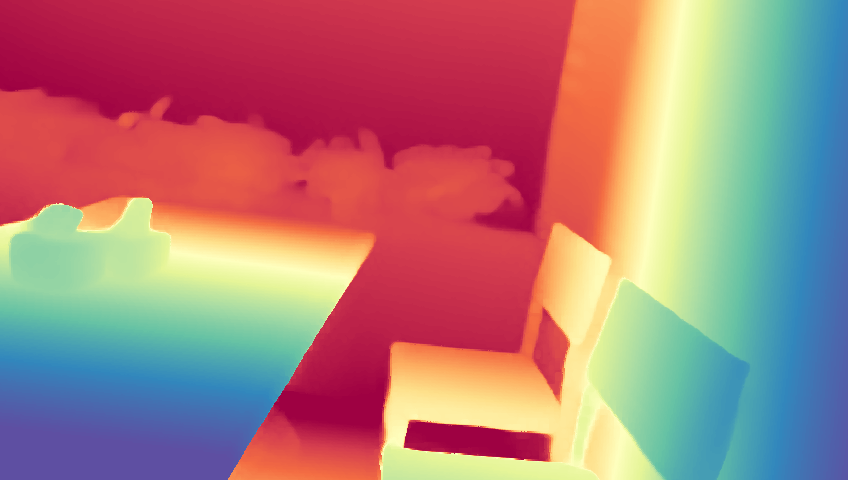} &
\raisebox{\height}{\begin{minipage}{0.02\textwidth}\centering\tiny 2.10\\\includegraphics[width=0.5\linewidth]{figures/colorbar.png}\\\tiny 0.65\end{minipage}} \\
\end{tabular}
\caption{Qualitative depth comparison on D435 real captures. Depth values are shown in meters.}
\label{fig:depth_qual}
\end{figure*}

Figure~\ref{fig:depth_qual} shows the same overall trend on real captures. Raw D435 depth and LibSGM exhibit similar active-stereo artifacts, making LibSGM a reasonable classical proxy for structured-light decoding in the experiments. LingBot-Depth and NSL remove much of this noise, but still tend to oversmooth some depth discontinuities and compress layered geometry. In contrast, \ours{} more faithfully preserves boundaries and local depth ordering. These qualitative results are consistent with the quantitative gains on Replica-SL, and suggest that the visual prior improves the decoding process itself rather than only refining depth after reconstruction.
The supplementary material provides additional D435 examples and qualitative D415 and Orbbec Gemini 335L results produced by the same synthetic-only model.

\subsection{Full-System Comparison on Replica-SL}
\label{sec:full_system}

\begin{table*}[t]
\caption{Replica-SL full-system comparison across 8 scenes. All methods use our structured-light depth. Each scene reports first-frame-aligned translation ATE RMSE (m) and reconstruction P/R/F (\%) at 5\,cm. First-frame alignment is the practically relevant setting because it matches online operation without post-hoc best-fit registration. \colorbox{rankone}{\textbf{bold}}, \colorbox{ranktwo}{yellow}, and \colorbox{rankthree}{orange} mark the top three values per scene; lower is better for ATE and higher is better for reconstruction F-score.}
\label{tab:full_system}
\centering
\scriptsize
\setlength{\tabcolsep}{3.5pt}
\renewcommand{\arraystretch}{1.15}
\resizebox{\textwidth}{!}{
\begin{tabular}{l|cc|cc|cc|cc|cc|cc|cc|cc}
\toprule
& \multicolumn{2}{c|}{room0} & \multicolumn{2}{c|}{room1} & \multicolumn{2}{c|}{room2} & \multicolumn{2}{c|}{office0} & \multicolumn{2}{c|}{office1} & \multicolumn{2}{c|}{office2} & \multicolumn{2}{c|}{office3} & \multicolumn{2}{c}{office4} \\
\cmidrule(lr){2-3} \cmidrule(lr){4-5} \cmidrule(lr){6-7} \cmidrule(lr){8-9} \cmidrule(lr){10-11} \cmidrule(lr){12-13} \cmidrule(lr){14-15} \cmidrule(lr){16-17}
Method & ATE$\downarrow$ & Recon$\uparrow$ & ATE$\downarrow$ & Recon$\uparrow$ & ATE$\downarrow$ & Recon$\uparrow$ & ATE$\downarrow$ & Recon$\uparrow$ & ATE$\downarrow$ & Recon$\uparrow$ & ATE$\downarrow$ & Recon$\uparrow$ & ATE$\downarrow$ & Recon$\uparrow$ & ATE$\downarrow$ & Recon$\uparrow$ \\
\midrule
GS-ICP & \third{0.0235} & \best{97.3 / 89.7 / 93.4} & 0.0680 & \third{96.2 / 89.4 / 92.7} & 0.0569 & \third{97.3 / 86.4 / 91.5} & 0.2565 & 86.2 / 76.5 / 81.1 & \second{0.0116} & \best{99.5 / 78.4 / 87.7} & 0.2951 & 75.3 / 73.8 / 74.6 & \second{0.0331} & \second{94.2 / 87.4 / 90.7} & 0.5242 & 71.0 / 86.9 / 78.2 \\
DROID & 0.0457 & \third{94.8 / 90.9 / 92.8} & 0.0367 & 93.7 / 89.5 / 91.5 & \third{0.0286} & \second{97.4 / 87.2 / 92.0} & 0.0592 & \third{90.4 / 82.3 / 86.2} & 0.0476 & \second{99.3 / 77.8 / 87.3} & \second{0.0467} & \second{92.1 / 84.7 / 88.3} & 0.0508 & 85.4 / 89.1 / 87.2 & 0.0738 & \third{90.2 / 88.6 / 89.4} \\
ViPE & 0.0574 & 88.5 / 88.9 / 88.7 & \best{0.0134} & \best{99.0 / 89.6 / 94.1} & 0.0660 & 85.3 / 85.9 / 85.6 & 0.0517 & \second{89.5 / 83.6 / 86.5} & 0.1549 & 65.5 / 70.9 / 68.1 & 0.1176 & \third{69.3 / 82.9 / 75.5} & \third{0.0458} & \third{88.8 / 87.6 / 88.2} & \third{0.0502} & \second{97.5 / 87.1 / 92.0} \\
RemixFusion & \best{0.0198} & 80.5 / 88.4 / 84.3 & \third{0.0223} & 73.3 / 88.3 / 80.1 & \second{0.0142} & 71.7 / 87.0 / 78.6 & \best{0.0100} & 71.9 / 81.4 / 76.4 & \third{0.0169} & 78.9 / 79.2 / 79.1 & 0.2944 & 60.8 / 65.3 / 63.0 & 0.5164 & 60.0 / 65.1 / 62.5 & \second{0.0488} & 78.9 / 87.0 / 82.8 \\
RTG-SLAM & 0.1333 & 62.5 / 77.2 / 69.1 & 0.2484 & 65.3 / 84.5 / 73.6 & 0.0682 & 82.8 / 81.8 / 82.3 & \third{0.0317} & 89.7 / 80.5 / 84.9 & 0.0326 & 90.1 / 74.6 / 81.6 & \third{0.0650} & 81.2 / 79.1 / 80.1 & 0.0648 & 73.4 / 79.3 / 76.2 & 0.0794 & 67.8 / 75.0 / 71.2 \\
\ours{} & \second{0.0199} & \second{97.7 / 88.8 / 93.1} & \second{0.0172} & \second{99.3 / 87.5 / 93.0} & \best{0.0083} & \best{99.4 / 86.3 / 92.4} & \second{0.0206} & \best{99.1 / 80.6 / 88.9} & \best{0.0099} & \third{99.6 / 77.1 / 86.9} & \best{0.0141} & \best{97.5 / 83.0 / 89.7} & \best{0.0181} & \best{95.2 / 86.6 / 90.7} & \best{0.0156} & \best{99.2 / 86.6 / 92.5} \\
\bottomrule
\end{tabular}
}
\end{table*}

Table~\ref{tab:full_system} evaluates the full system on Replica-SL using first-frame-aligned ATE, rather than post-hoc best-fit alignment, so that the reported trajectory error reflects the drift behavior faced in practical online operation. Under this protocol, all compared SLAM methods receive the same neural structured-light depth input, and the remaining difference lies in how effectively each system exploits that depth for registration, stabilization, and global correction.

Under this setting, \ours{} achieves the best ATE on five scenes, the best or tied reconstruction F-score on five scenes, and remains within the top three reconstruction results on all eight scenes. These results indicate that the proposed system makes consistently strong use of structured-light geometry across both trajectory estimation and final reconstruction.

This trend is consistent with the design in Sec.~\ref{sec:method}. Our method keeps dense structured-light depth as the primary geometric signal, uses the Gaussian scene map for stable dense registration, introduces sparse visual constraints only when local geometry becomes ambiguous, and applies lightweight backend optimization to recover long-range consistency. Together, these components improve robustness in exactly the cases where dense geometry alone becomes unreliable, while preserving the advantages of accurate metric depth when it is available.

\subsection{Real-World D435 Sequences}
\label{sec:d435}

\begin{table*}[t]
\caption{D435 real-world tracking stability. Values are first-frame-aligned translation gaps to the T265 reference trajectory (m). \textcolor{red}{\textbf{X}} marks sequences with gap $> 0.3$\,m. The T265 trajectory is not ground truth: agreement with it does not prove correctness, but large divergence is a reliable indicator of catastrophic failure.}
\label{tab:d435_ref}
\centering
\small
\begin{tabular}{l|cccccccc|c}
\toprule
Method & printer & discussion\_room & office & treadmill & tea\_room & stairs & workstation & robotic\_arm & \textbf{Failed} \\
\midrule
GS-ICP & 0.044 & 0.197 & 0.107 & 0.360 \textcolor{red}{\textbf{X}} & 1.423 \textcolor{red}{\textbf{X}} & 0.581 \textcolor{red}{\textbf{X}} & 0.088 & 0.742 \textcolor{red}{\textbf{X}} & \textbf{4/8} \\
DROID & 0.040 & 0.148 & 0.084 & 0.066 & 0.317 \textcolor{red}{\textbf{X}} & 0.515 \textcolor{red}{\textbf{X}} & 0.095 & 0.244 & \textbf{2/8} \\
ViPE & 0.128 & 0.097 & 0.158 & 0.113 & 0.485 \textcolor{red}{\textbf{X}} & 0.387 \textcolor{red}{\textbf{X}} & 0.204 & 0.308 \textcolor{red}{\textbf{X}} & \textbf{3/8} \\
RemixFusion & 0.039 & 0.121 & 0.102 & 0.141 & 0.155 & 1.457 \textcolor{red}{\textbf{X}} & 0.095 & 0.079 & \textbf{1/8} \\
RTG-SLAM & 0.039 & 0.199 & 0.186 & 0.154 & 0.134 & 0.463 \textcolor{red}{\textbf{X}} & 0.091 & 0.229 & \textbf{1/8} \\
\ours{} & 0.034 & 0.112 & 0.104 & 0.110 & 0.130 & 0.156 & 0.086 & 0.122 & \best{\textbf{0/8}} \\
\bottomrule
\end{tabular}
\end{table*}

Table~\ref{tab:d435_ref} evaluates robustness on real sequences against the T265 reference tracker. Unlike Replica-SL, this benchmark does not provide ground-truth trajectories, so it is not meant to measure absolute trajectory accuracy. Instead, it tests whether a SLAM system remains in the same operational regime as a reliable real-time reference or suffers severe drift and tracking failure on challenging captures. Under this criterion, \ours{} is the most stable method: it is the only approach with no failed sequence, whereas every baseline fails on at least one sequence and GS-ICP fails on half of them.

The result is important because it mirrors the intended system behavior in real capture conditions.
Stronger depth reduces the amount of local geometric corruption entering the frontend, sparse visual correspondences stabilize the ambiguous cases that remain, and lightweight global correction prevents those local failures from accumulating into large trajectory drift.

\begin{figure*}[t]
\centering
\setlength{\tabcolsep}{3pt}
\renewcommand{\arraystretch}{1.0}
\begin{tabular}{c c c c}
D435+T265: View1 & \textbf{Ours: View1} & D435+T265: View2 & \textbf{Ours: View2} \\
\includegraphics[width=0.24\textwidth]{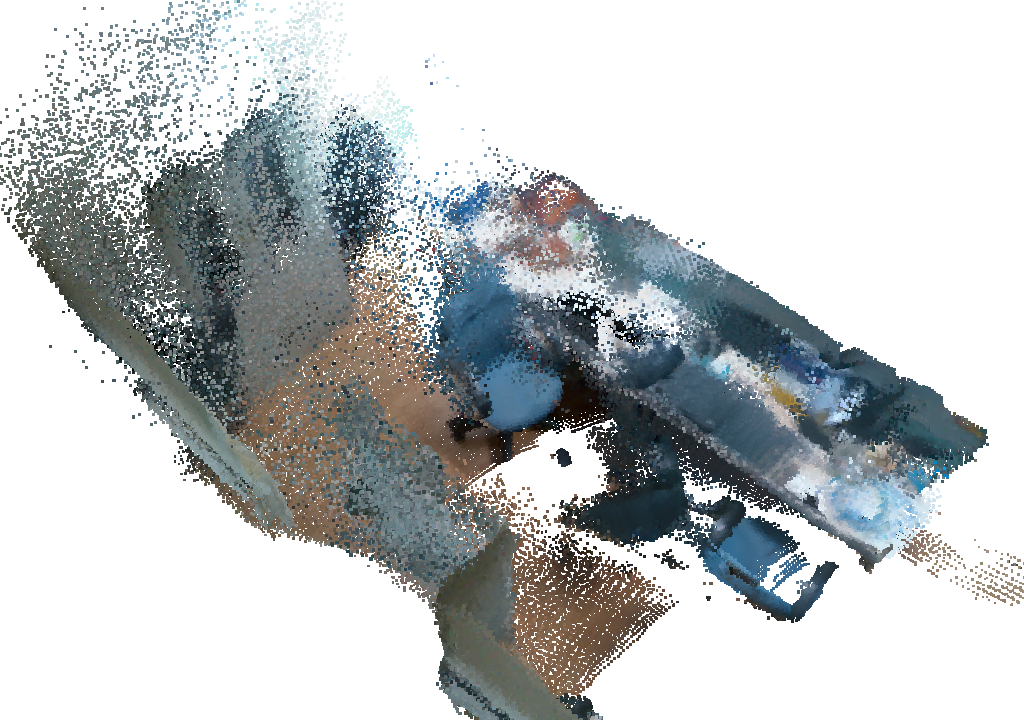} &
\includegraphics[width=0.24\textwidth]{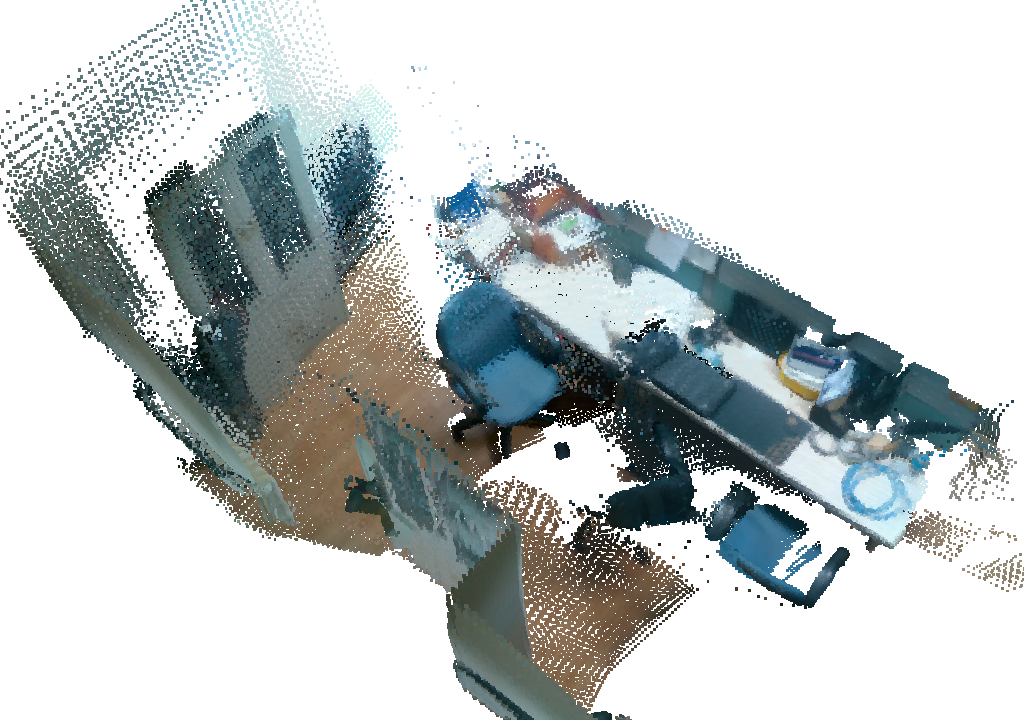} &
\includegraphics[width=0.24\textwidth]{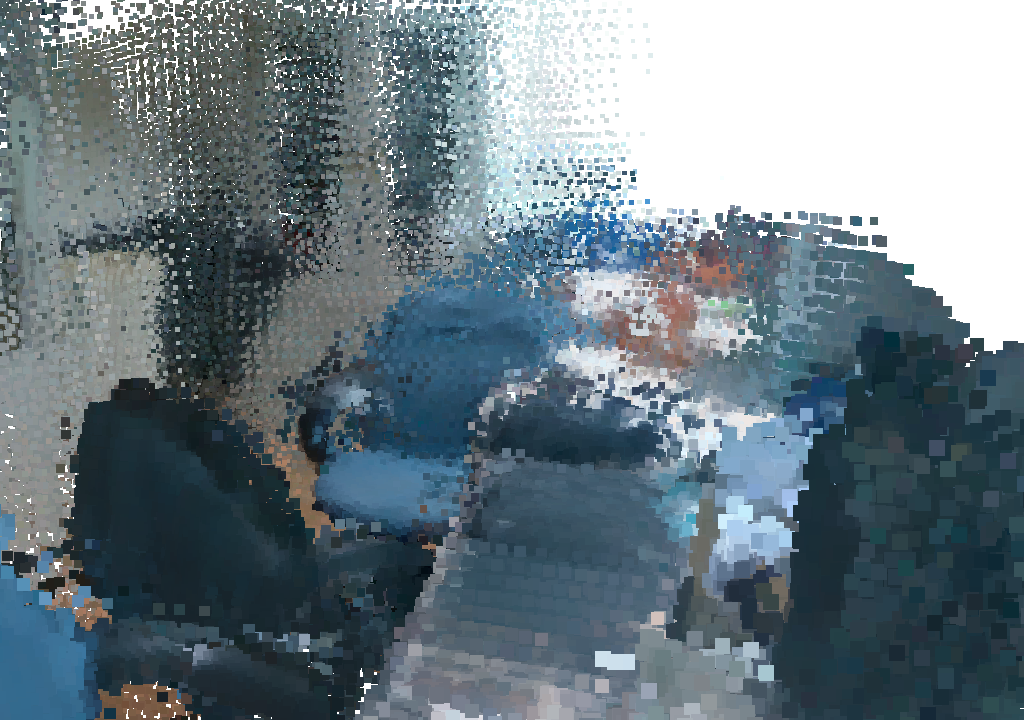} &
\includegraphics[width=0.24\textwidth]{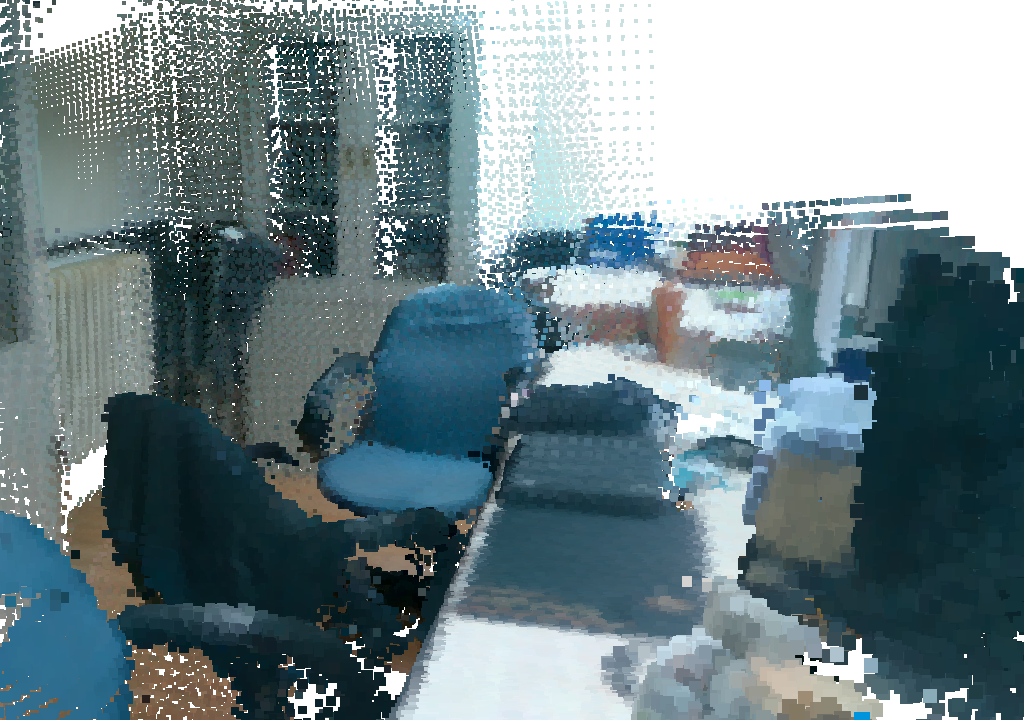} \\

\includegraphics[width=0.24\textwidth]{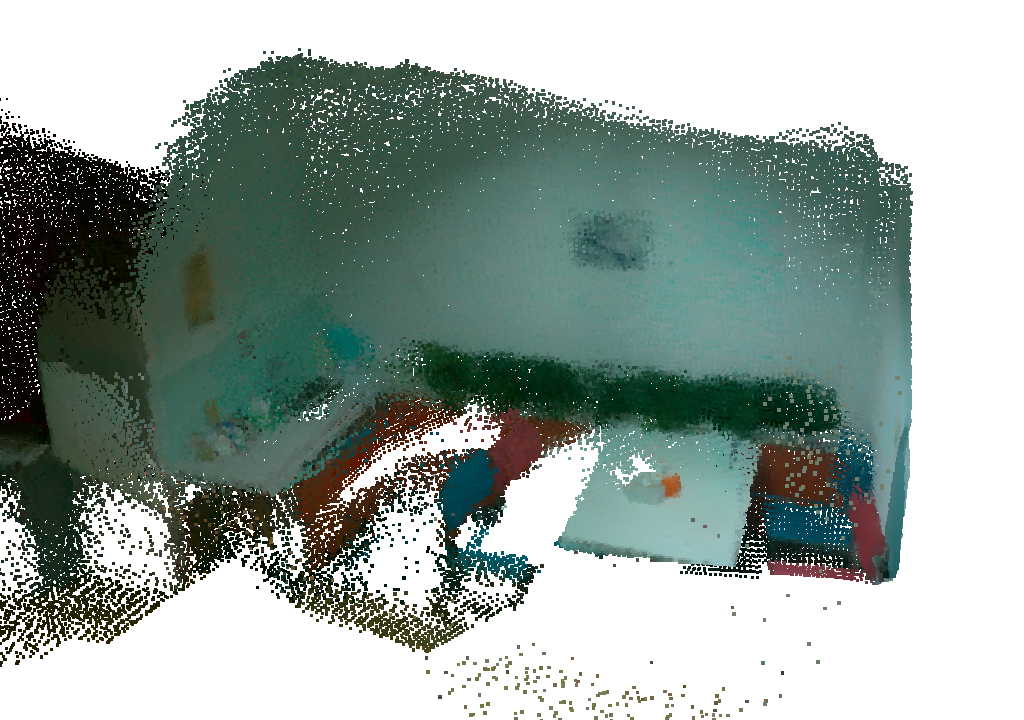} &
\includegraphics[width=0.24\textwidth]{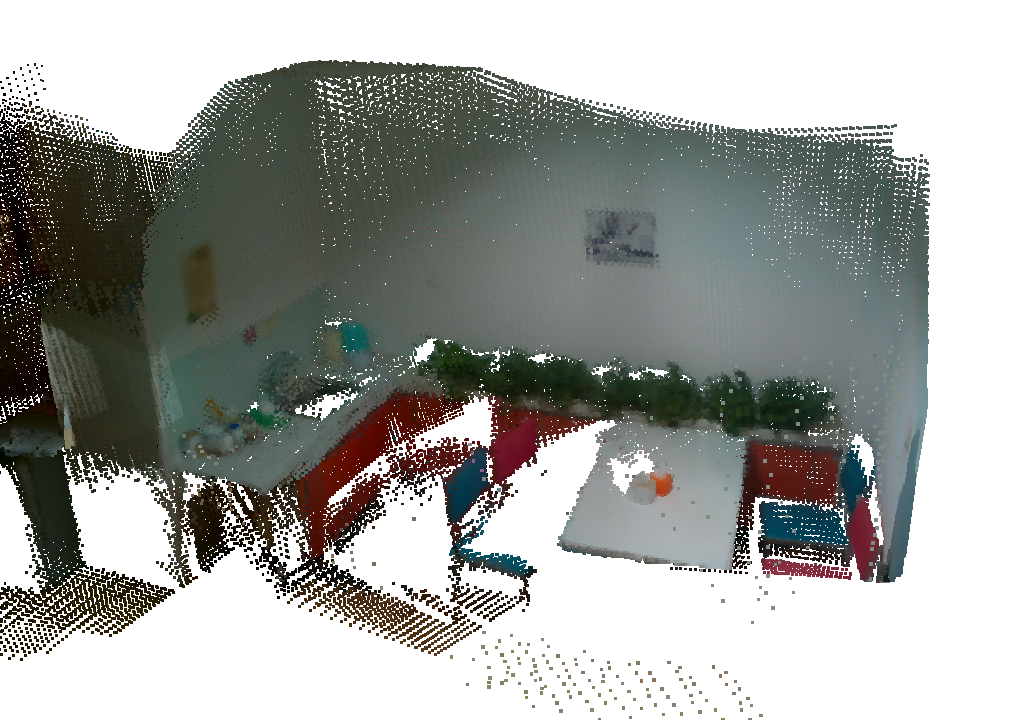} &
\includegraphics[width=0.24\textwidth]{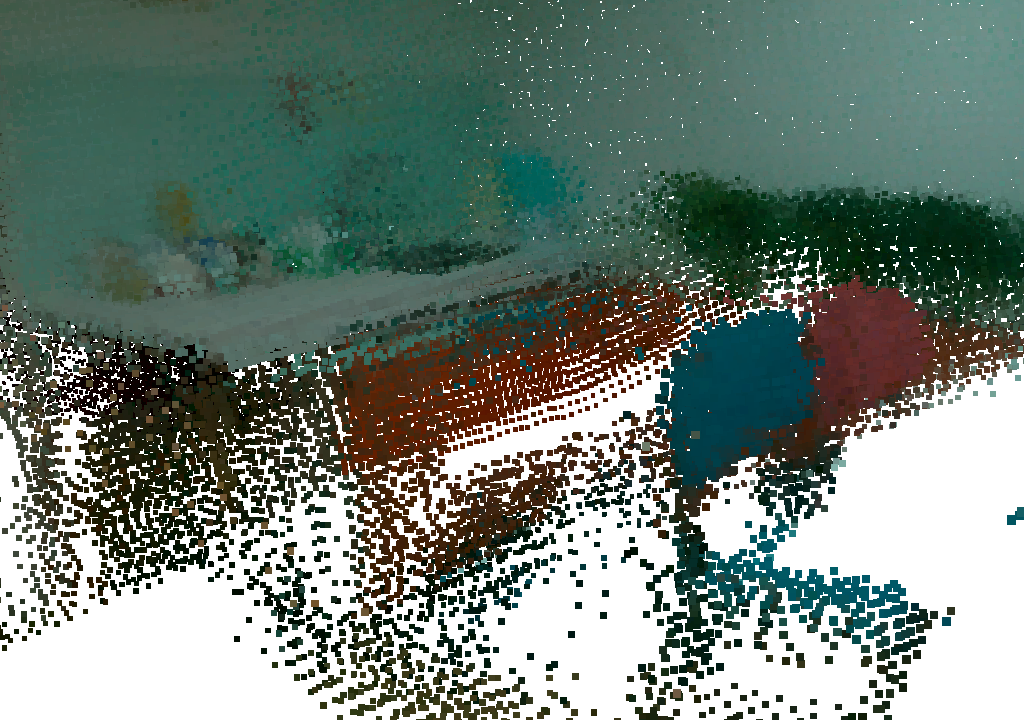} &
\includegraphics[width=0.24\textwidth]{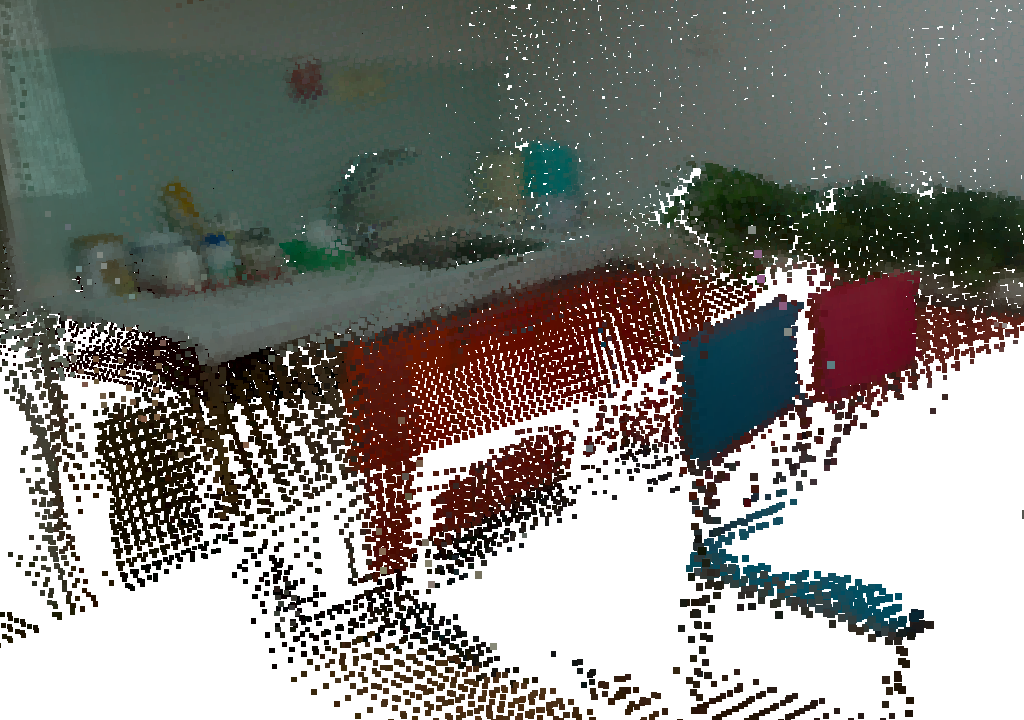} \\

\includegraphics[width=0.24\textwidth]{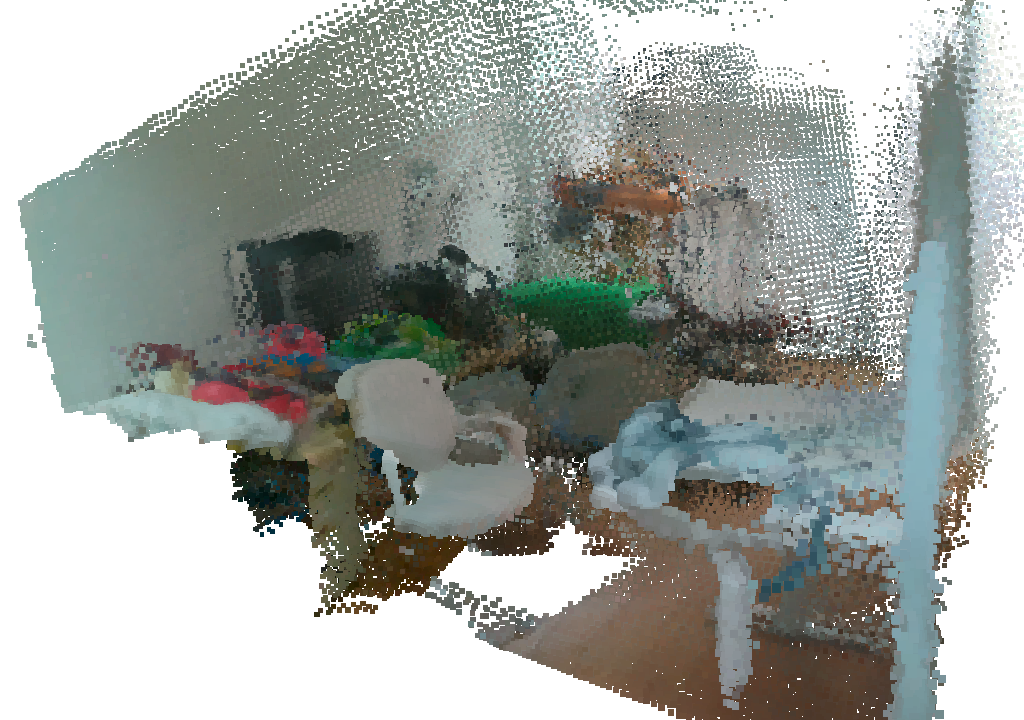} &
\includegraphics[width=0.24\textwidth]{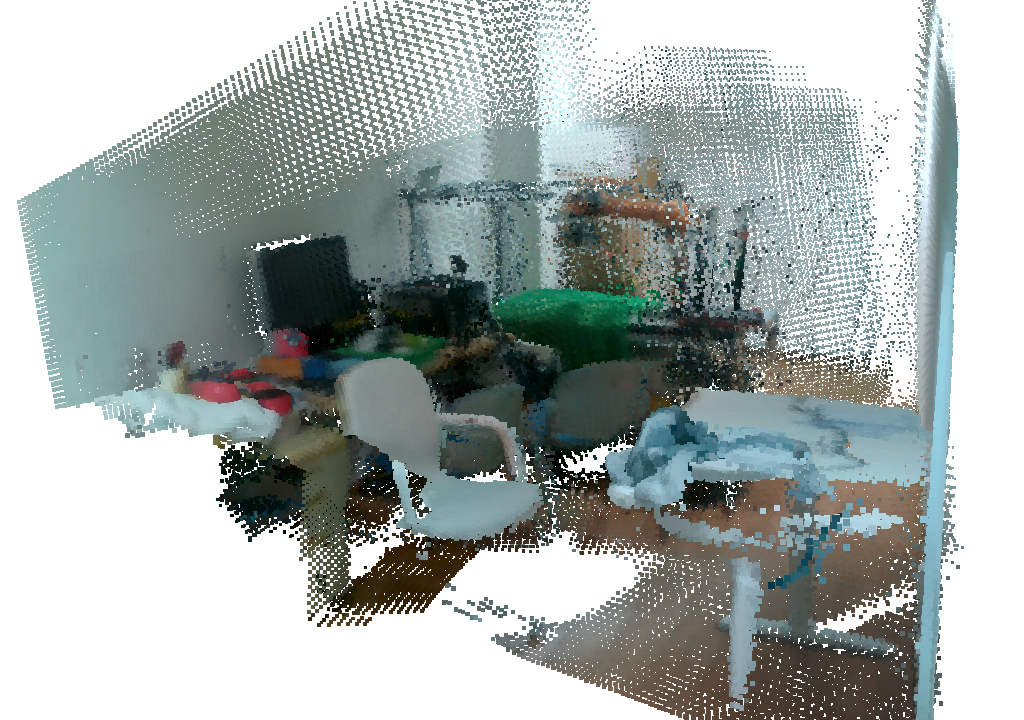} &
\includegraphics[width=0.24\textwidth]{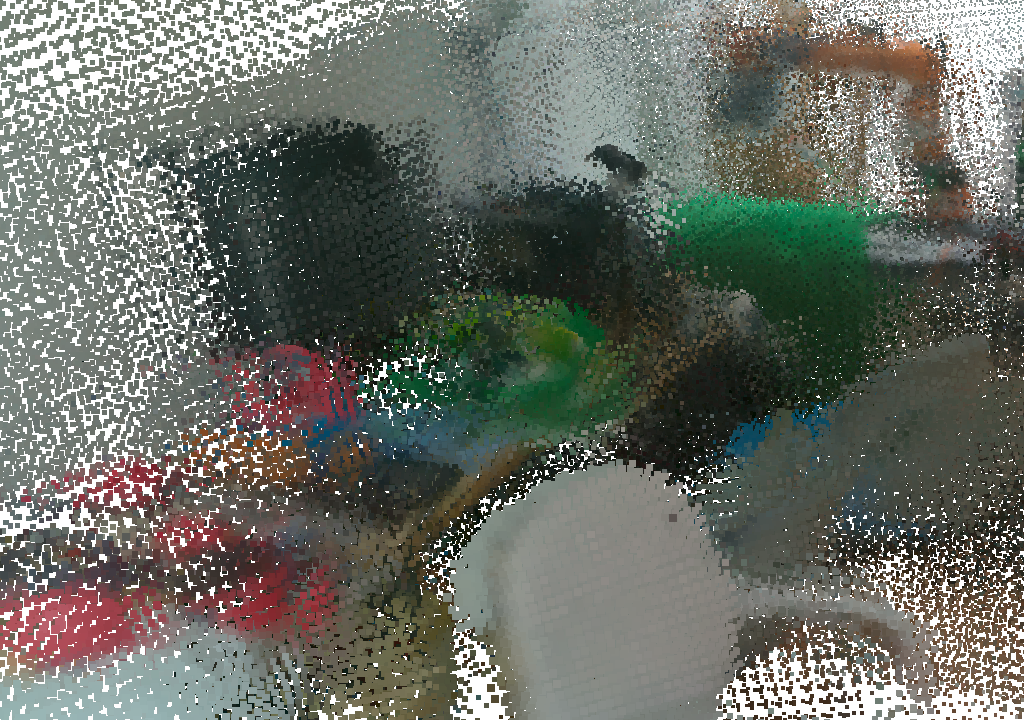} &
\includegraphics[width=0.24\textwidth]{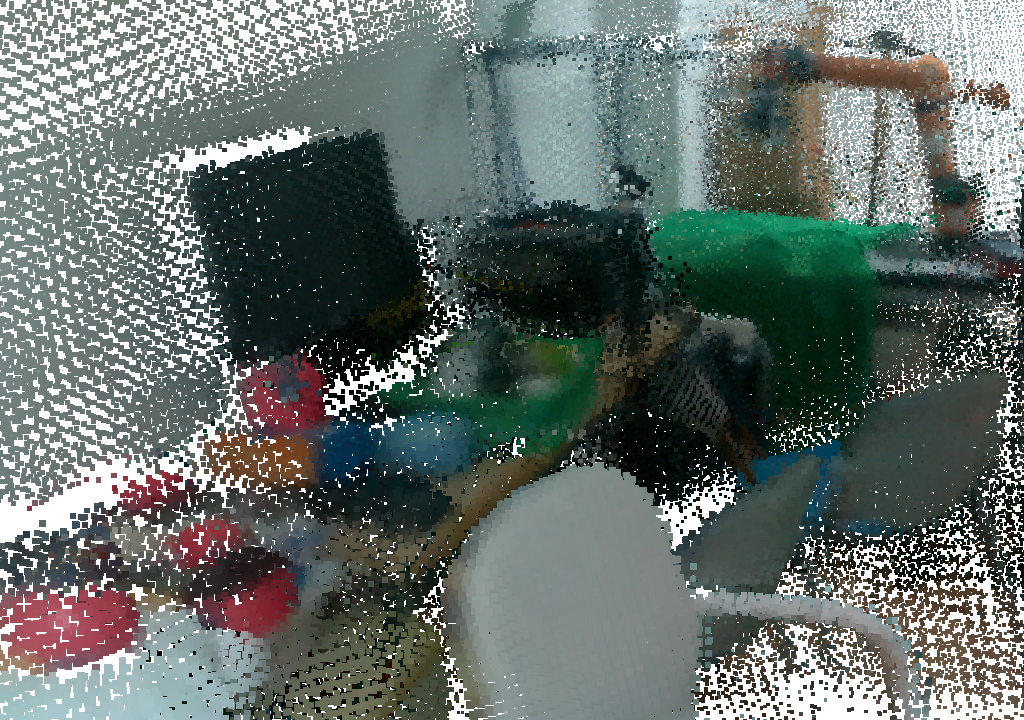} \\
\end{tabular}
\caption{Real-data reconstruction comparison between a commodity baseline formed by D435 sensor depth with the T265 reference trajectory and \ours{}.}
\label{fig:recon_qual}
\end{figure*}

Figure~\ref{fig:recon_qual} further shows that this robustness advantage carries over to the final reconstruction. We compare against a practical commodity reference pipeline formed by D435 depth and the T265 trajectory. Even relative to this strong baseline, \ours{} reconstructs cleaner surfaces and less duplicated structure across all three scenes, indicating that improved depth quality and the downstream SLAM design jointly improve the final geometry.
The supplementary material provides additional real-scene reconstructions.

\subsection{Ablation Study}
\label{sec:ablation}

We study the contribution of each major system component by averaging the ablation over all 8 Replica-SL scenes.
The baseline uses LibSGM depth together with the same Gaussian-mapper-only downstream pipeline, where LibSGM serves as a proxy for the depth quality of traditional commodity structured-light pipelines.
We then replace the baseline depth with \ours{} Depth, which denotes the depth predicted by our proposed neural structured-light decoder, and incrementally add sparse visual correspondences, backend BA, and pose feedback.
All values in Table~\ref{tab:ablation} are first-frame-aligned translation ATE RMSE.

\begin{table}[t]
\caption{Ablation study on Replica-SL averaged over 8 scenes. Lower is better.}
\label{tab:ablation}
\centering
\small
\begin{tabular}{l|c}
\toprule
Configuration & Avg. ATE$\downarrow$ \\
\midrule
LibSGM depth + Gaussian mapper & 1.5460 \\
\ours{} Depth + Gaussian mapper & 0.1529 \\
+ sparse visual correspondences & 0.0280 \\
+ backend BA & \second{0.0188} \\
+ hard feedback & 0.0265 \\
+ soft feedback (full) & \best{0.0155} \\
\bottomrule
\end{tabular}
\end{table}

Table~\ref{tab:ablation} reports the ablation averaged over all 8 Replica-SL scenes.
Replacing LibSGM depth with \ours{} Depth reduces the average ATE from 1.5460\,m to 0.1529\,m, showing that moving beyond commodity-quality structured-light depth is the first requirement for practical SLAM in this setting.
Once that stronger depth foundation is in place, adding sparse visual correspondences further reduces the average ATE to 0.0280\,m, indicating that the dominant residual failures come from geometrically ambiguous frames where dense registration alone is not sufficiently constrained.
Backend BA then reduces the average error to 0.0188\,m, confirming that long-range inconsistency remains even after local tracking becomes much more stable.
Finally, soft feedback outperforms hard feedback by a clear margin (0.0155\,m vs.\ 0.0265\,m), showing that gradual absorption of corrected poses yields a more stable online map than abrupt pose-driven map updates.

\subsection{Runtime Analysis}
\label{sec:runtime}

On an RTX 4090 at $848\times480$, the TensorRT depth estimator runs at 27.5 FPS (31 FPS on RTX 5090), and the full online SLAM system runs at 20.9 FPS. The remaining latency comes mainly from CPU GICP and frontend--backend synchronization; the supplementary material provides the runtime breakdown.

\subsection{Limitations}
\label{sec:limitations}

\ours{} still has several important limitations.
First, geometric ambiguity and visual ambiguity can co-occur.
Large planes, repeated structures, weak texture, or reflective regions can simultaneously weaken dense alignment and sparse visual support, leaving too little information for stable tracking.
Second, scenes dominated by very thin structures or cluttered boundaries remain challenging because residual depth smoothing still affects the final fused geometry.
Third, the backend is intentionally lightweight rather than globally heavy-handed; when the frontend has already drifted far outside the correct basin, the resulting correction range is correspondingly limited.
Fourth, although the ablation provides substantially stronger scene-averaged system-level evidence, the paper still does not provide a separate per-failure-mode causal dissection of every component.
Finally, the real sequences lack ground-truth trajectories and use the T265 only as a reference, and the real reconstructions lack quantitative ground-truth evaluation.

\section{Conclusion}
\label{sec:conclusion}

We have presented \ours{}, a practical neural structured-light SLAM system that advances both per-frame depth quality and downstream reconstruction capability.
By injecting frozen monocular priors directly into the disparity-refinement stage of structured-light decoding, the structured-light depth estimator establishes a stronger depth foundation for the practical SLAM regime targeted in this paper and transfers those gains to SLAM and reconstruction.
Built on that stronger geometry, our system couples depth-guided tracking, an online Gaussian scene representation, and lightweight backend correction into a coherent pipeline rather than treating depth and SLAM as separate problems.
Under the evaluation protocols in this paper, the resulting system shows a strong overall tradeoff on Replica-SL and the most stable behavior on real D435 captures relative to the T265 reference.
More broadly, the results support the paper's main claim: once neural structured-light depth becomes strong enough, practical SLAM with robust downstream reconstruction is possible with a simple, explicit system design that preserves dense depth as the main signal and addresses the remaining failure cases instead of hiding them inside heavier optimization.

\begin{acks}
This work is supported by the projects of Beijing Science and Technology Program
(Z251100008125028).
\end{acks}

\bibliographystyle{unsrt}
\bibliography{references}

\clearpage
\appendix
\section*{Supplementary Material}

\section{Implementation Details}
\label{sec:implementation_details}

\subsection{Depth Estimation Details}

Our structured-light depth estimator is built on RAFT-Stereo~\cite{RAFTStereo} and incorporates a frozen Depth Anything v2 ViT-L prior~\cite{DepthAnythingV2} through feature concatenation in the context branch. The model is trained on the NSL\_train dataset~\cite{NSL} with 16 recurrent update iterations, and the same number of iterations is used at inference time. Training uses AdamW with a learning rate of $6\times10^{-5}$ for 200k optimization steps and a total batch size of 32.

\subsection{Streaming SLAM Pipeline}

The streaming pipeline takes RGB images and predicted depth maps at a default depth resolution of $848\times480$. Frame-to-model GICP registration is augmented with sparse visual anchors from a TensorRT-accelerated LightGlue matcher. Tracking keyframes are inserted when overlap drops below 0.8, additional mapping keyframes every 10 frames, and the tracking target retains the latest 30 keyframes. Sparse matching retains up to 2048 keypoints and requires at least 5 matches. The adaptive $\alpha$ is clipped to $[10^{-3},10^{3}]$.

\subsection{Online Gaussian Mapping}

An asynchronous Gaussian mapper maintains the online scene representation and is optimized with Adam using separate parameter groups for Gaussian position, appearance, opacity, scale, and rotation. In the default setting, the corresponding learning rates are $1.6\times10^{-6}$ for position, $2.5\times10^{-3}$ for appearance features, $5\times10^{-2}$ for opacity, $5\times10^{-3}$ for scale, and $1\times10^{-3}$ for rotation.

On receiving backend-corrected poses, the mapper performs 5 feedback steps on the updated keyframes. The position learning rate increases to $5\times10^{-3}$, while appearance, opacity, scale, and rotation rates decrease to $2.5\times10^{-5}$, $5\times10^{-4}$, $5\times10^{-5}$, and $1\times10^{-5}$, respectively, emphasizing geometric adjustment to the corrected poses. Depth supervision weights are 1.0 for backend-updated keyframes and 0.1 for newly inserted or randomly sampled keyframes.

A separate backend performs bundle adjustment for long-range consistency using LightGlue-TRT correspondences and an active window of 20 keyframes. Backend keyframes have a minimum interval of 5 frames and relative motion thresholds of 0.08\,m in translation and $20^\circ$ in rotation, distinct from the 10-frame mapping interval. Final reconstruction combines the corrected trajectory with per-frame dense depth reprojection.

\subsection{Hyperparameter Sensitivity}
\label{supp:hyperparameter_sensitivity}

Table~\ref{tab:supp_hyperparameter_sensitivity} reports a one-factor sweep around the defaults above ($\alpha$ follows Eq.~5); multipliers such as $0.5\times$ scale only the indicated parameter while keeping others fixed. Most hyperparameters show limited sensitivity, and the default settings consistently perform among the top-performing settings across the sweep.

\begin{table}[ht]
\caption{One-factor sensitivity. ATE is averaged over room0 and office0 (cm); all non-varied parameters stay at default.}
\label{tab:supp_hyperparameter_sensitivity}
\centering
\small
\begin{tabular}{l|cc|cc|cc}
\toprule
\multirow{2}{*}{Sweep} & \multicolumn{2}{c|}{Default} & \multicolumn{2}{c|}{Var. 1} & \multicolumn{2}{c}{Var. 2} \\
& setting & ATE & setting & ATE & setting & ATE \\
\midrule
$\alpha$ & $1\times$ & 1.87 & $0.5\times$ & 1.91 & $2\times$ & 2.11 \\
Window & 20 & 1.87 & 10 & 1.97 & 30 & 1.91 \\
KF int. & 5 & 1.87 & 3 & 1.82 & 8 & 2.03 \\
Motion & $1\times$ & 1.87 & $0.5\times$ & 1.88 & $1.5\times$ & 1.75 \\
\bottomrule
\end{tabular}
\end{table}

\subsection{Runtime Breakdown}
\label{supp:runtime_breakdown}

\begin{figure}[t]
\centering
\includegraphics[width=\linewidth]{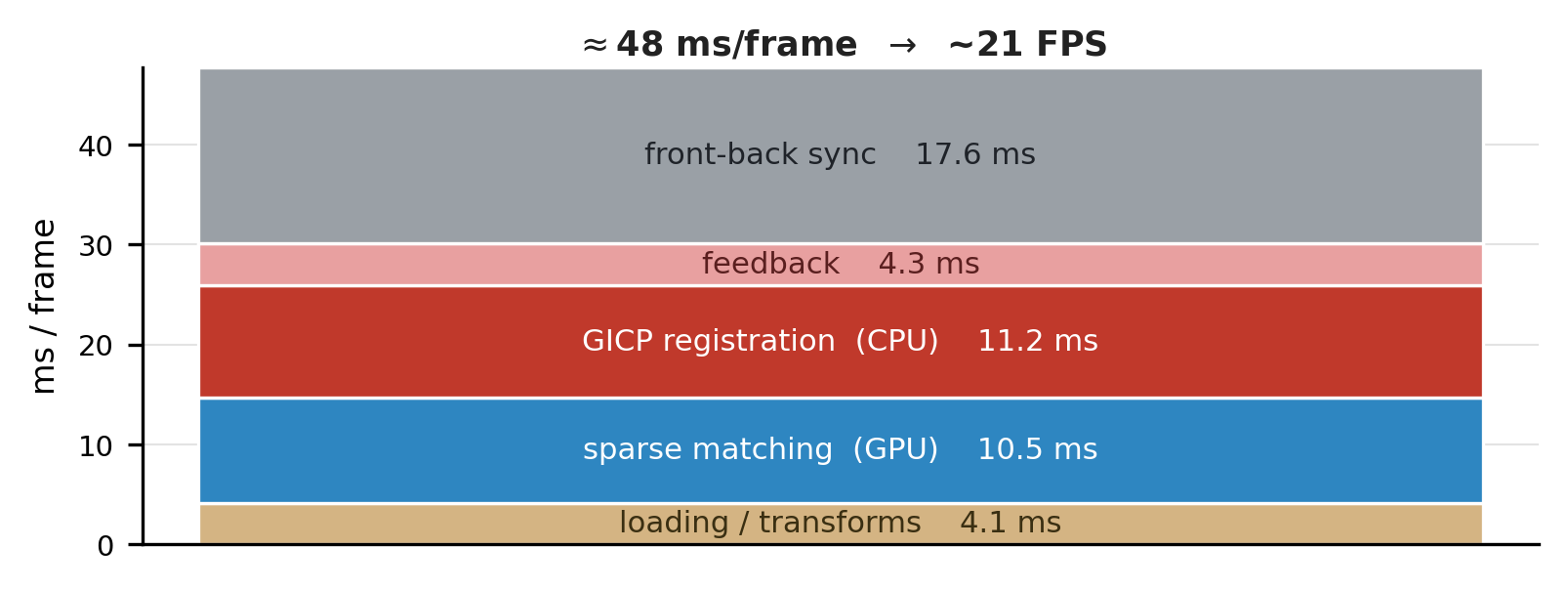}
\caption{Runtime breakdown on an RTX 4090 at $848\times480$. TensorRT depth runs at 27.5 FPS and the full online SLAM system runs at 20.9 FPS.}
\label{fig:supp_runtime_breakdown_rtx4090}
\end{figure}

On an RTX 4090 at $848\times480$, TensorRT depth runs at 27.5 FPS and the online SLAM system runs at 20.9 FPS. The remaining latency comes mainly from CPU GICP and frontend--backend synchronization, as detailed in Fig.~\ref{fig:supp_runtime_breakdown_rtx4090}.

\begin{figure*}[!t]
\centering
\setlength{\tabcolsep}{1pt}
\renewcommand{\arraystretch}{0.4}
\begin{tabular}{c c c c c c}

\includegraphics[width=0.145\textwidth]{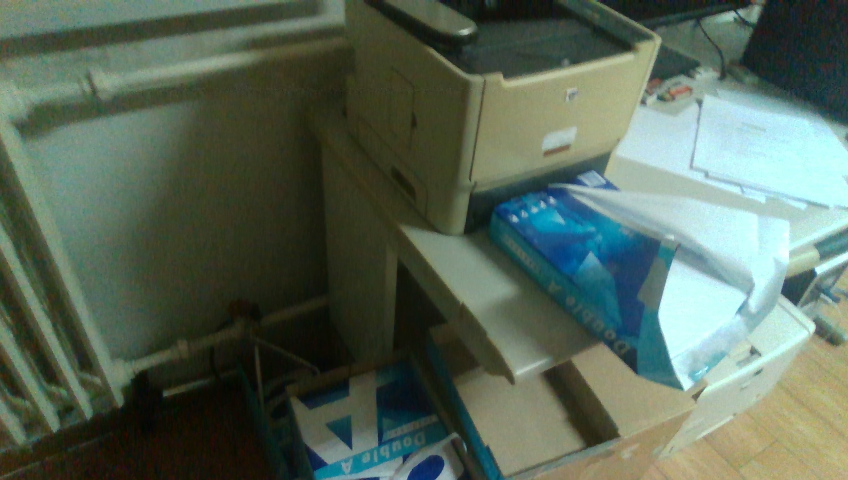} &
\includegraphics[width=0.145\textwidth]{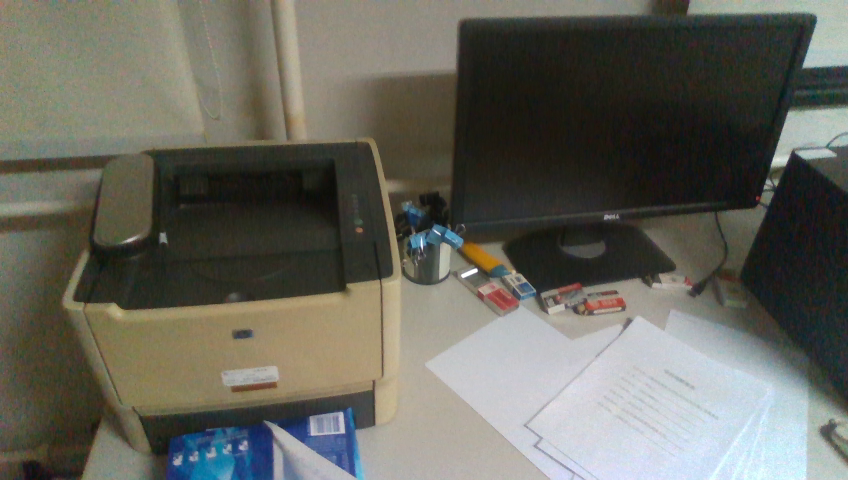} &
\includegraphics[width=0.145\textwidth]{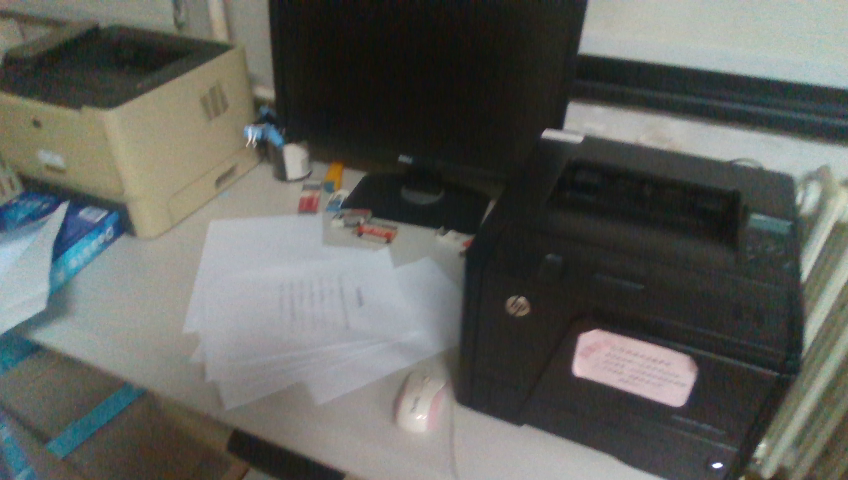} &
\includegraphics[width=0.145\textwidth]{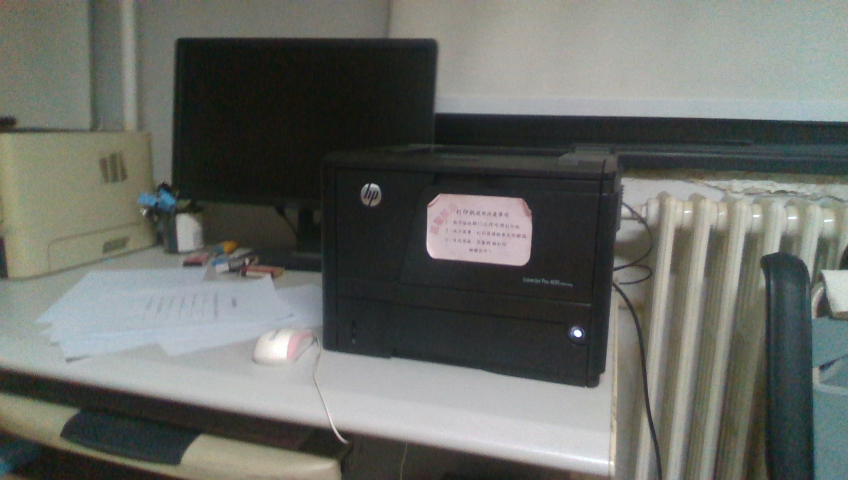} &
\includegraphics[width=0.145\textwidth]{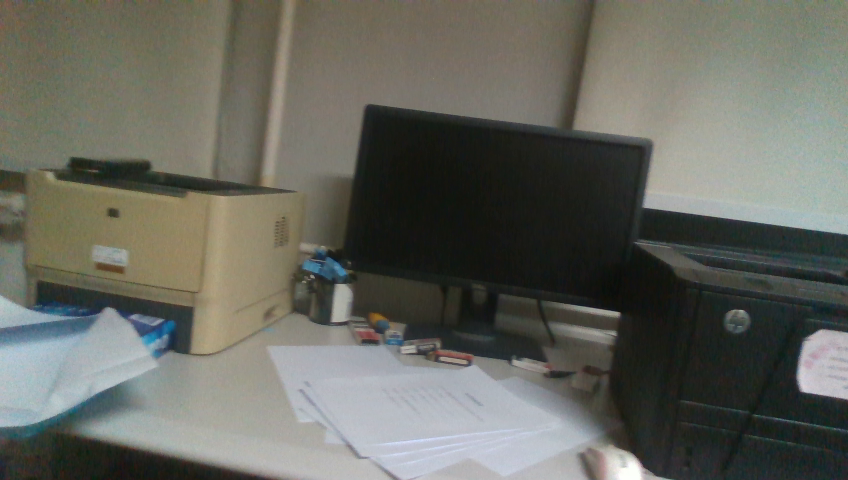} &
\includegraphics[width=0.145\textwidth]{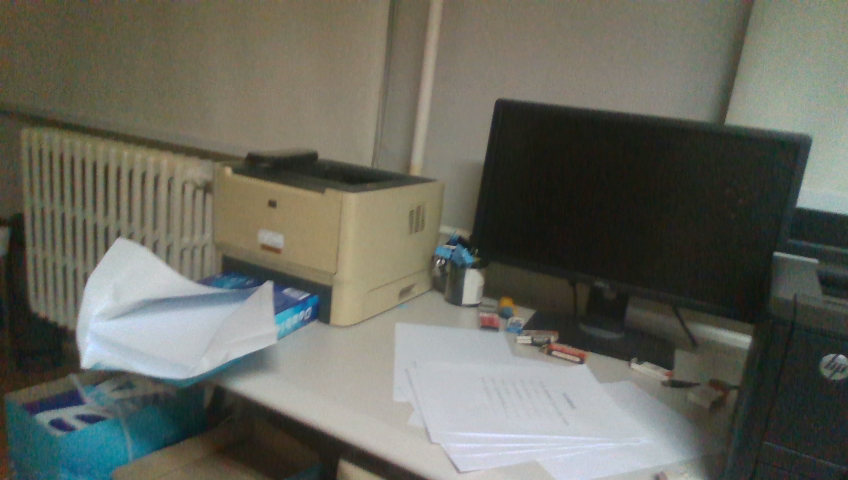} \\
\includegraphics[width=0.145\textwidth]{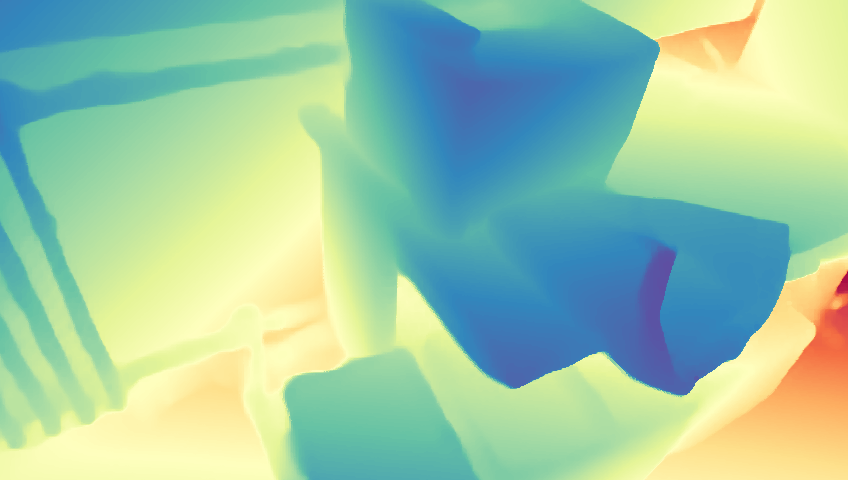} &
\includegraphics[width=0.145\textwidth]{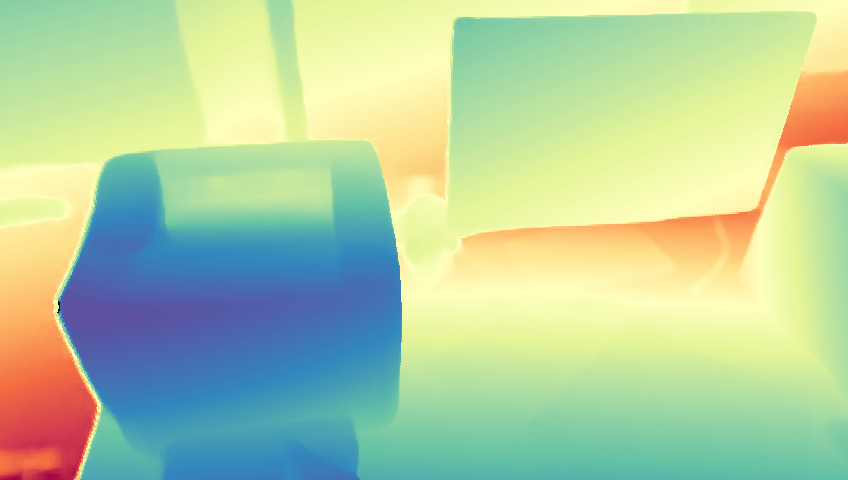} &
\includegraphics[width=0.145\textwidth]{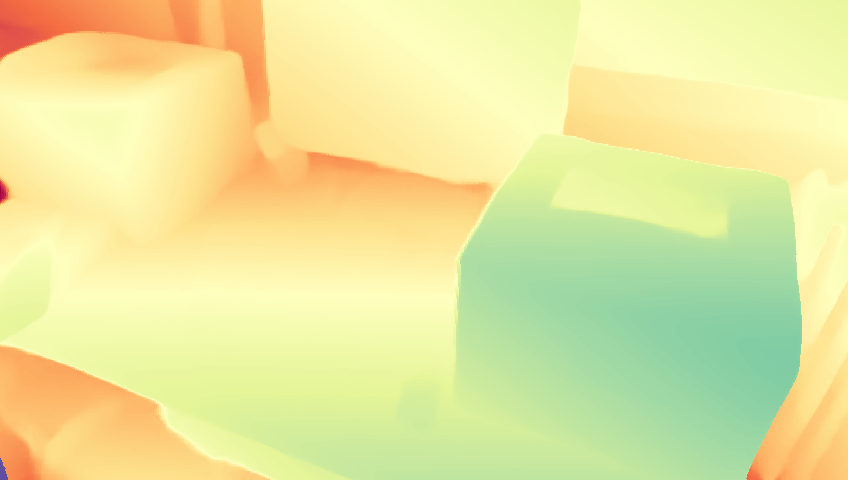} &
\includegraphics[width=0.145\textwidth]{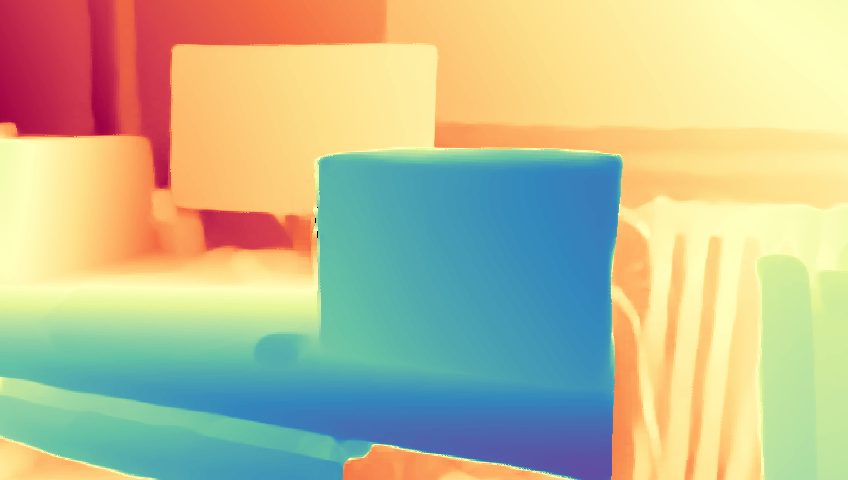} &
\includegraphics[width=0.145\textwidth]{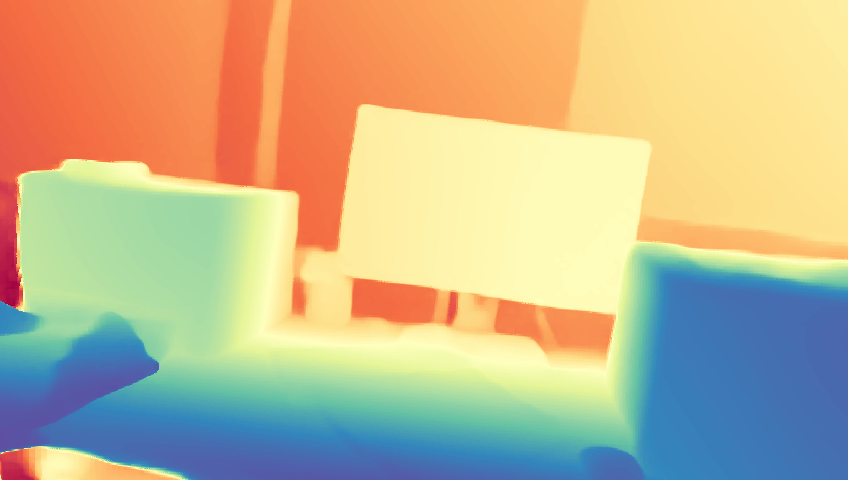} &
\includegraphics[width=0.145\textwidth]{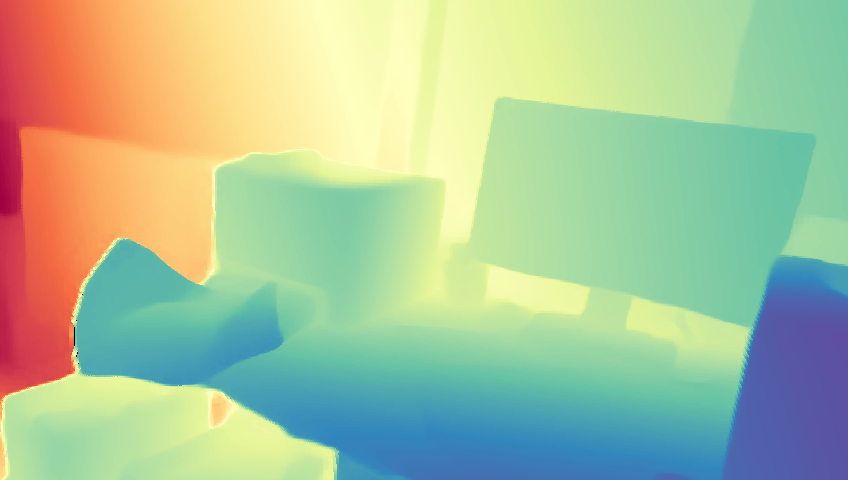} \\

\includegraphics[width=0.145\textwidth]{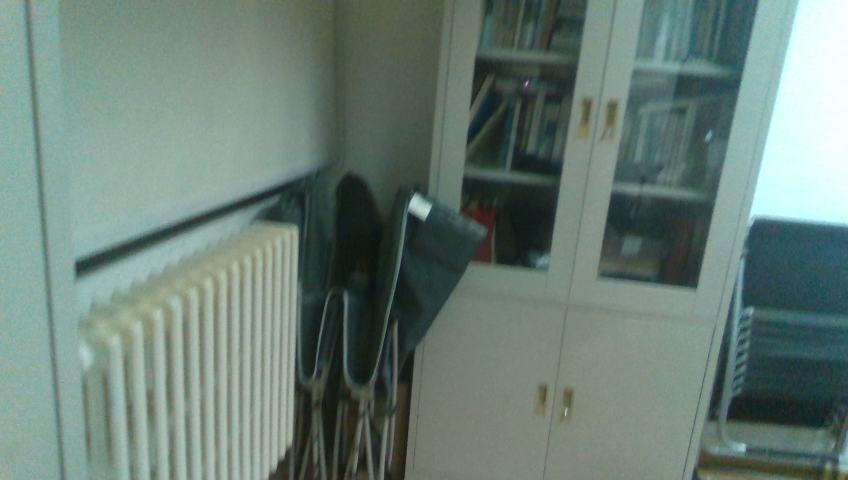} &
\includegraphics[width=0.145\textwidth]{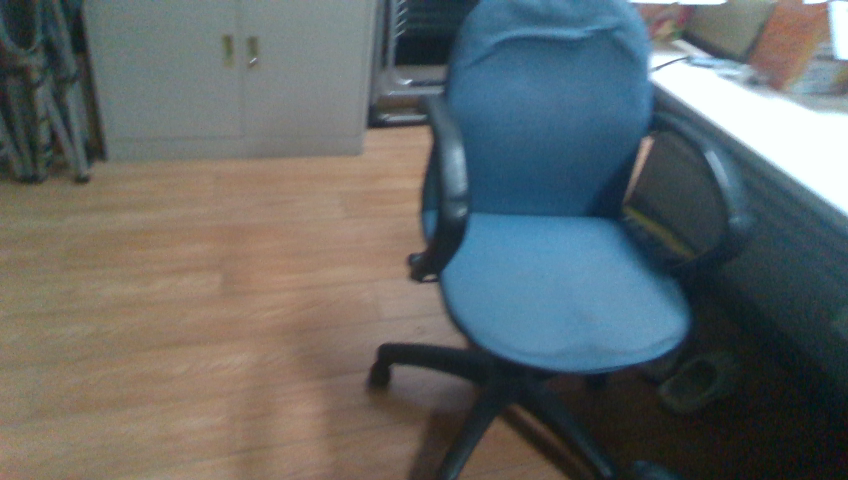} &
\includegraphics[width=0.145\textwidth]{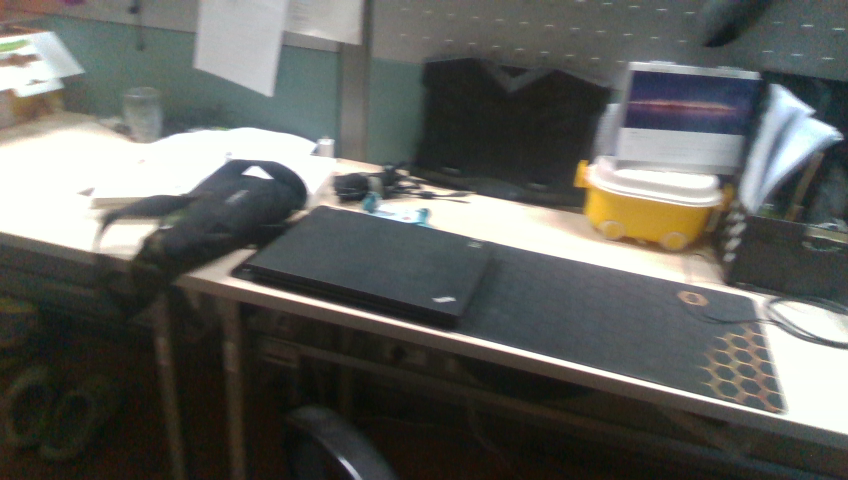} &
\includegraphics[width=0.145\textwidth]{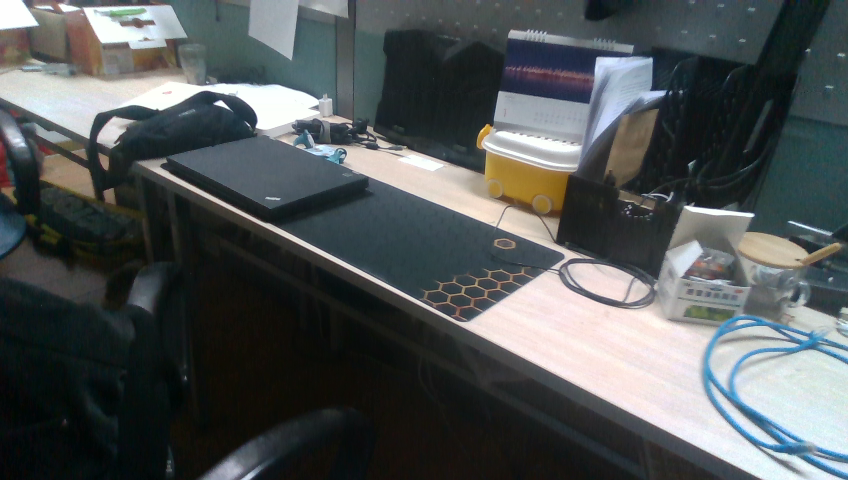} &
\includegraphics[width=0.145\textwidth]{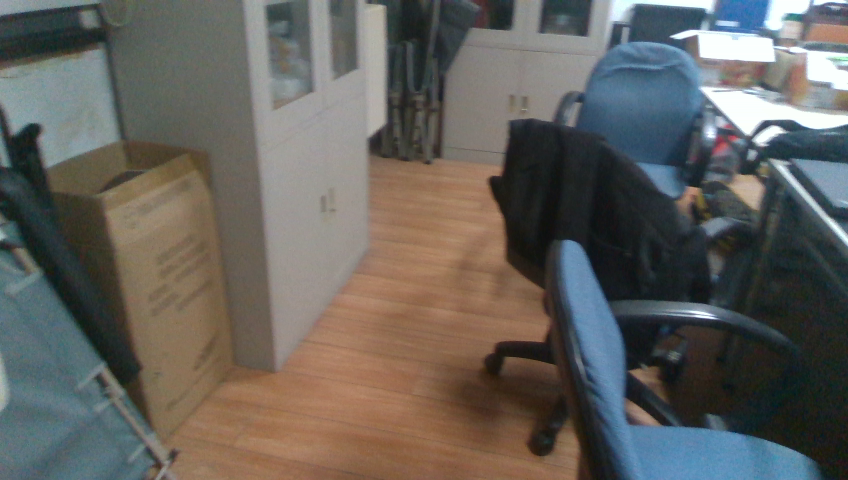} &
\includegraphics[width=0.145\textwidth]{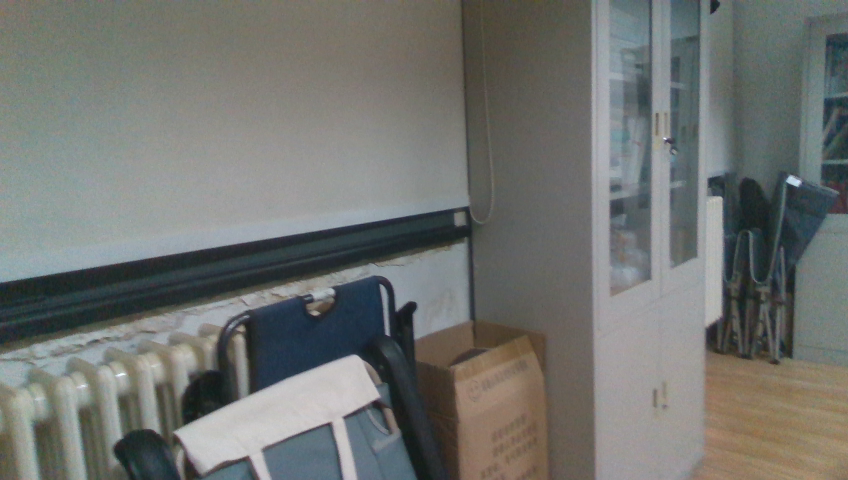} \\
\includegraphics[width=0.145\textwidth]{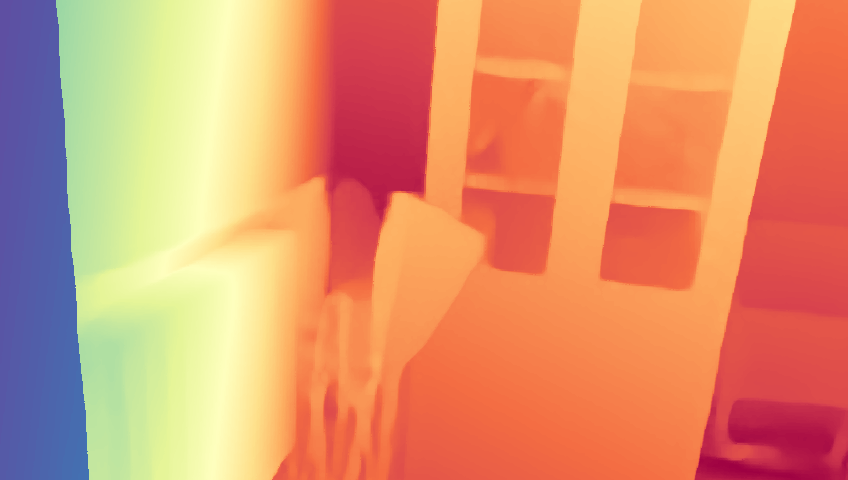} &
\includegraphics[width=0.145\textwidth]{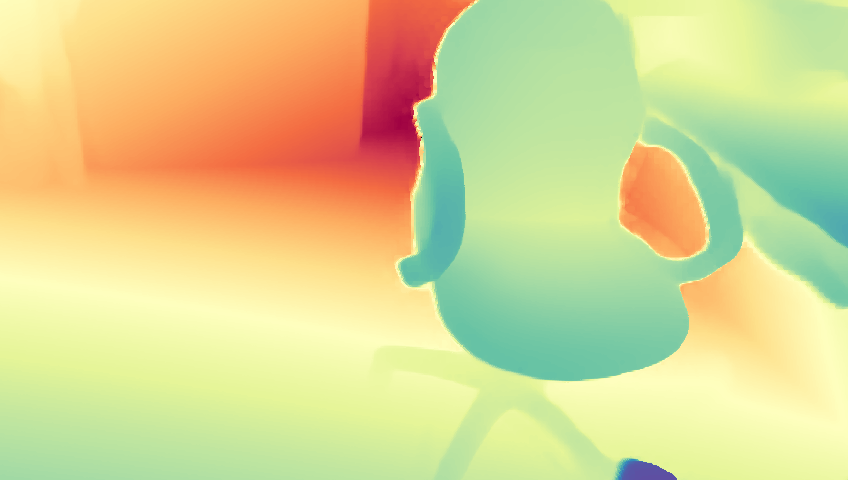} &
\includegraphics[width=0.145\textwidth]{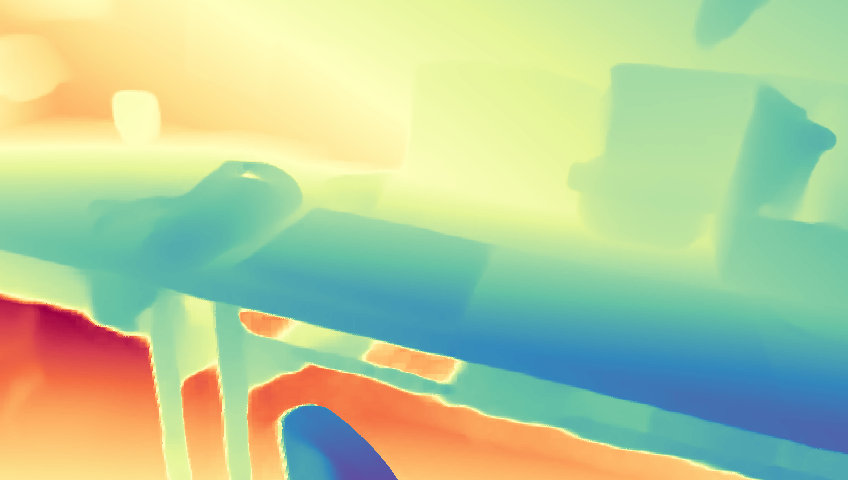} &
\includegraphics[width=0.145\textwidth]{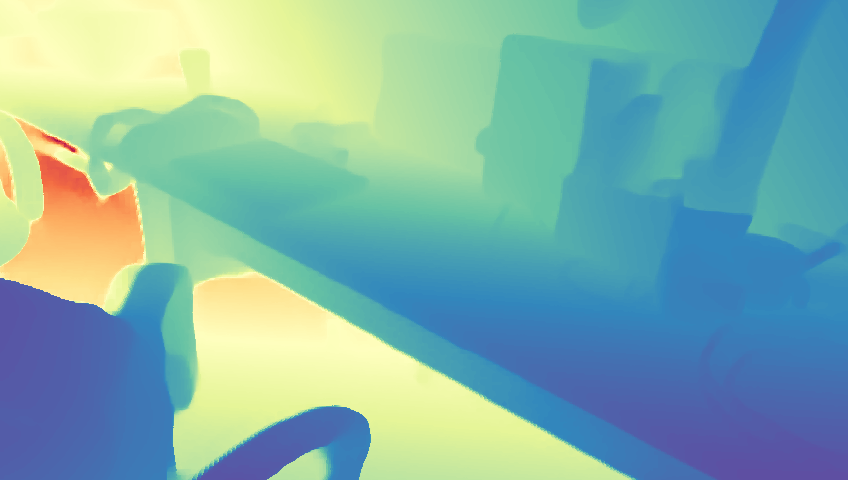} &
\includegraphics[width=0.145\textwidth]{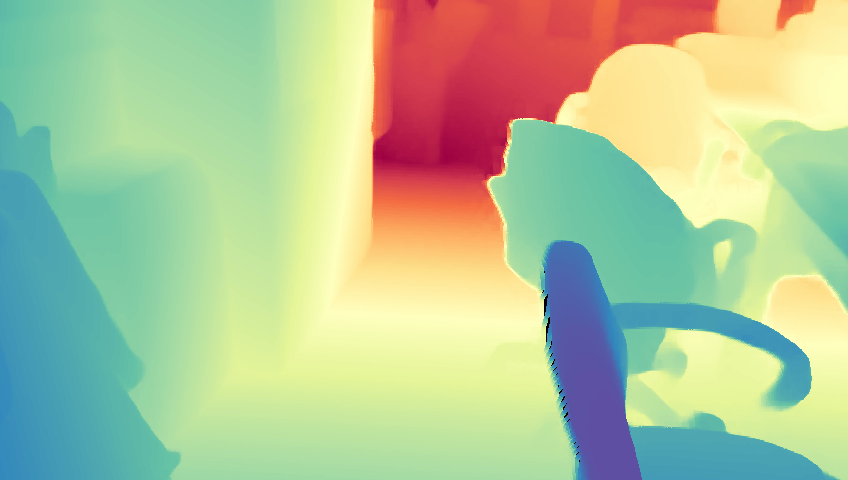} &
\includegraphics[width=0.145\textwidth]{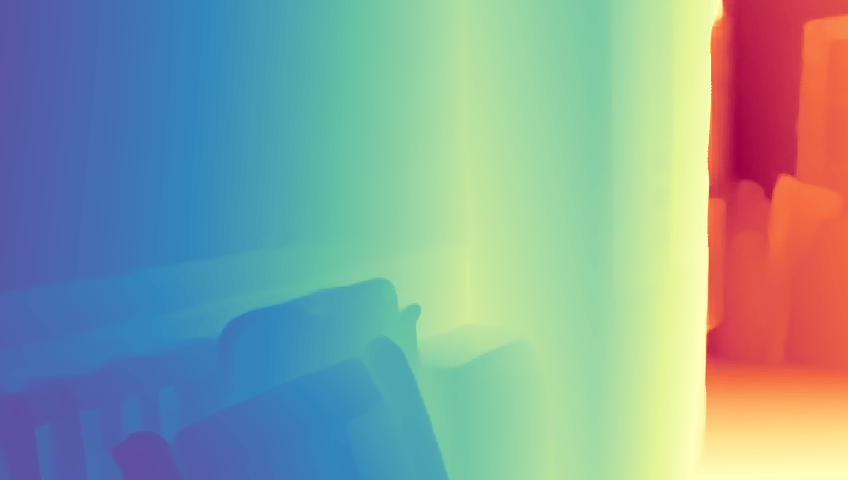} \\

\includegraphics[width=0.145\textwidth]{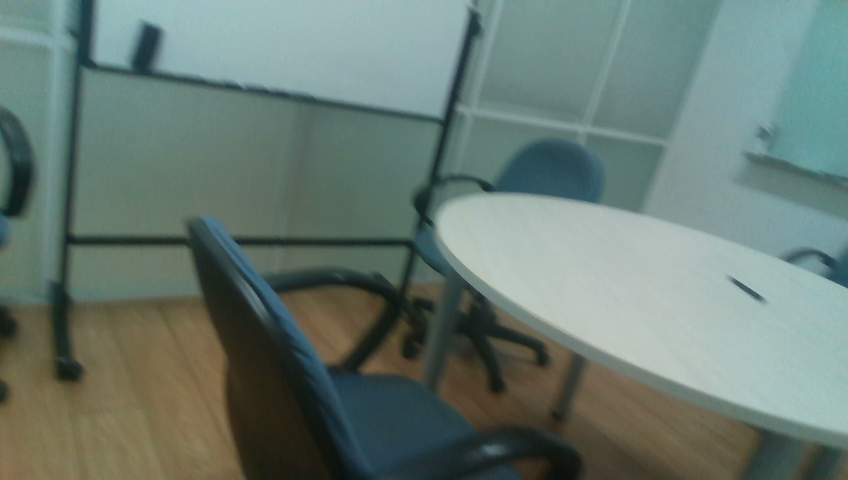} &
\includegraphics[width=0.145\textwidth]{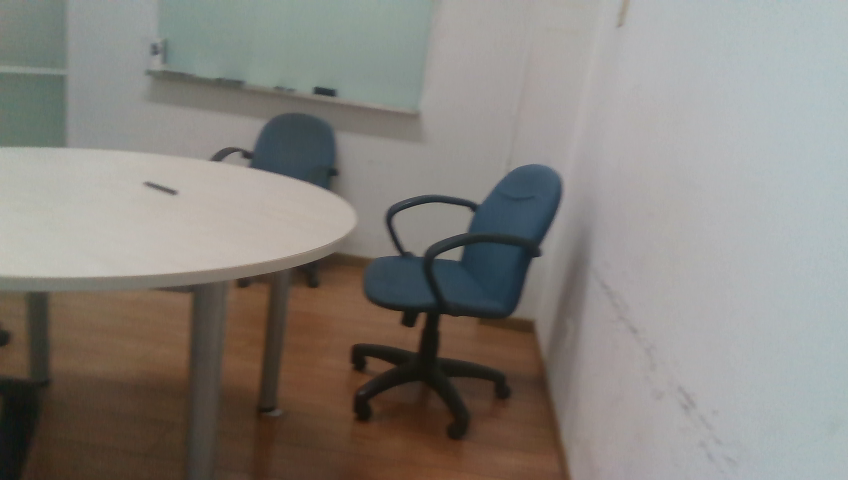} &
\includegraphics[width=0.145\textwidth]{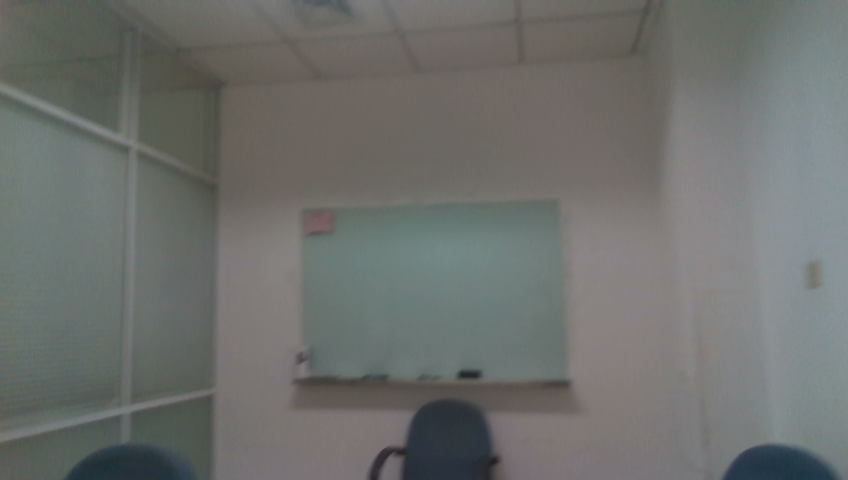} &
\includegraphics[width=0.145\textwidth]{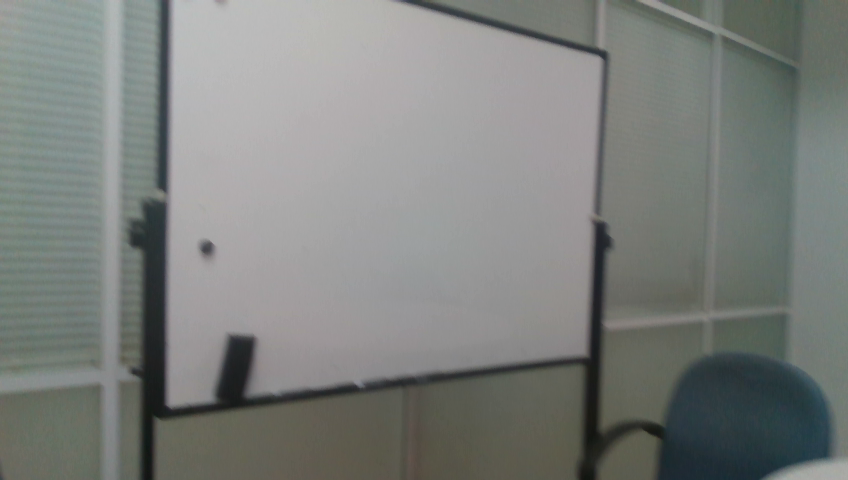} &
\includegraphics[width=0.145\textwidth]{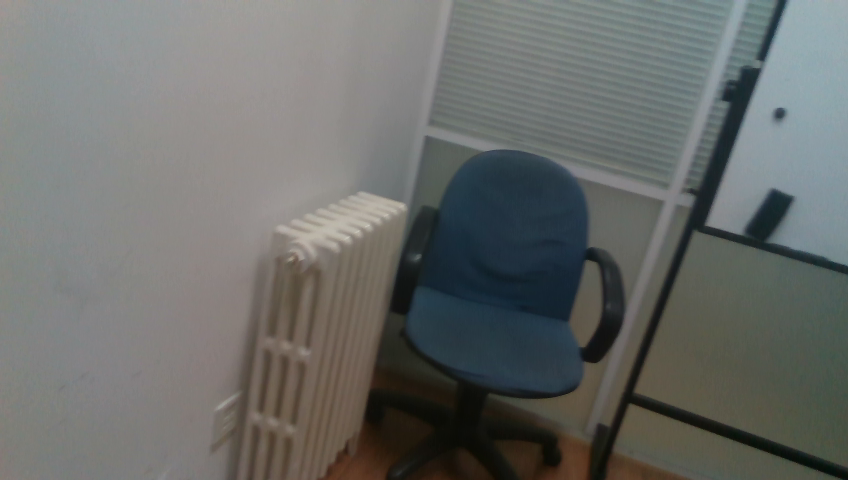} &
\includegraphics[width=0.145\textwidth]{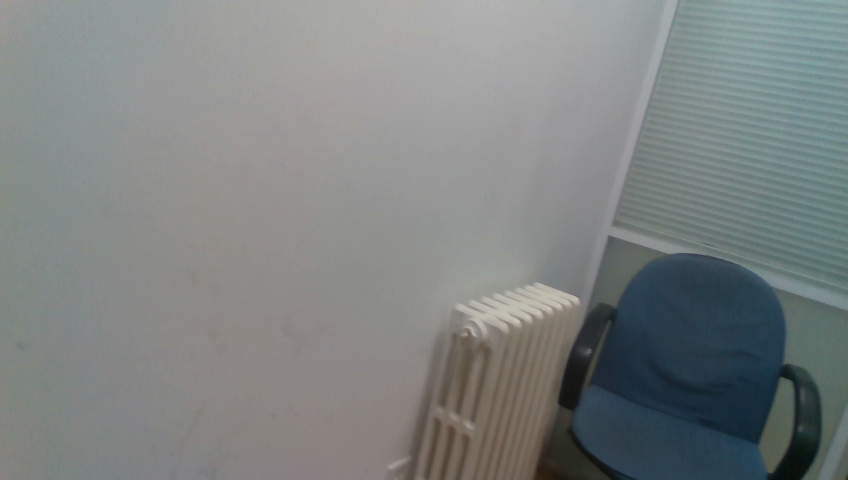} \\
\includegraphics[width=0.145\textwidth]{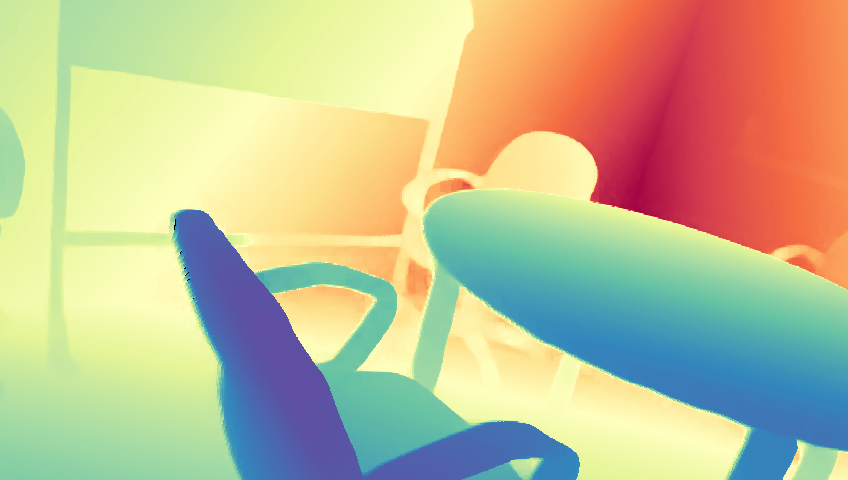} &
\includegraphics[width=0.145\textwidth]{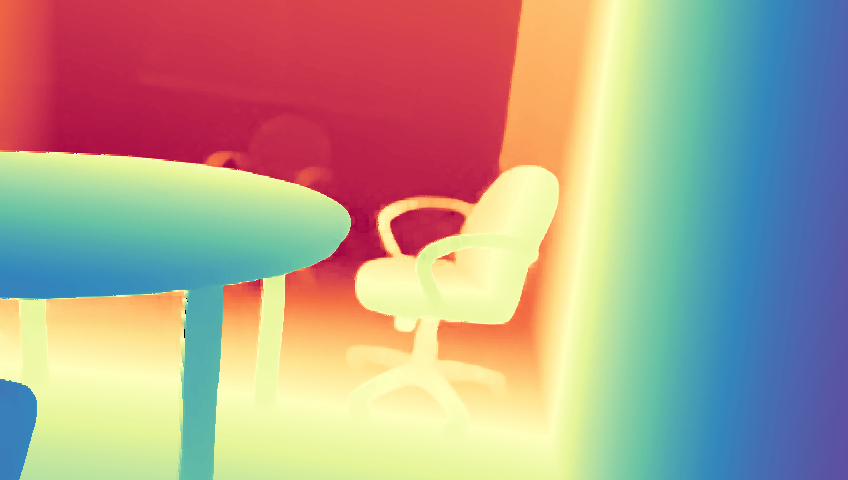} &
\includegraphics[width=0.145\textwidth]{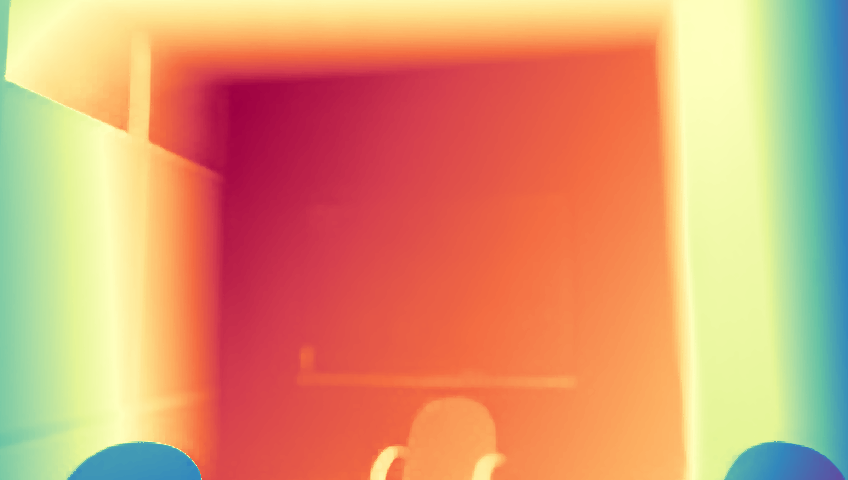} &
\includegraphics[width=0.145\textwidth]{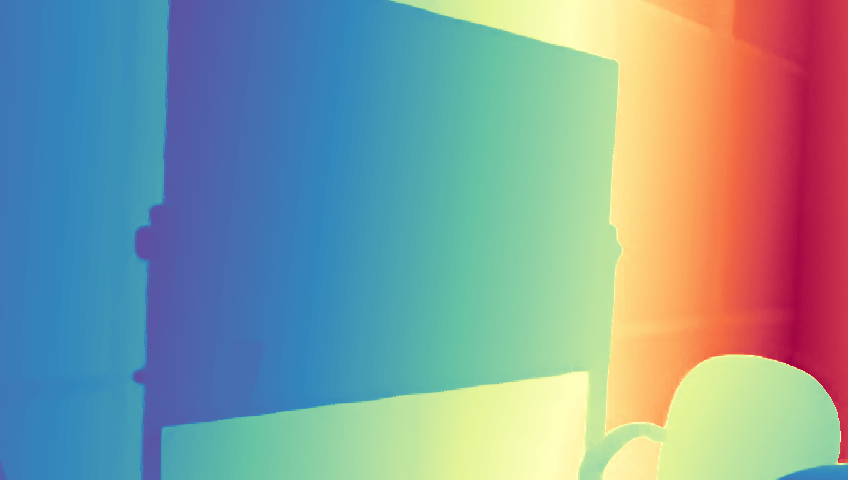} &
\includegraphics[width=0.145\textwidth]{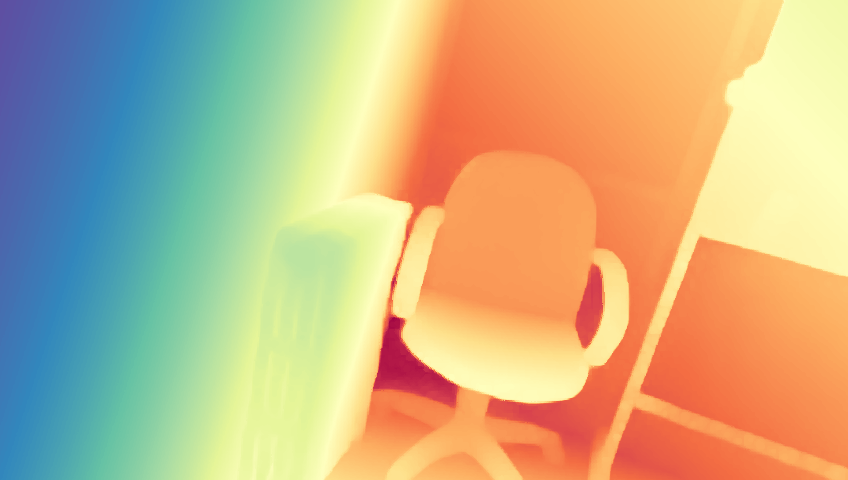} &
\includegraphics[width=0.145\textwidth]{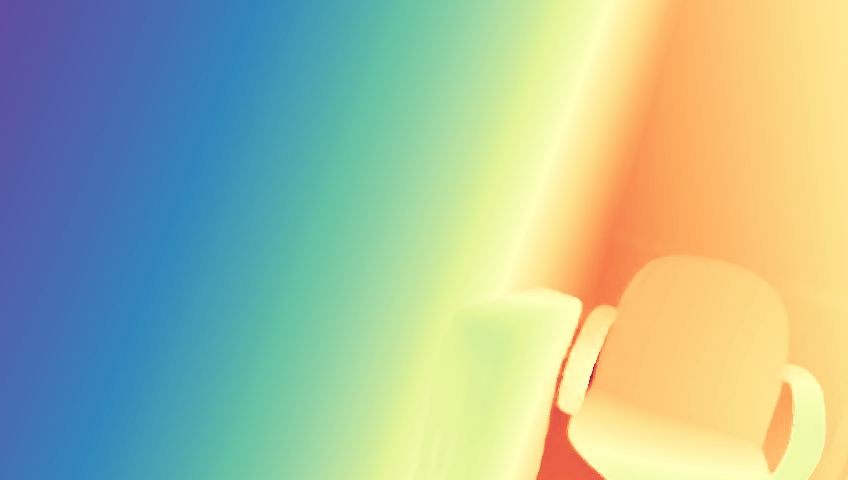} \\

\includegraphics[width=0.145\textwidth]{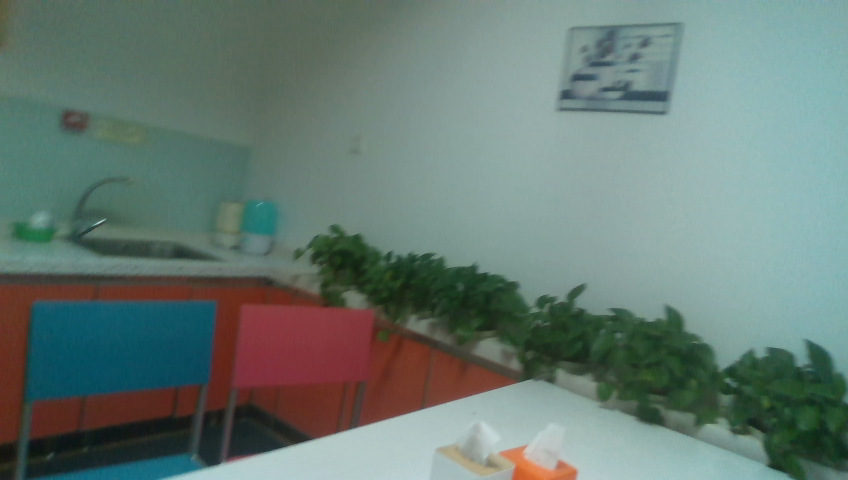} &
\includegraphics[width=0.145\textwidth]{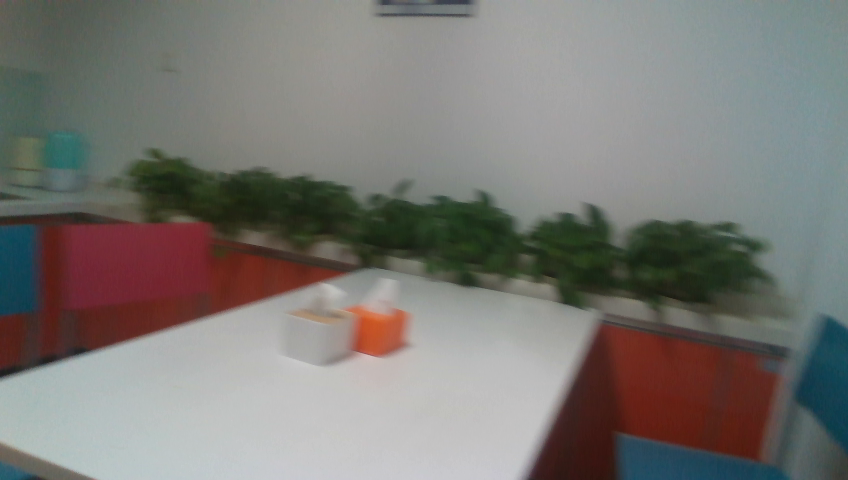} &
\includegraphics[width=0.145\textwidth]{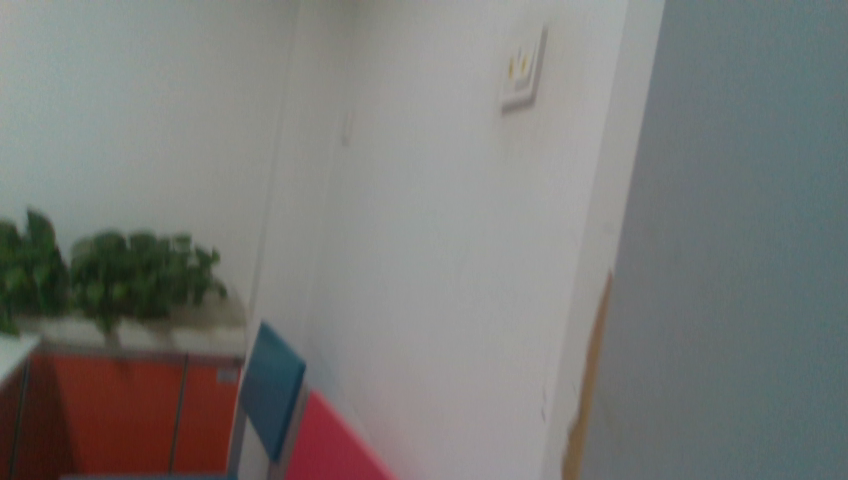} &
\includegraphics[width=0.145\textwidth]{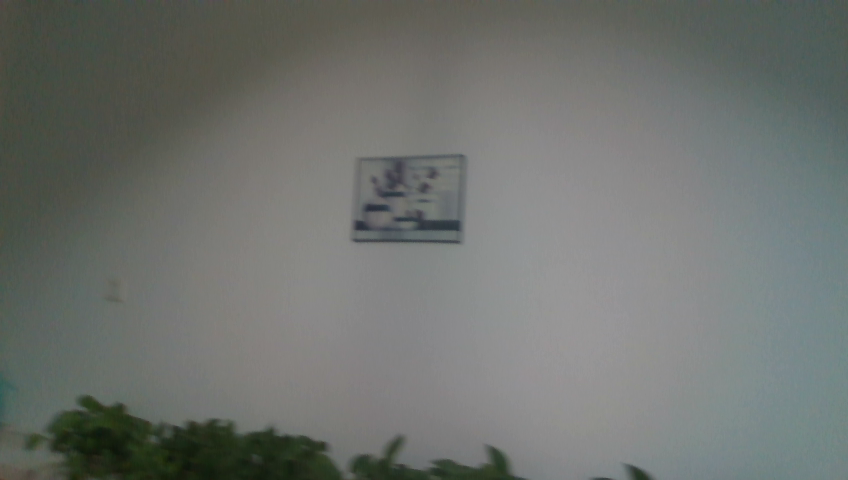} &
\includegraphics[width=0.145\textwidth]{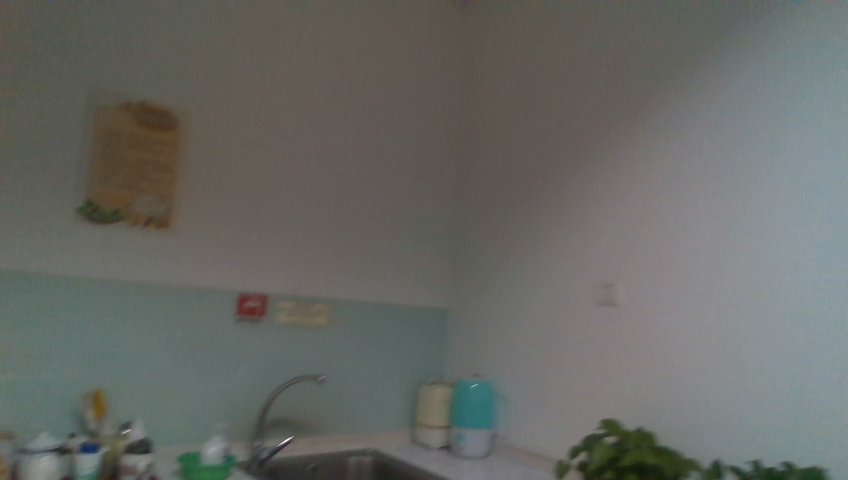} &
\includegraphics[width=0.145\textwidth]{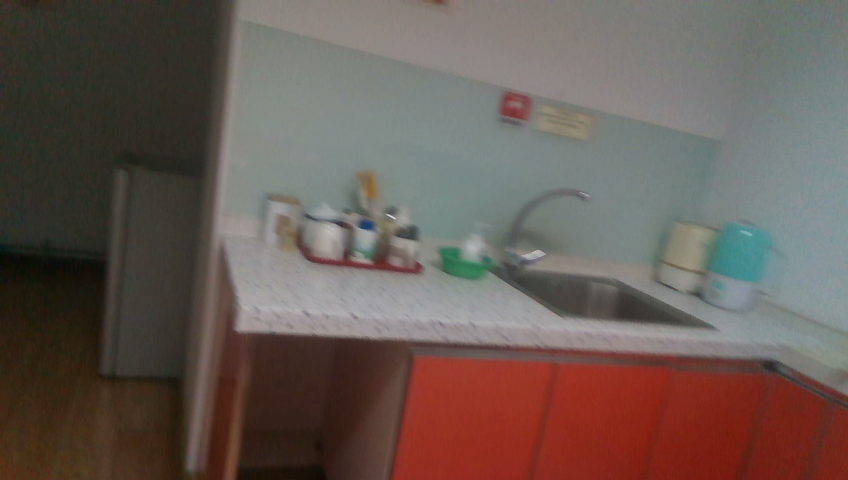} \\
\includegraphics[width=0.145\textwidth]{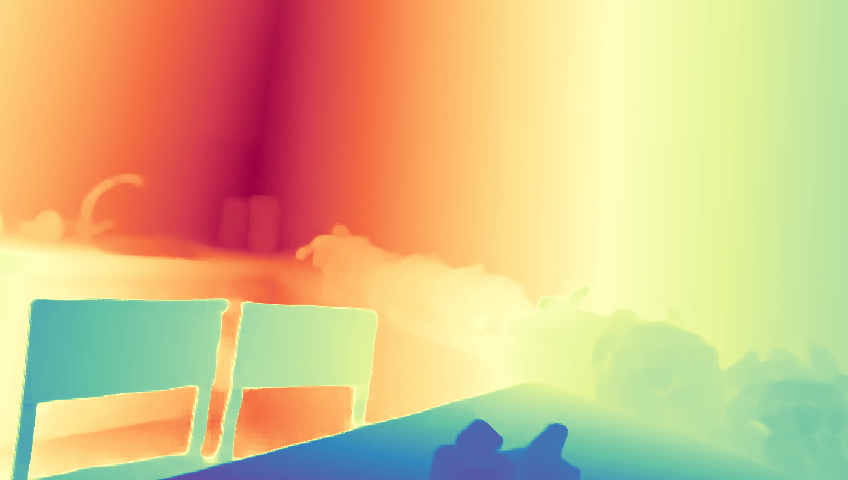} &
\includegraphics[width=0.145\textwidth]{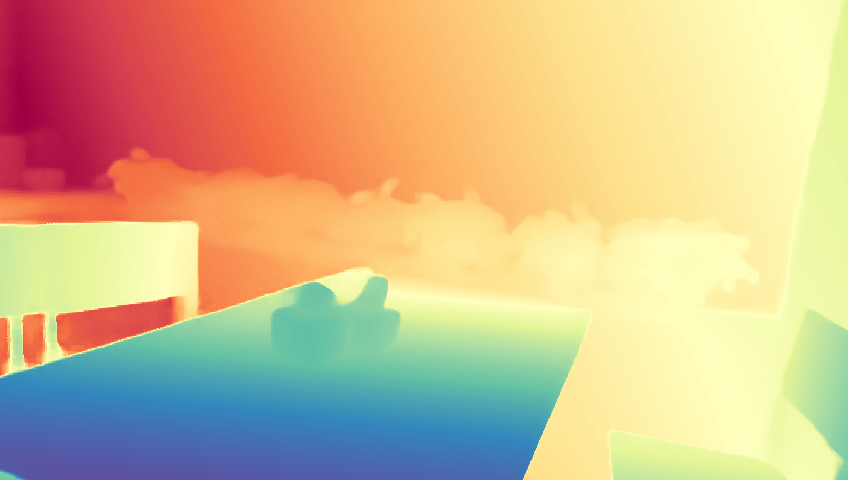} &
\includegraphics[width=0.145\textwidth]{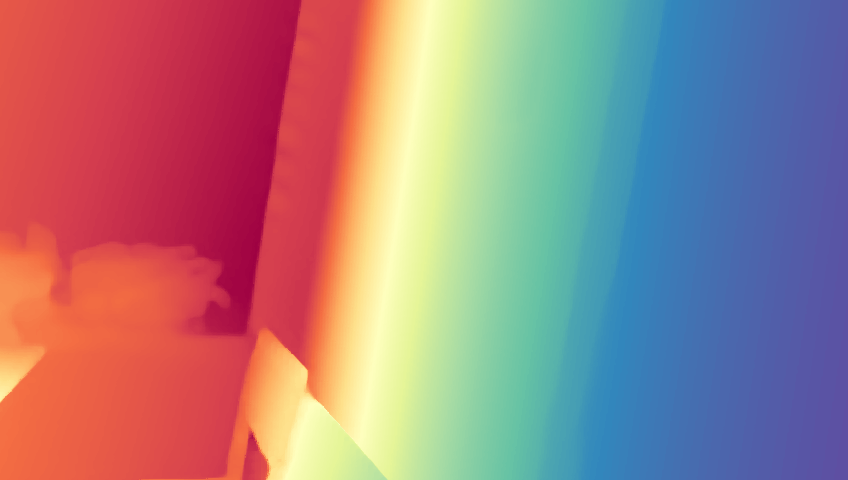} &
\includegraphics[width=0.145\textwidth]{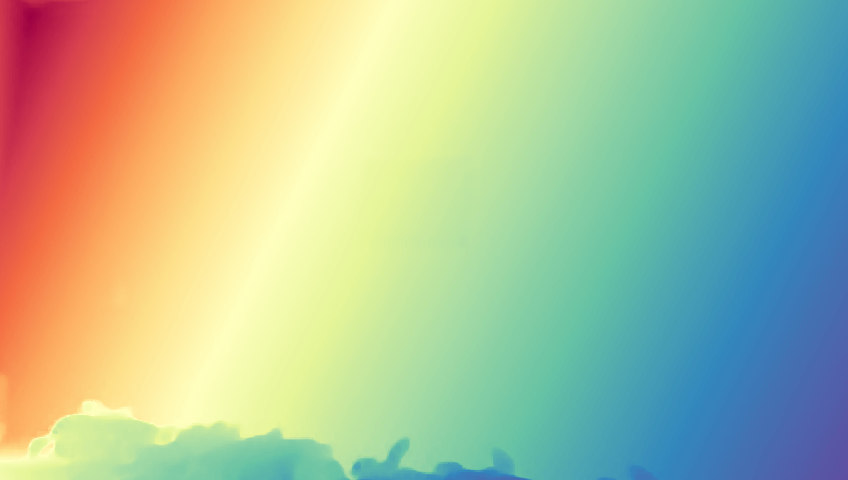} &
\includegraphics[width=0.145\textwidth]{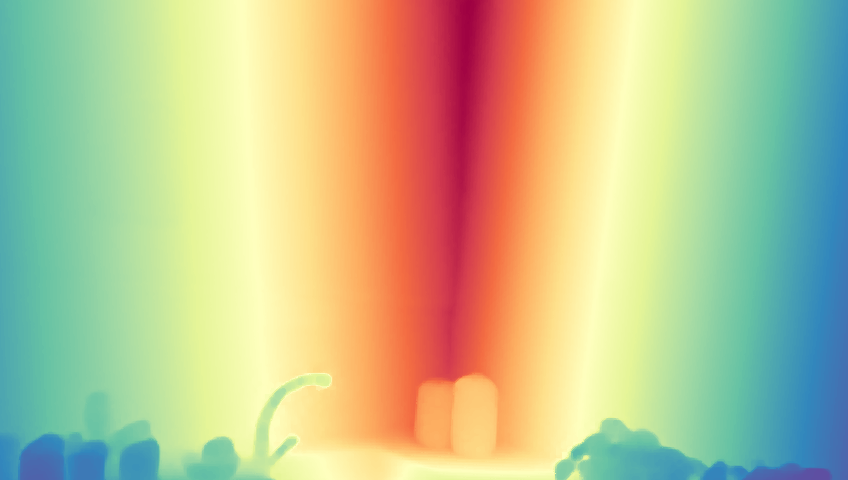} &
\includegraphics[width=0.145\textwidth]{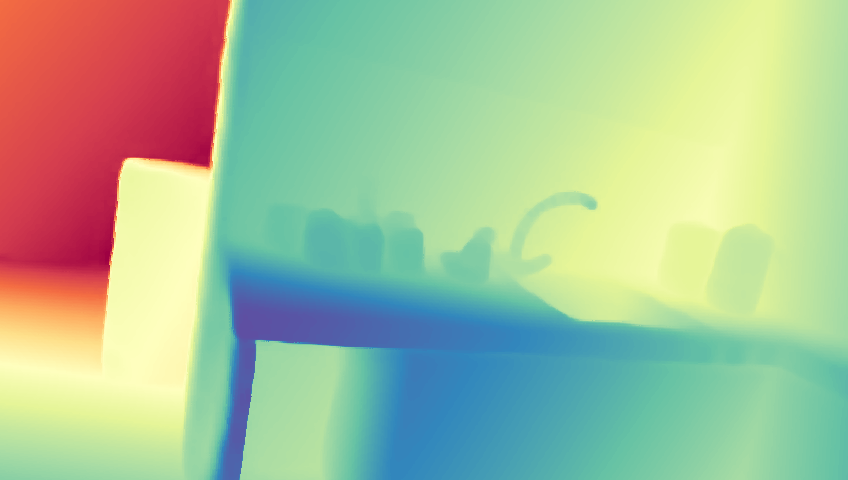} \\
\end{tabular}
\caption{Additional real-scene depth examples on four D435 sequences. The top row of each scene shows the RGB input, and the bottom row shows the corresponding predicted depth.}
\label{fig:more_depth_qual}
\end{figure*}

\section{Replica-SL Synthesis}
\label{sec:replica_sl_synthesis}

To construct the synthetic structured-light version of Replica~\cite{Replica}, we preserve the original camera trajectories and depth maps and synthesize a projected-pattern observation for each frame from the rendered RGB image and ground-truth depth. Let $\mathbf{K}_c$ and $\mathbf{K}_p$ denote the camera and projector intrinsics, and let $\mathbf{T}_{c\rightarrow p}$ denote the camera-to-projector transform. For a camera pixel $\mathbf{u}=(u,v,1)^\top$ with depth $d(\mathbf{u})$, the corresponding 3D point in the camera frame is first recovered as
\begin{equation}
\mathbf{x}_c(\mathbf{u}) = d(\mathbf{u})\,\mathbf{K}_c^{-1}\mathbf{u},
\end{equation}
and then transformed into the projector frame,
\begin{equation}
\mathbf{x}_p(\mathbf{u}) = \mathbf{T}_{c\rightarrow p}
\begin{bmatrix}
\mathbf{x}_c(\mathbf{u})\\
1
\end{bmatrix}.
\end{equation}
The projector image coordinate is obtained by perspective projection,
\begin{equation}
\mathbf{\pi}_p(\mathbf{u}) = \Pi\!\left(\mathbf{K}_p\,\mathbf{x}_p(\mathbf{u})\right),
\end{equation}
where $\Pi(\cdot)$ denotes homogeneous normalization. The projected pattern is then sampled at $\mathbf{\pi}_p(\mathbf{u})$ by bilinear interpolation.

The synthesized structured-light image is generated with a simple image-formation model,
\begin{equation}
\tilde{I}(\mathbf{u}) =
\operatorname{clip}\!\left(
\gamma \left[
\alpha I(\mathbf{u}) +
\beta\,P\!\left(\mathbf{\pi}_p(\mathbf{u})\right)\,
R(\mathbf{u})\,
\frac{1}{1+\eta z_p(\mathbf{u})^2}
\right]
\right),
\end{equation}
where $I(\mathbf{u})$ is the original Replica RGB intensity, $P(\cdot)$ is the projected pattern image, $R(\mathbf{u})$ is an estimated surface reflectance term, $z_p(\mathbf{u})$ is the depth of the point in the projector frame, and $\operatorname{clip}(\cdot)$ truncates the result to $[0,1]$. Here $\alpha$ controls the ambient component, $\beta$ controls the projector contribution, $\gamma$ is a global exposure factor, and $\eta$ is the distance-attenuation coefficient. The camera-projector baseline is 8\,cm with zero relative rotation.

The reflectance map $R$ is estimated from the input RGB image with a Retinex-style decomposition on luminance, in which illumination is obtained by Gaussian smoothing in the log-intensity domain and reflectance is recovered by division. To keep the projected pattern visible without excessive saturation, the projector strength $\beta$ is automatically calibrated on a reference frame by enforcing low saturation and sufficient projected-pattern contrast; the calibrated value is then used for the whole dataset.
\par\medskip
\section{Low-Cost Sensing Hardware}
\label{supp:sensing_hardware}

Although tested with RealSense and Orbbec cameras, our algorithm does not require these relatively costly devices. An alternative below CNY~200 combines an stereo camera with a self-assembled VCSEL--DOE projector and collimating lenses. Table~\ref{tab:supp_sensing_hardware} estimates the materials at CNY~197 using low-end component budgets.

\noindent\begin{minipage}{\linewidth}
\captionof{table}{Estimated material costs (CNY).}
\label{tab:supp_sensing_hardware}
\centering
\small
\setlength{\tabcolsep}{3pt}
\renewcommand{\arraystretch}{1.0}
\begin{tabular}{@{}p{0.20\linewidth}p{0.65\linewidth}r@{}}
\toprule
Component & Brand / model & CNY \\
\midrule
Stereo camera &
HBVCAM HBV-1780-2 OV9732 Full-spectrum & 150 \\
VCSEL &
Silverlight G3535N1VIRB3U12SX-850 & 7 \\
DOE+Lenses & Generic dot-pattern DOE \& collimating lenses & 25 \\
Integration & Shell, wiring, and mount & 15 \\
\midrule
\multicolumn{2}{l}{\textbf{Total}} & \textbf{197} \\
\bottomrule
\end{tabular}
\par
\end{minipage}
\par

\section{More qualitative results}
\label{supp:more_ret}

\begin{figure*}[t]
\centering
\setlength{\tabcolsep}{3pt}
\renewcommand{\arraystretch}{1.0}
\begin{tabular}{c c c c}
D435+T265: View1 & \textbf{Ours: View1} & D435+T265: View2 & \textbf{Ours: View2} \\
\includegraphics[width=0.20\textwidth]{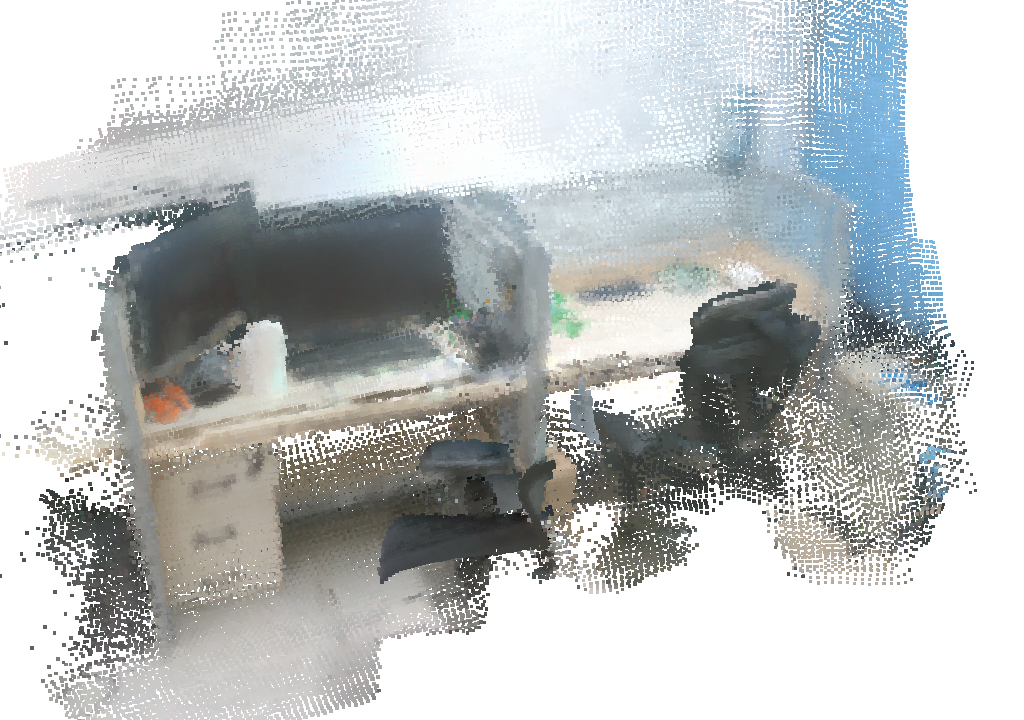} &
\includegraphics[width=0.20\textwidth]{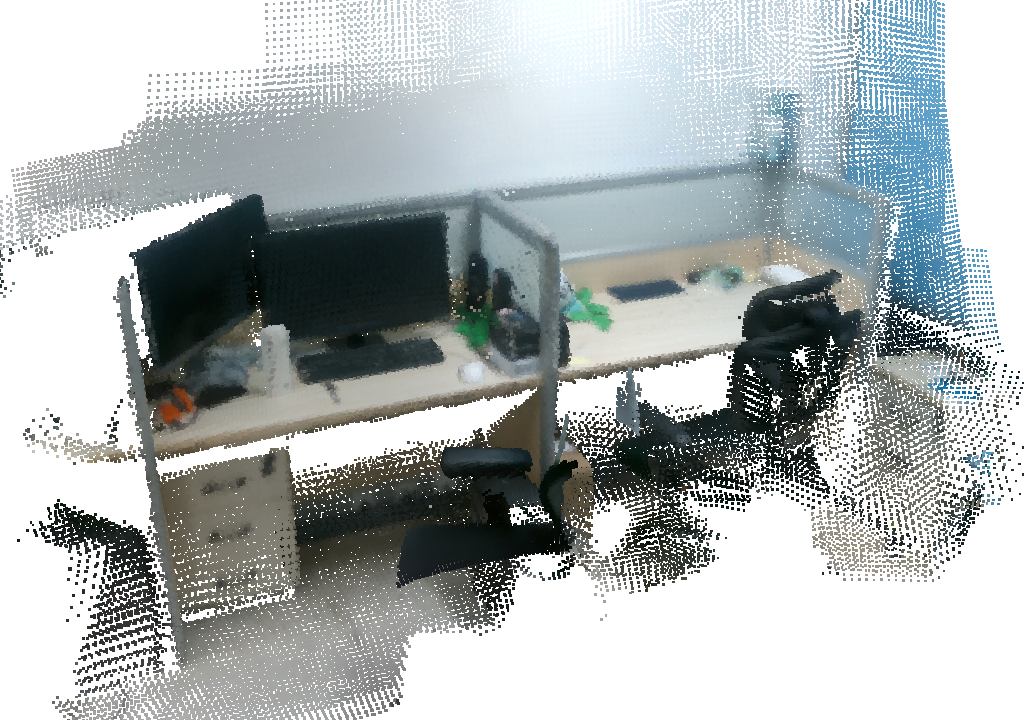} &
\includegraphics[width=0.20\textwidth]{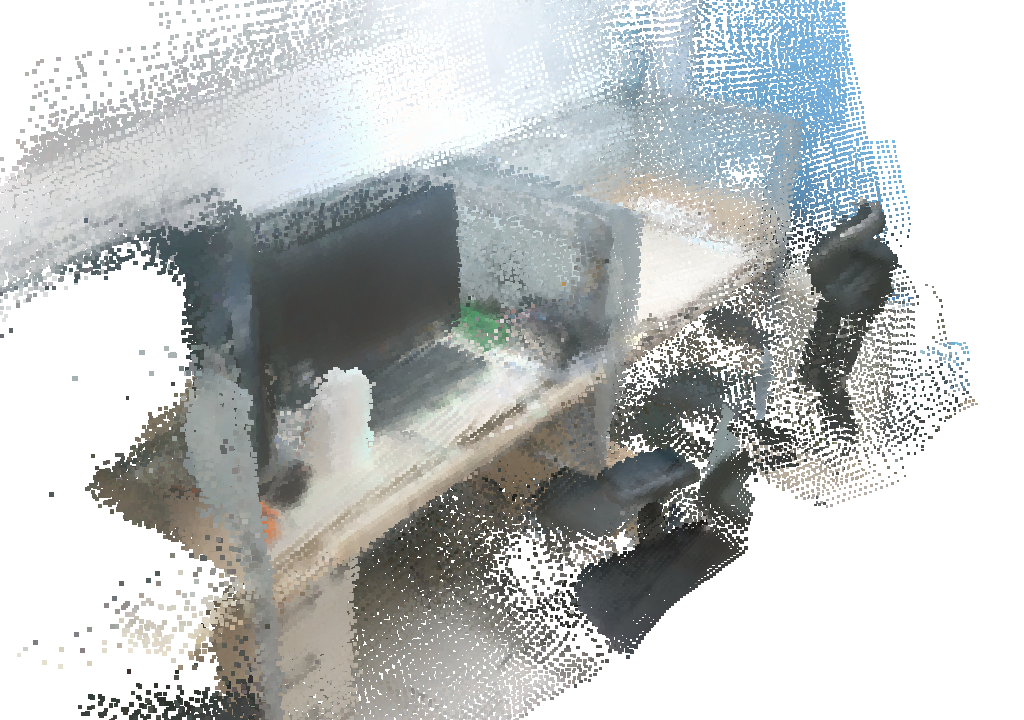} &
\includegraphics[width=0.20\textwidth]{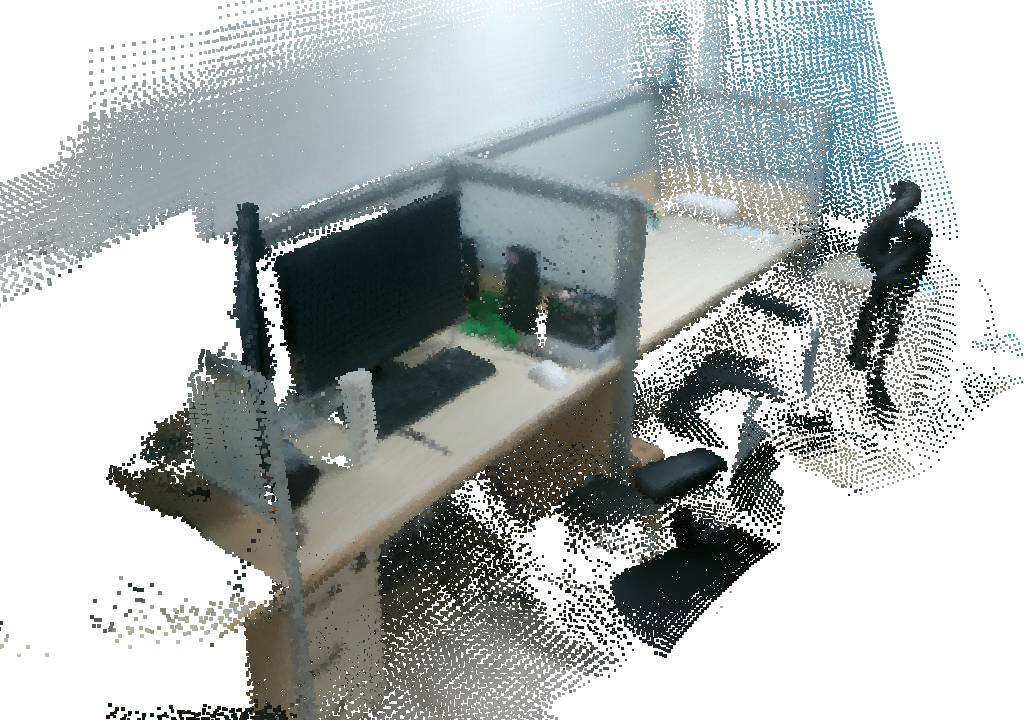} \\

\includegraphics[width=0.20\textwidth]{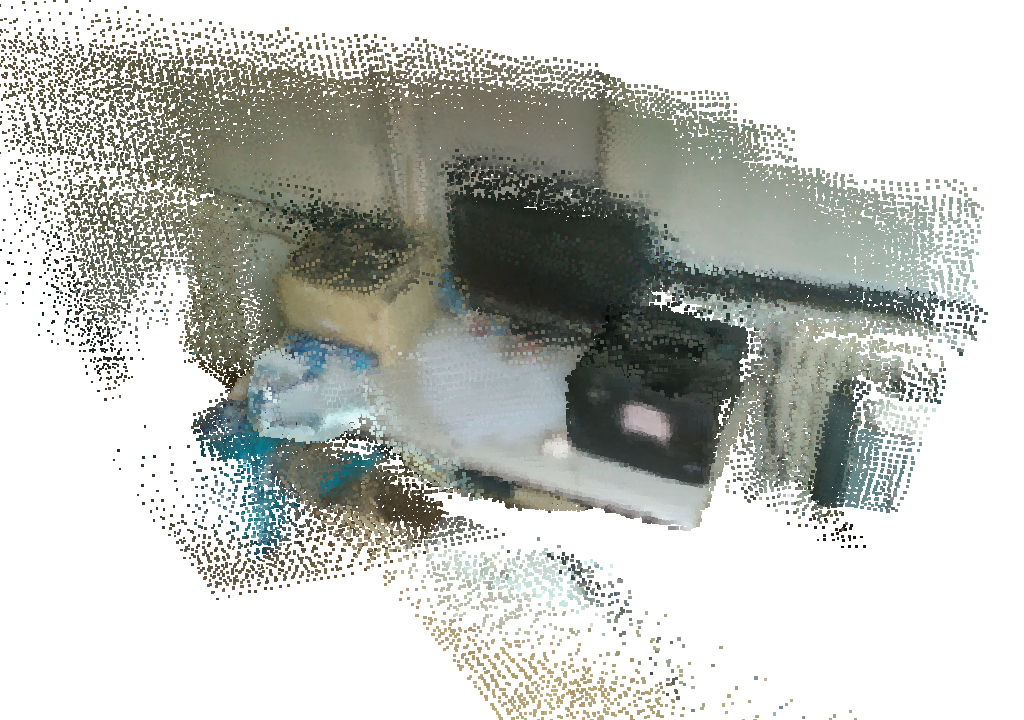} &
\includegraphics[width=0.20\textwidth]{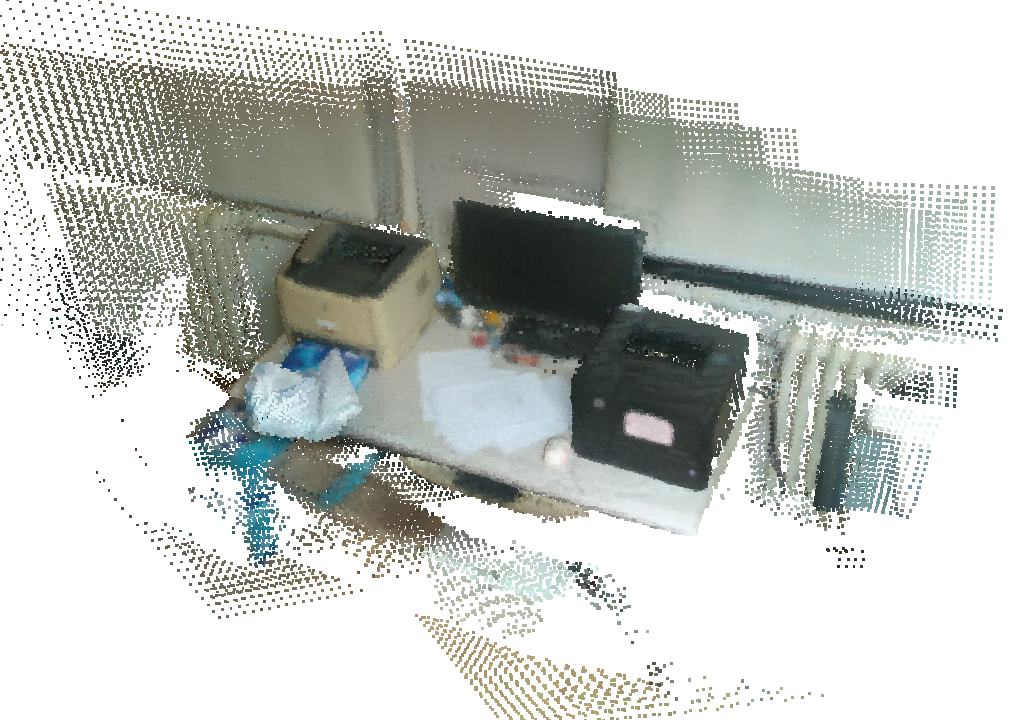} &
\includegraphics[width=0.20\textwidth]{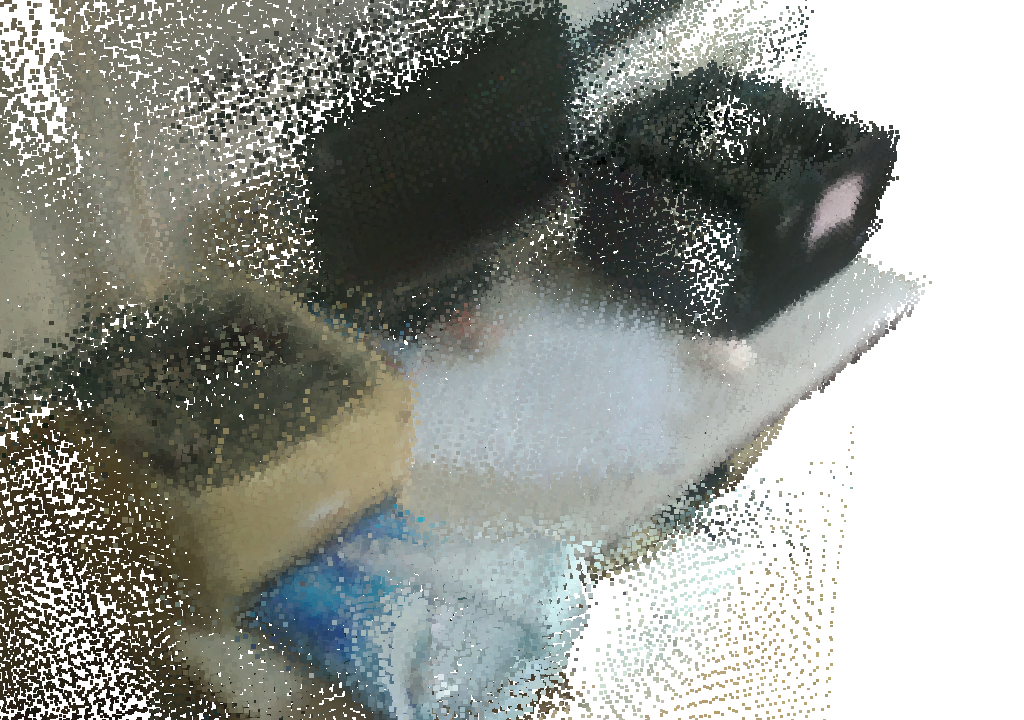} &
\includegraphics[width=0.20\textwidth]{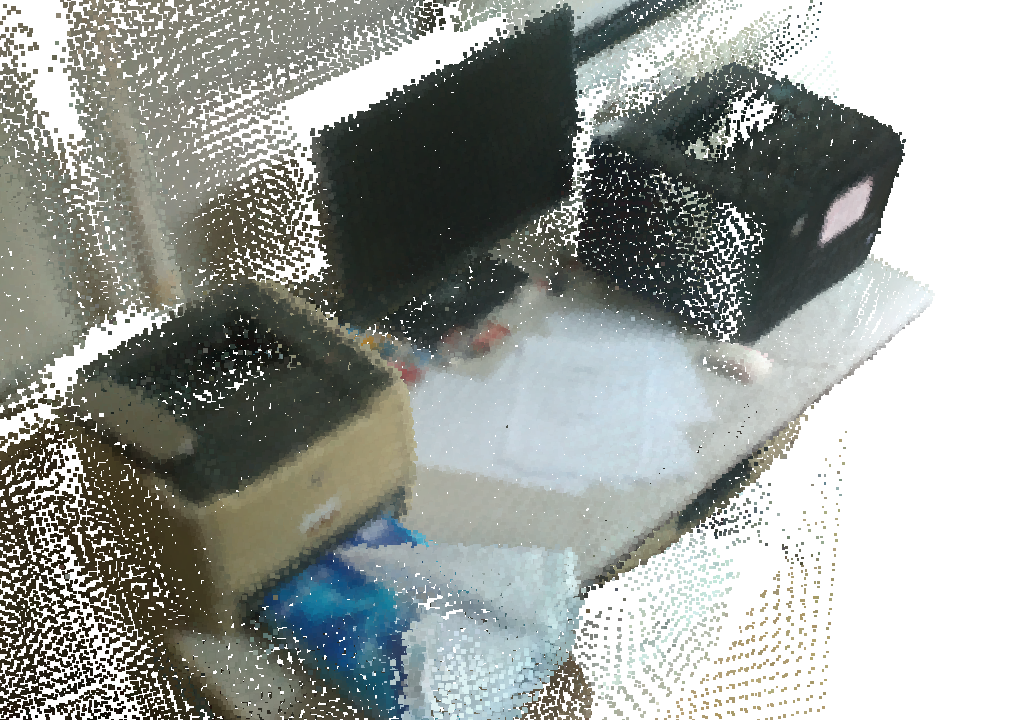} \\

\includegraphics[width=0.20\textwidth]{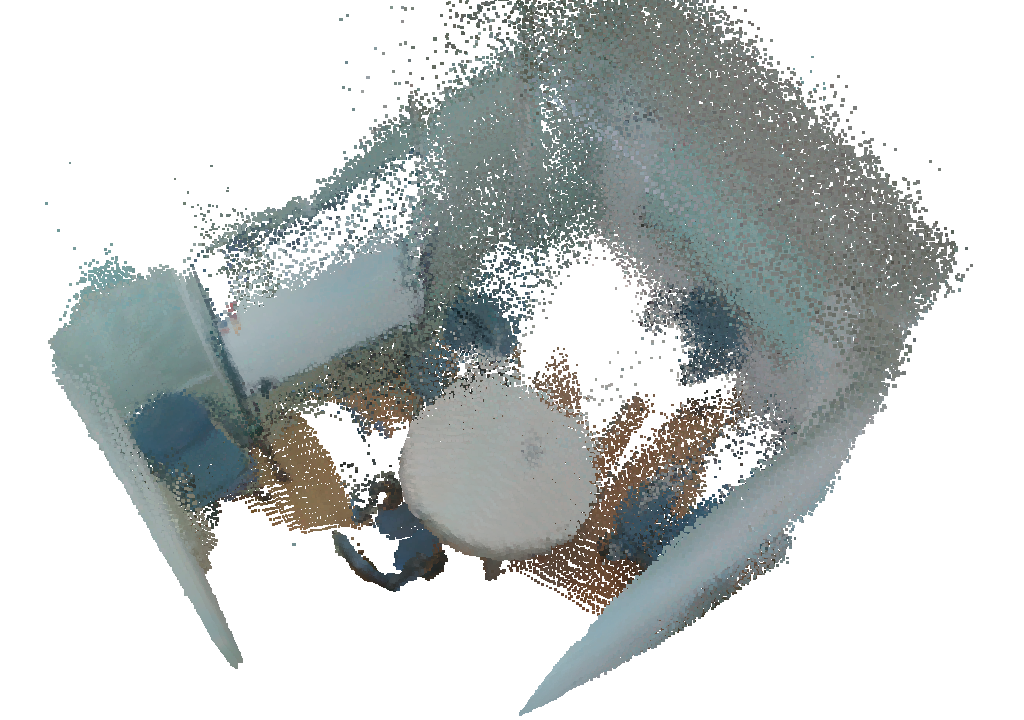} &
\includegraphics[width=0.20\textwidth]{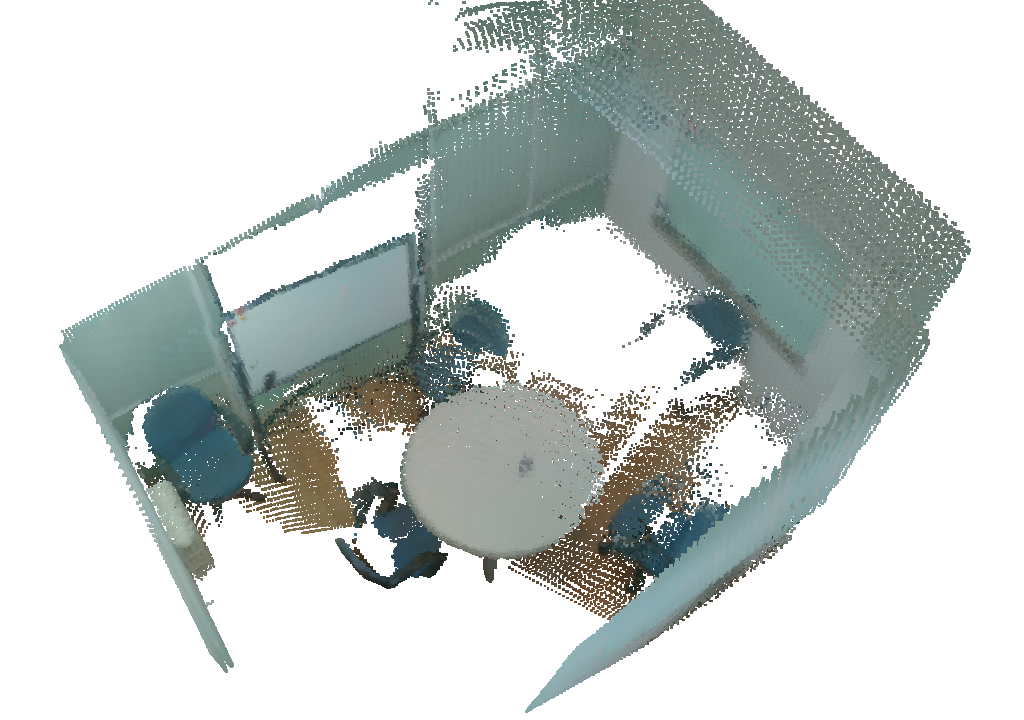} &
\includegraphics[width=0.20\textwidth]{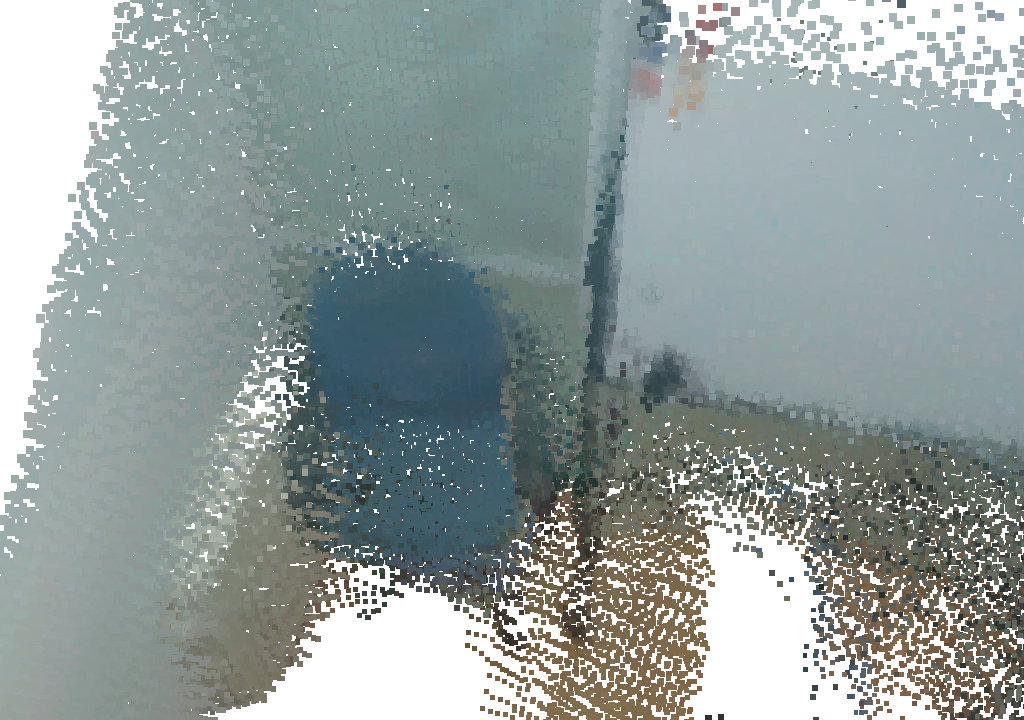} &
\includegraphics[width=0.20\textwidth]{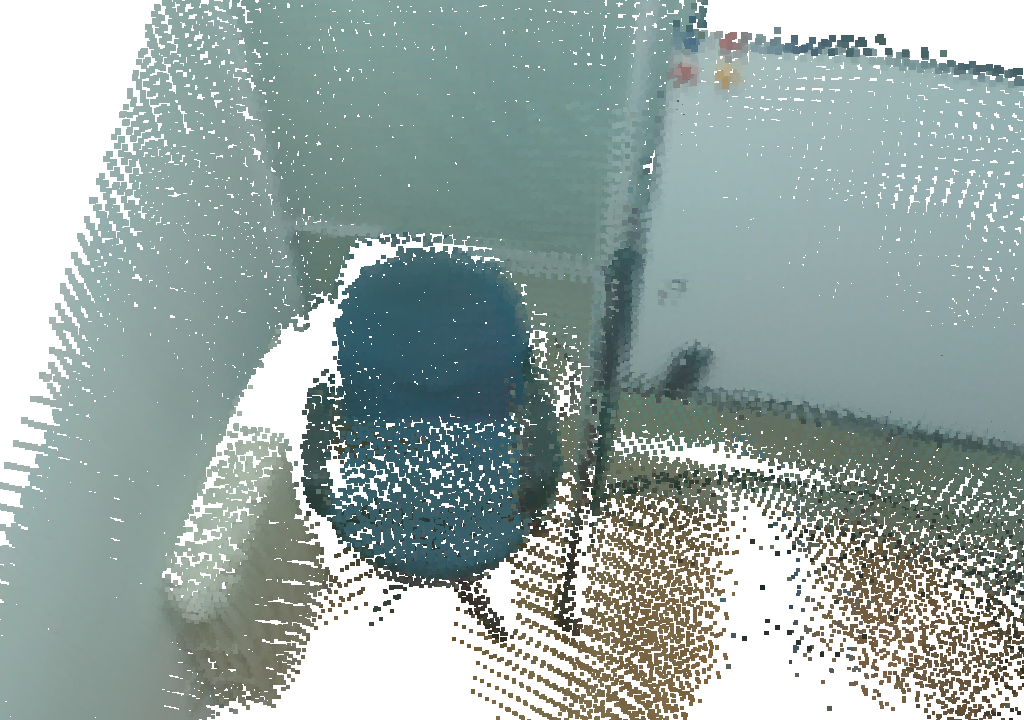} \\

\includegraphics[width=0.20\textwidth]{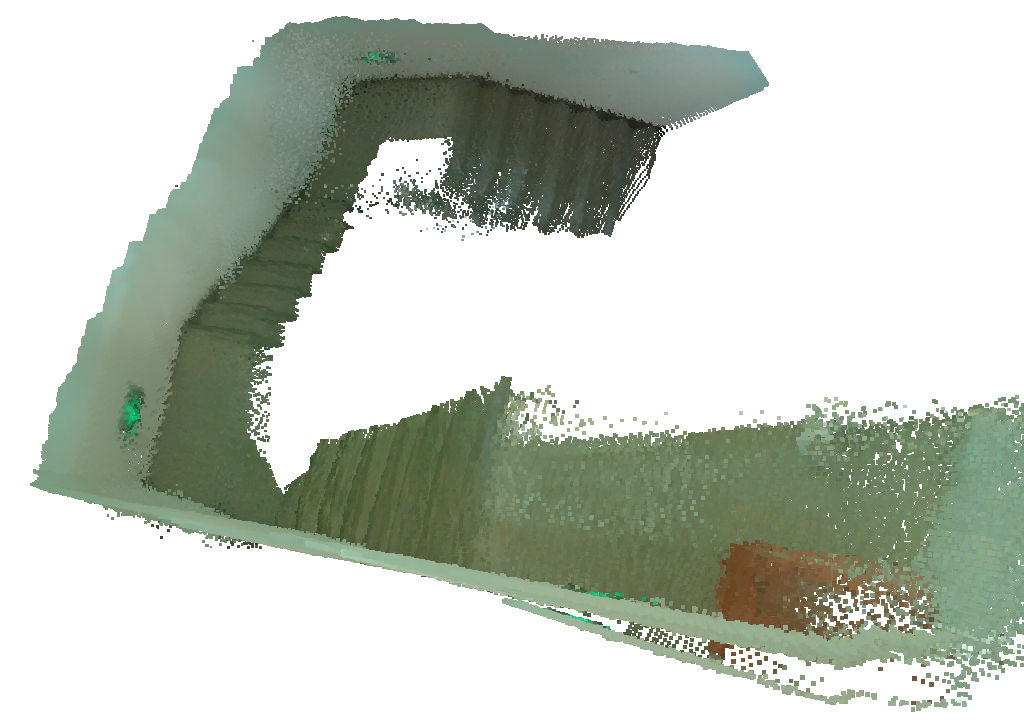} &
\includegraphics[width=0.20\textwidth]{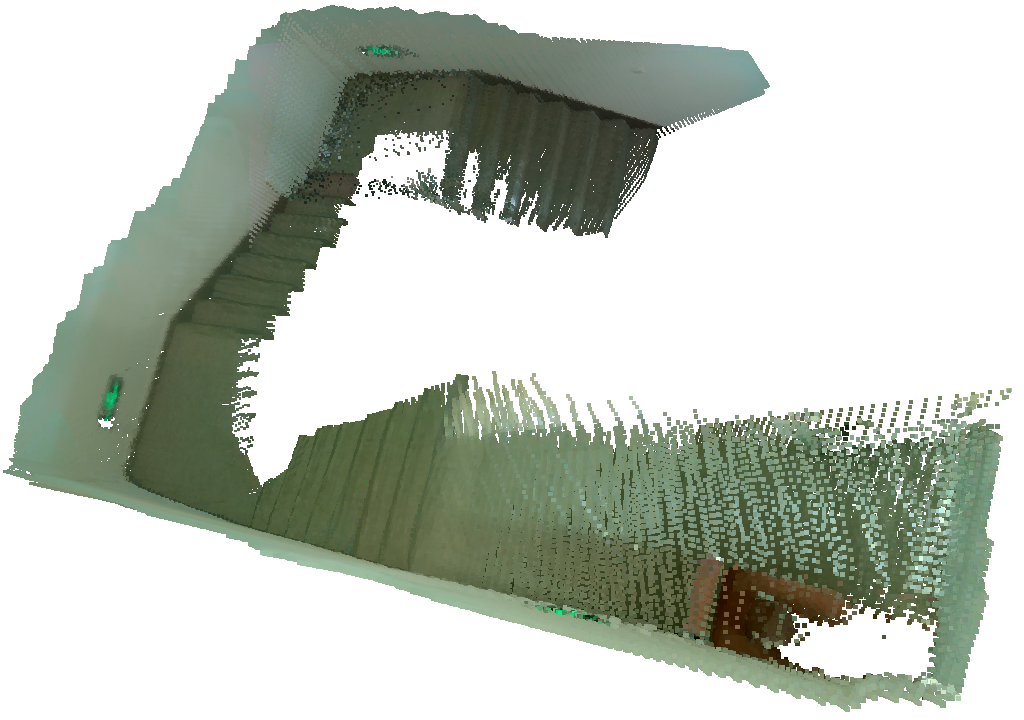} &
\includegraphics[width=0.20\textwidth]{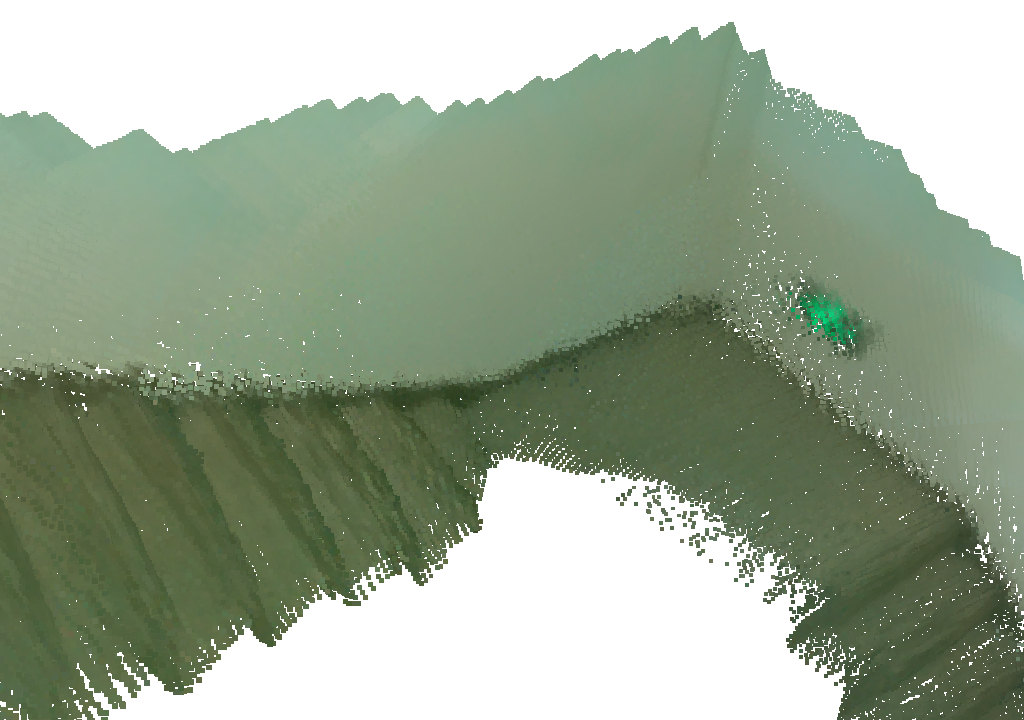} &
\includegraphics[width=0.20\textwidth]{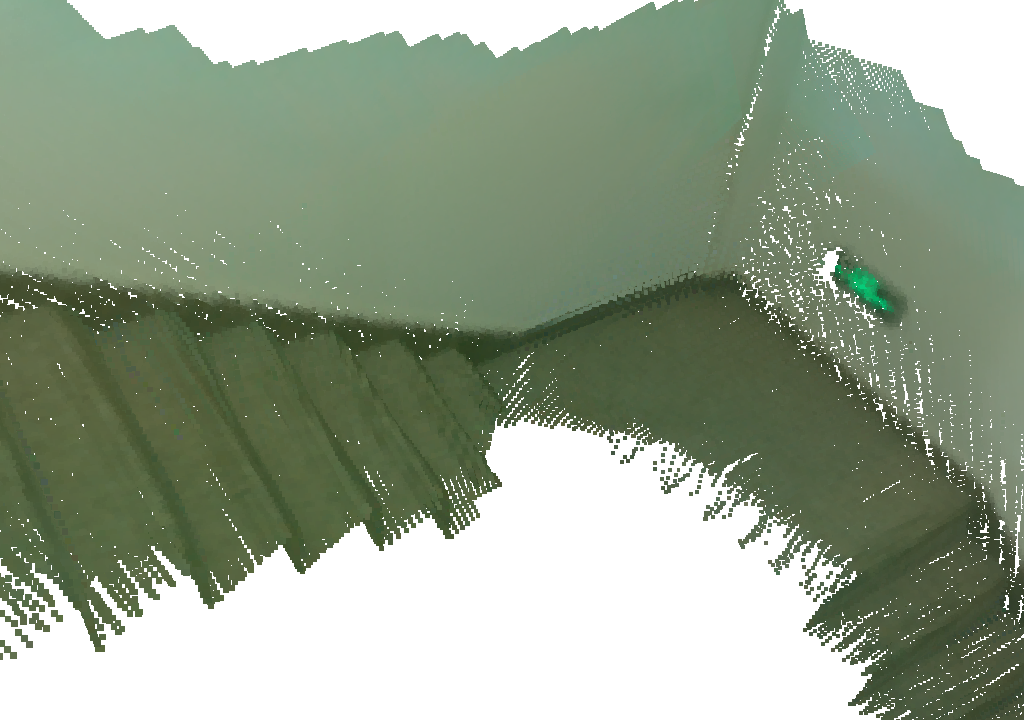} \\

\includegraphics[width=0.20\textwidth]{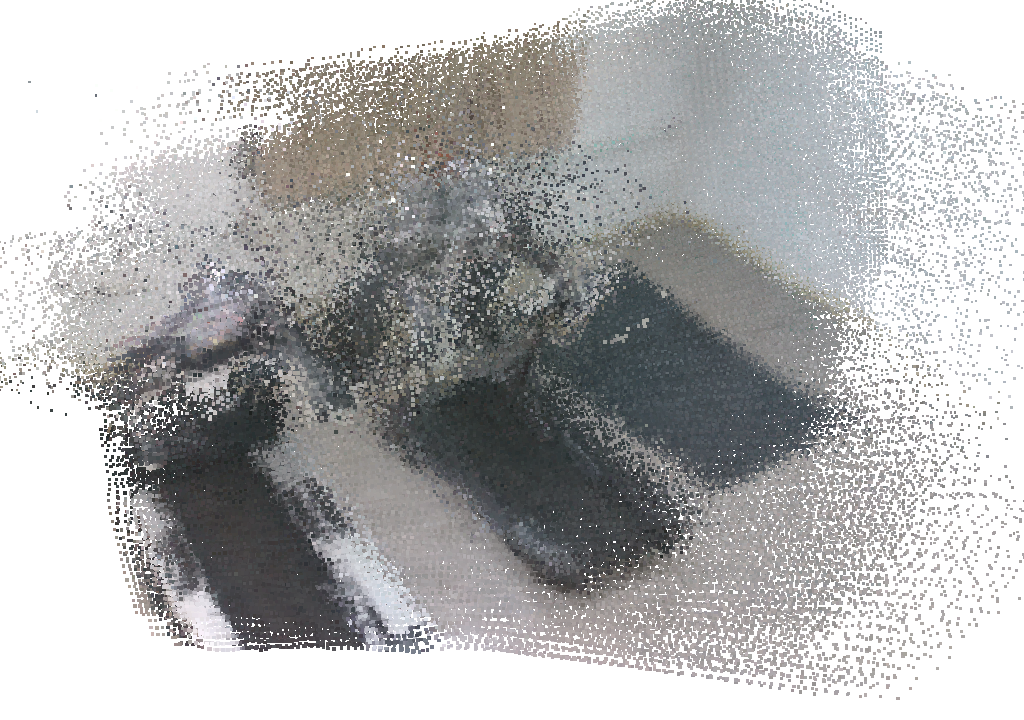} &
\includegraphics[width=0.20\textwidth]{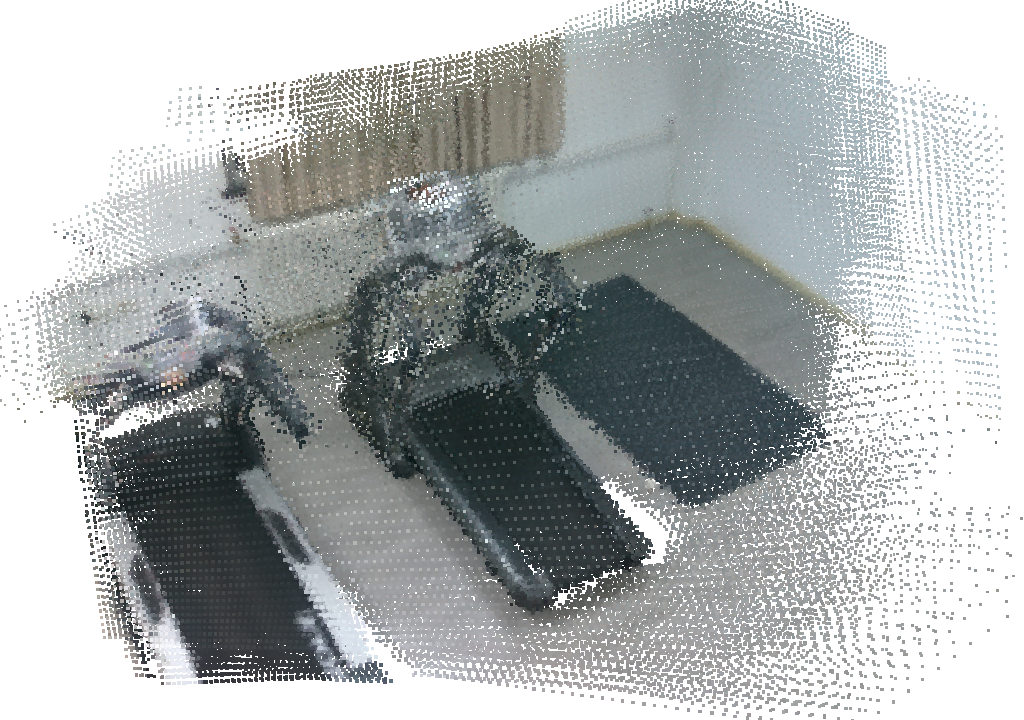} &
\includegraphics[width=0.20\textwidth]{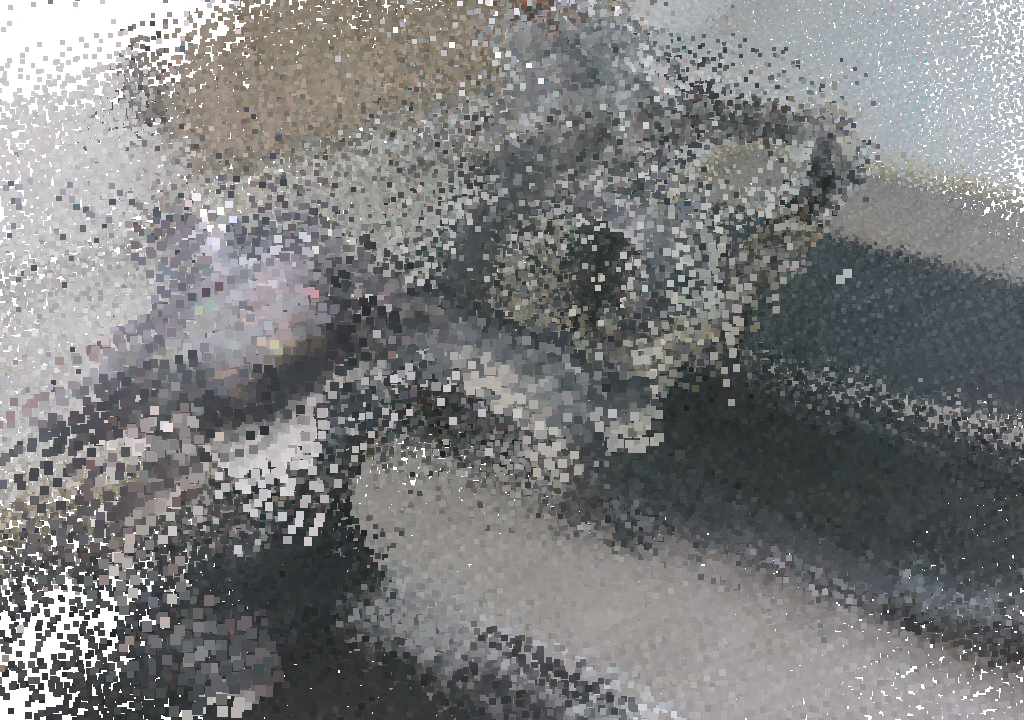} &
\includegraphics[width=0.20\textwidth]{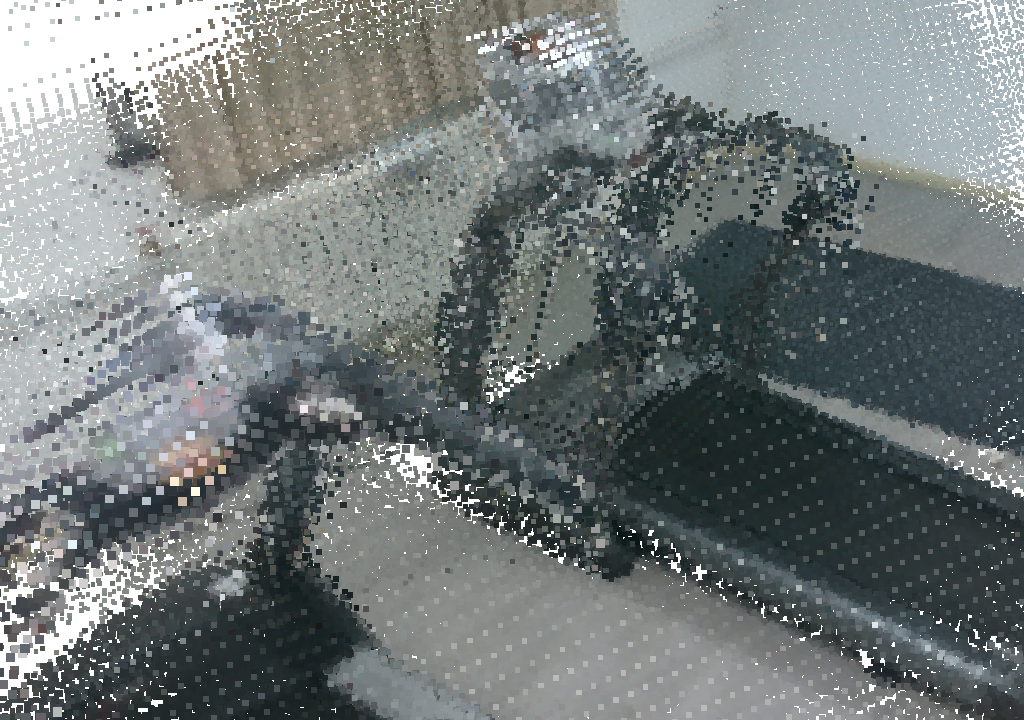} \\

\end{tabular}
\caption{Real-data reconstruction comparison between a commodity baseline formed by D435 sensor depth with the T265 reference trajectory and \ours{}. These are the rest 5 scenes not presented in the main text.}
\label{fig:supp_recon_qual}
\end{figure*}

Figure~\ref{fig:more_depth_qual} presents additional real-scene RGB observations and the corresponding predicted depth maps. Figure~\ref{fig:supp_recon_qual} shows additional qualitative reconstruction comparisons on the remaining real scenes.

\subsection{Cross-Sensor Qualitative Transfer}
\label{supp:cross_sensor_transfer}

The same synthetic-only NSL\_train model is also evaluated on RealSense D415 and Orbbec Gemini 335L. Despite differences in baseline, filtering, and projector patterns, the predicted depth and point clouds remain coherent. 

\begin{figure*}[t]
\centering
\setlength{\tabcolsep}{2pt}
\renewcommand{\arraystretch}{1.0}
\begin{tabular}{c c c c}
\multicolumn{4}{c}{\textbf{RealSense D415}} \\
RGB & IR & Predicted depth & Point cloud \\
\includegraphics[width=0.20\textwidth]{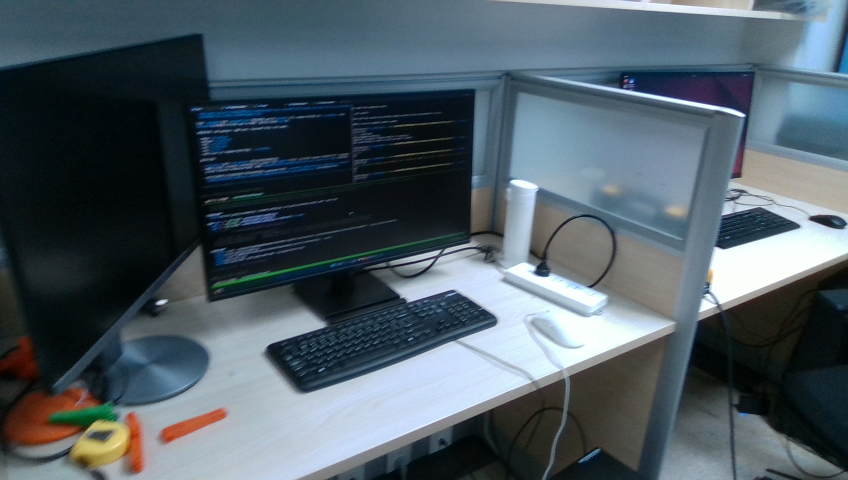} &
\includegraphics[width=0.20\textwidth]{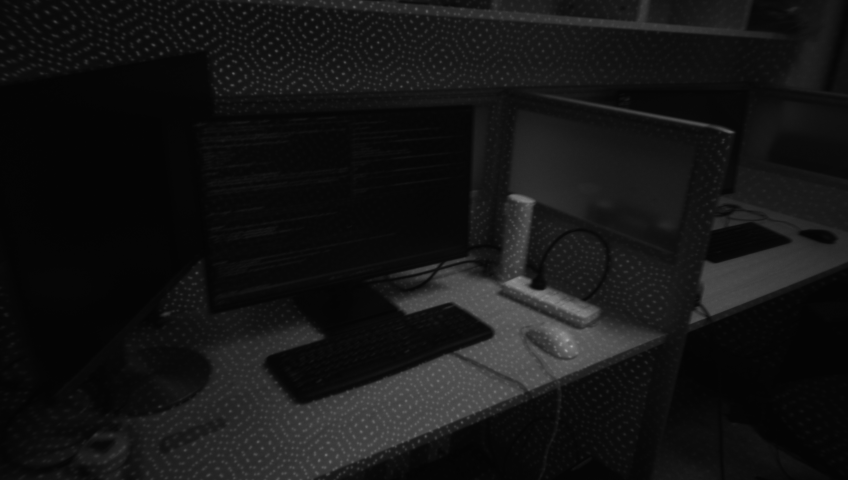} &
\includegraphics[width=0.20\textwidth]{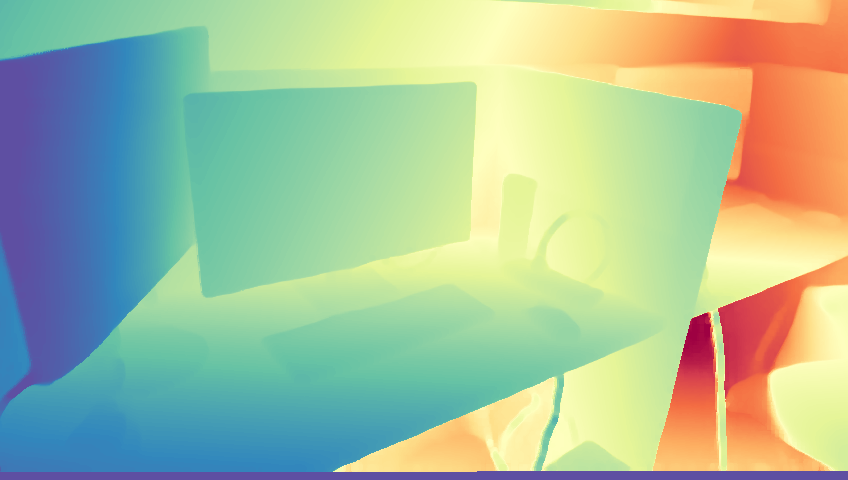} &
\includegraphics[width=0.20\textwidth]{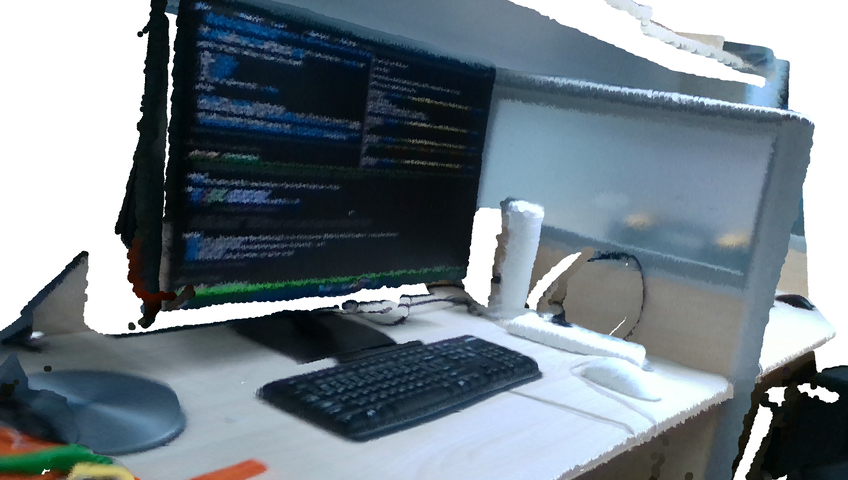} \\
\multicolumn{4}{c}{\textbf{Orbbec Gemini 335L}} \\
RGB & IR & Predicted depth & Point cloud \\
\includegraphics[width=0.20\textwidth]{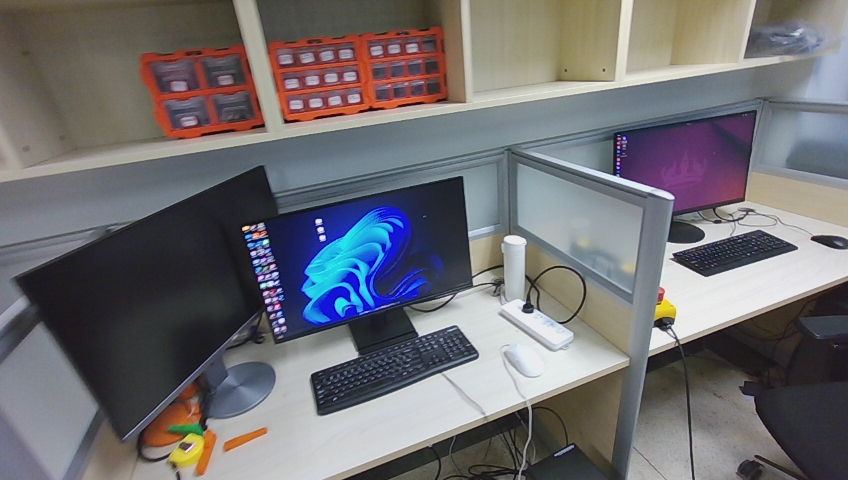} &
\includegraphics[width=0.20\textwidth]{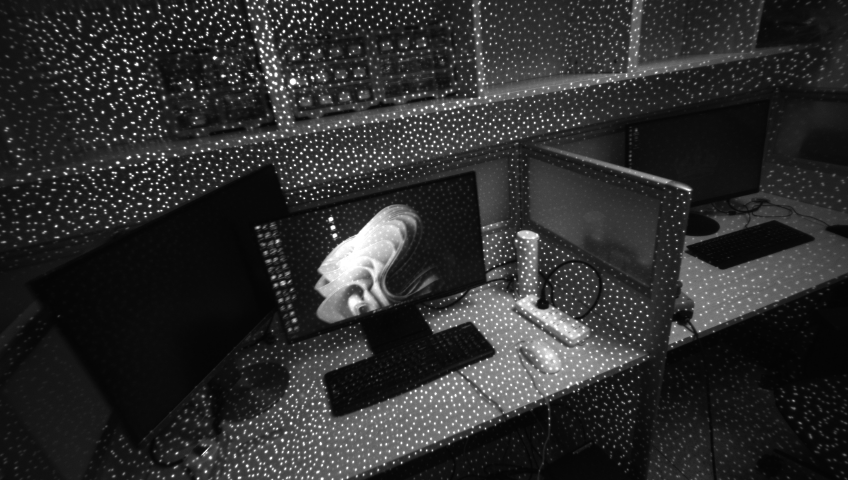} &
\includegraphics[width=0.20\textwidth]{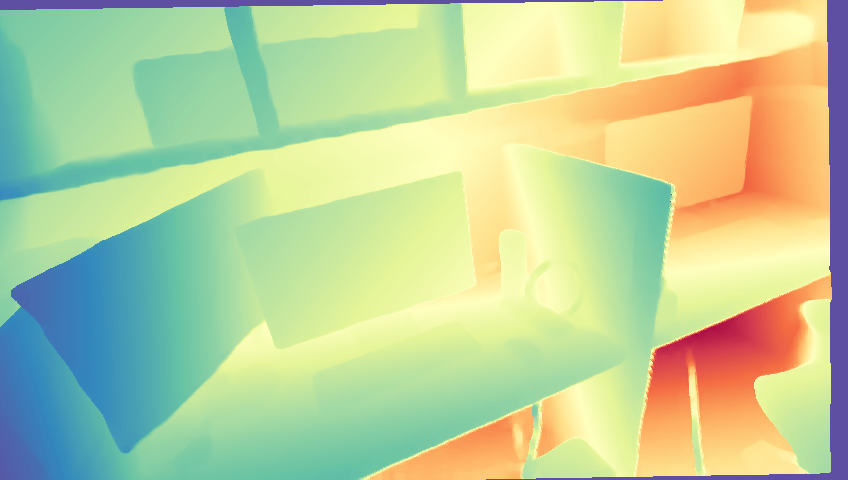} &
\includegraphics[width=0.20\textwidth]{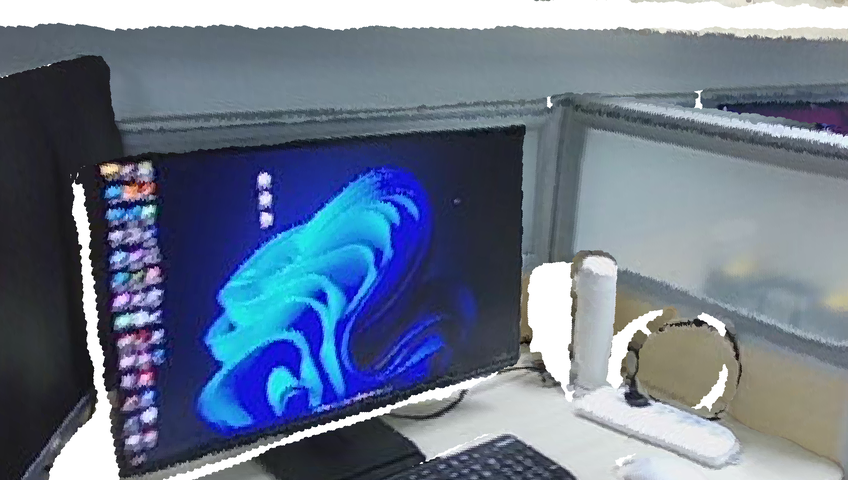} \\
\end{tabular}
\caption{Cross-sensor qualitative transfer from the same synthetic-only NSL\_train model to one capture per sensor from the RealSense D415 and Gemini 335L.}
\label{fig:supp_cross_sensor_transfer}
\end{figure*}

\end{document}